\documentclass{article}

\usepackage[utf8]{inputenc} 
\usepackage[T1]{fontenc}    
\usepackage{hyperref}       
\usepackage{url}            
\usepackage{booktabs}       
\usepackage{amsfonts}       
\usepackage{nicefrac}       
\usepackage{microtype}      

\usepackage{dsfont}
\usepackage{amssymb}
\usepackage{amsmath}
\usepackage{amsthm}
\usepackage{color}
\usepackage{algorithmic}
\usepackage{algorithm}
\usepackage{graphicx}
\usepackage{multirow}
\usepackage{enumerate}
\usepackage{verbatim}
\usepackage{hyperref}
\usepackage{microtype}
\usepackage{geometry}
\usepackage{footnote}
\usepackage{chngcntr}
\usepackage{caption}

\def\bbE{\mathop{\mathbb{E}}}

\newcommand{\pd}{\partial}
\newcommand{\grad}{\nabla}
\newcommand{\identity}{\mathbf{I}}

\def\bbR{\mathbb{R}}

\def\calF{\mathcal{F}}
\newcommand{\Svalid}{S_{valid}}

\def\sgn{\text{sgn}}
\def\proj{\text{proj}}
\def\Strain{S_{train}}
\def\calD{\mathcal{D}}
\newcommand{\Sinv}{S_{validinv}}

\title{Data Poisoning Attacks against Online Learning}

\author{
  Yizhen~Wang \\
  University of California, San Diego\\
  \texttt{yiw248@eng.ucsd.edu} \\
  \and
  Kamalika~Chaudhuri \\
  University of California, San Diego\\
  \texttt{kamalika@cs.ucsd.edu} \\
}

\begin{document}

\maketitle

\begin{abstract}
We consider data poisoning attacks, a class of adversarial attacks on machine learning where an adversary has the power to alter a small fraction of the training data in order to make the trained classifier satisfy certain objectives. While there has been much prior work on data poisoning, most of it is in the offline setting, and attacks for online learning, where training data arrives in a streaming manner, are not well understood.

In this work, we initiate a systematic investigation of data poisoning attacks for online learning. We formalize the problem into two settings, and we propose a general attack strategy, formulated as an optimization problem, that applies to both with some modifications. We propose three solution strategies, and perform extensive experimental evaluation. Finally, we discuss the implications of our findings for building successful defenses.
\end{abstract}

\section{Introduction}

As machine learning algorithms are increasing used in security-critical applications, there is a growing need to design them with active adversaries in mind. A class of adversarial attacks on machine learning that have received much attention is data poisoning attacks \cite{xiao2015support, xiao2015feature, xiao2012adversarial, burkard2017analysis, koh2017understanding, meizhu}. Here, an adversary is aware of the learner's training data and algorithm, and has the power to alter a small fraction of the training data in order to make the trained classifier satisfy certain objectives. For example, a sabotage adversary may try to degrade the overall accuracy of the trained classifier as part of an industrial sabotage campaign, or a profit-oriented adversary may try to poison the training data so that the resulting model favors it -- say, by recommending the its products over others.

While there has been a long line of prior work on data poisoning \cite{xiao2015support, xiao2015feature, xiao2012adversarial, burkard2017analysis, koh2017understanding, meizhu, biggio2013data, li2016data}, most of it has focussed in the offline setting, where a classifier or some other model is trained on a fixed input. In many applications of machine learning however, training is done online as data arrives sequentially in a stream; data poisoning attacks in this context are not well-understood.

In this work, we initiate a systematic investigation of data poisoning attacks for online learning. We begin by formalizing the problem into two settings, semi-online and fully-online, that reflect two separate use cases for online learning algorithms. In the semi-online setting, only the classifier obtained at the end of training is used downstream, and hence the adversary's objective involves only this final classifier. In the fully-online setting, the classifiers is updated and evaluated continually, corresponding to applications where an agent continually learns and adapts to a changing environment -- for example, a tool predicting price changes in the stock market. In this case, the adversary's objective involves classifiers accumulated over the entire online window.

We next consider online classification via online gradient descent, and formulate the adversary's attack strategy as an optimization problem; our formulation covers both semi-online and fully-online settings with suitable modifications and applies quite generally to a number of attack objectives. We show that this formulation has two critical differences with the corresponding offline formulation~\cite{meizhu} that lead to a difference in solution approaches. The first is that unlike the offline case where the estimated classifier is the minimizer of an empirical risk, here the classifier is a much more complex function of the data stream which makes gradient computation time-consuming.The second difference is that data order now matters, which implies that modifying training points at certain positions in the stream may yield high benefits; this property can be potentially exploited by a successful attack to reduce the search space.

We then propose a solution that uses three key steps. First, we simplify the optimization problem by smoothing the objective function and using a novel trick called label inversion if needed. Second, we recursively compute the gradient using Chain rule. The third and final step is to narrow down the search space of gradient ascent by modifying data points at certain positions in the input stream. We propose three such modification schemes -- the Incremental Attack which corresponds to an online version of the greedy scheme, the Interval Attack which corresponds to finding the best consecutive sequence to modify, and finally a Teach-and-Reinforce Attack, which seeks to teach a classifier that adheres to the adversary's goal by selectively modifying inputs in the beginning of the data stream, and then reinforce the teaching by uniformly modifying inputs along the rest of the stream. 

Finally, we carry out detailed experimentation which evaluates these attacks in the context of an adversary who seeks to degrade the classification error for four data sets, two settings (semi-online and fully-online) as well as three styles of learning rates. Our experiments demonstrate that online adversaries are indeed significantly more powerful in this context than adversaries who are oblivious to the online nature of the problem. Additionally, we show that the severity of the attacks depend on the learning rate as well as the setting; online learners with rapidly decaying learning rates are more susceptible to attacks, and so is the semi-online setting. We conclude with a brief discussion about the implication of our work for constructing effective defenses.

\paragraph{Related Work.} Prior work on data poisoning has mostly looked at offline learning algorithms. This includes attacks on spam filters~\cite{nelson2008exploiting} and malware detection~\cite{newsome2006paragraph} to more recent work on sentiment analysis \cite{newell2014practicality} and collaborative filtering \cite{li2016data}. \cite{biggio2012poisoning} developed the first gradient-ascent-based poisoning attack on offline SVMs; this technique was later extended to neural networks~\cite{munoz2017towards}. Mei and Zhu \cite{meizhu} propose a general optimization framework for poisoning attacks in the offline setting; they show how to use KKT conditions to efficiently calculate the gradients for gradient ascent. While our work is inspired by \cite{biggio2012poisoning} and \cite{meizhu}, the optimization problem in our case is considerably more complex and the KKT conditions in \cite{meizhu} no longer apply. Finally, Steinhardt \emph{et al.}~\cite{steinhardt2017certified} proposes a certified defense against offline data poisoning attacks.

In contrast, little is known about poisoning attacks in more online and adaptive settings. The work closest to ours is \cite{burkard2017analysis}, which poisons a data stream so that the final classifier misclassifies a pre-specified input; in contrast, our settings are more general. \cite{alfeld2016data} attacks an autoregressive model to yield desired outputs by altering the initial input; however their attack does not alter the model parameters, nor does it have access to the training process. Steinhardt \emph{et al.}~\cite{steinhardt2017certified} uses an online learning framework to generate a poisoning data set, but their attack and certified defense both apply to offline learning algorithms. Lastly, \cite{huang2017adversarial} and \cite{lin2017tactics} consider attacks against an reinforcement learning agent. However, the attacks only provide adversarial examples at test time using gradient methods; the policies are already learned and the attacker has no access to the training process.

\section{The Setting}

\label{sec:setting}
\subsection{The Classification Setting and Algorithm}

We consider online learning for binary linear classification. Specifically, we have instances $x$ drawn from the instance space $\bbR^d$, and binary labels $y \in \{ -1, +1 \}$. A linear classifier is represented by a weight vector $w$; for an instance $x$, it predicts a label $\sgn(w^{\top} x)$. For a particular classification problem, this weight vector $w$ is determined based on training data.

In offline learning, the weight vector $w$ is learnt by minimizing a convex empirical loss $\ell$ on training data plus a regularizer $\Omega(w)$. In contrast, in online learning -- the setting of this paper -- the training data set $S$ arrives in a stream $\{ (x_0, y_0), \ldots, (x_t, y_t), \ldots \}$; starting with an initial $w_0$, at time $t$, the weight vector $w_{t}$ is iteratively updated to $w_{t+1}$ based on the current example $S_t = (x_t, y_t)$. 

Since the classical work of~\cite{LW88}, there has been a large body of work in online learning in a number of settings~\cite{cesabianchibook}. In this work, we consider online learning in a distributional setting where each $(x_t, y_t)$ is independently drawn from an underlying data distribution $\calD$. As our learning procedure, we select the popular Online Gradient Descent (OGD) algorithm \cite{zinkevich2003online}, which at time $t$, performs the following update:
\[  w_{t+1} = w_t - \eta_t ( \grad \ell(w_t, (x_t, y_t)) + \grad \Omega(w_t) ),\]
Recall that here $\ell$ is a convex loss function, and $\Omega$ is a regularizer. For example, in the popular $L_2$-regularized logistic regression, $\ell$ is the logistic function: $\ell(w, (x, y)) = \log(1 + e^{-y w^{\top} x})$ and $\Omega(w) = \frac{c}{2} \| w \|^2$ for some constant $c$. In addition, $\eta_t$ is a learning rate that typically diminishes over time. We use $T$ to denote the length of the data stream. 

\subsection{The Attacker}

Following prior work on data poisoning~\cite{meizhu,koh2017understanding}, we consider a white-box adversary. Specifically, the adversary knows the entire data stream (including data order), the model class, the training algorithm, any hyperparameters and any defenses that may be employed. Armed with this knowledge, the adversary has the power to arbitrarily modify at most $K$ examples $(x_t, y_t)$ in the stream. While this is a strong assumption, it has been used in a number of prior works~\cite{meizhu,koh2017understanding,biggio2012poisoning,burkard2017analysis}. Security by obscurity is well-known to be bad practice and to lead to pit-falls~\cite{biggio2013data}. Additionally, adversaries can often reverse-engineer the parameters of a machine learning system~\cite{modelsteal2016,membershipinference2016}.

We consider two different styles of attack objectives -- {\em{semi-online}} and {\em{fully-online}} -- that correspond to two different practical uses of online learning. In the sequel, we use the notation $f(w)$ to denote an attack objective function that depends on a specific weight vector $w$.

\textbf{Semi-Online.} Here, the attacker seeks to modify the training data stream so as to maximize its objective $f(w_T)$ on the classifier $w_T$ obtained {\em{at the end of training}}. This setting is applicable when an online or streaming algorithm is used to train a classifier, which is then used directly in a downstream application. 

Here, the training phase uses an online algorithm but the evaluation of the objective is only at the end of the stream. Compared to the classic offline data-poisoning attack setting~\cite{biggio2012poisoning,steinhardt2017certified}, the attacker now has the extra knowledge on the order in which training data is processed.

\textbf{Fully-online.} Here, the attacker seeks to modify the training data so as to maximize the accumulated objectives over the entire online learning window.  More specifically, the attacker's goal is to now maximize $\sum_{t=1}^{T} f(w_t)$. 

This setting is called fully-online because both the training process and the evaluation of the objective are online. It corresponds to adversaries in  applications where an agent continually learns online, thus constantly adapting to a changing environment; an example of such a learner is a financial tool predicting price changes in the stock market.   

\paragraph{An Example Attack Objective: Classification Error.} Our algorithm and ideas apply to a number of objective functions $f$; however, for specificity, while explaining our attack, we will consider an adversary whose objective is the classification error achieved by the classifier output by the online learner:
\begin{equation} \label{eqn:classifobjective} f(w) = \Pr_{(x, y) \sim \mathcal{D}}( \sgn(w^{\top} x) \neq y)  = \bbE_{(x, y) \sim \calD} [ \mathds{1}(\sgn(w^{\top} x) \neq y) ] 
\end{equation}

\paragraph{The Feasible Set Defense.} Following prior work~\cite{biggio2012poisoning, steinhardt2017certified} will also assume that the learner employs a {\em{feasible set}} defense, where each example $(x_t, y_t)$ on arrival is projected on to a  feasible set $\calF$ of bounded diameter. This defense rules out trivial attacks where outlier examples with very high norm may be used to arbitrarily modify a classifier at training time. Observe that under the feasible set defense, it is sufficient to consider adversaries which only alter training points to other points inside the feasible set $\calF$.

For the purpose of this work, we will consider $\calF = [-1, 1]^{d} \times \{ -1, 1 \}$.

\section{Attack Methods}
\label{sec:attacks}

In the setting described above, finding the attacker's optimal strategy reduces to solving a certain attack optimization problem. We begin by describing this problem.

In the sequel, given a data stream $S$, we use the notation $S_t$ to denote the $t$-th labeled example in the stream. Additionally, the difference between two data streams $S$ and $S'$, denoted by $S \setminus S'$, is defined as the set $\{ S_t | S_t \neq S'_t \}$; its cardinality $| S \setminus S'|$ is the number of of time steps $t$ for which $S_t \neq S'_t$. 

\subsection{Attacker's Optimization Problem} 

Under this notation, the attacker's optimal strategy under an input data stream $\Strain$ in the semi-online case can be described as the solution to the following optimization problem.

\begin{align}
& \max_{S \in \calF^T} f(w_T) \label{eqn:objective}\\
& \mbox{subject to:} && | S \setminus \Strain | \leq K, \label{eqn:L0constraint} \\
&&& w_t = w_0 - \sum_{\tau=0}^{t-1} \eta_{\tau} ( \grad \ell (w_{\tau}, S_{\tau}) + \grad \Omega(w_{\tau})), 1 \leq t \leq T \label{eqn:OGDconstraint}
\end{align}

Here, \eqref{eqn:L0constraint} ensures that at most $K$ examples in the input data stream are changed, and \eqref{eqn:OGDconstraint} ensures that $w_t$ is the OGD iterate at time $t$. In the fully online case, the objective changes to $\sum_{t=1}^{T} f(w_t)$.

\subsection{Attack Algorithm}

\noindent{\textbf{Challenges and Our Approach.}} Observe that this optimization problem has some similarities to the corresponding offline problem derived by~\cite{meizhu}. However, there are two important differences that leads to a difference in solution approaches. 

The first is that unlike the offline case where the estimated weight vector is the minimizer of an empirical risk, here $w_t$ is a more complex function of the data stream $S$. This has two consequences that make gradient computation more challenging -- first, changing $x_t$ now influences all later $w_{\tau}$ for $\tau > t$; second, the KKT conditions that were exploited by prior work \cite{biggio2012poisoning, steinhardt2017certified, koh2017understanding} for easy computation no longer hold. The second difference from the offline case is that data order now matters, which implies that modifying training points at certain positions in the stream may yield high benefits. This property could be potentially exploited by a successful attack to reduce the search space. We next provide a solution approach that involves three steps, corresponding to three key ideas.

The first key idea is to simplify the optimization problem and make it amenable for gradient ascent; this is done by smoothing the objective function and using a novel tool called {\em{label inversion}} if needed. The second idea is to compute the gradient of the objective function with respect to points in the data stream by use of the chain rule and recursion. Finally, our third key idea is to narrow the search space by confining our search to specific positions in the data stream. We next describe each of these steps.

\paragraph{Idea 1: Simplify Optimization Problem.} We begin by observing that in many cases, the objective function~\eqref{eqn:objective} may not be suited for gradient ascent. An example is the classification error objective function~\eqref{eqn:classifobjective}; being a $0/1$ loss, it is non-differentiable, and second, it involves an expectation over $\calD$ which we cannot compute as $\calD$ is generally unknown.  In these cases, we smooth the objective function using standard tools -- we use a convex surrogate, the logistic loss, instead of the $0/1$ loss, and evaluating the expectation over a separate validation data set $\Svalid$ \cite{biggio2012poisoning}. 

However, the surrogate to the classification loss objective~\eqref{eqn:classifobjective} still has local maxima at the constant classifier. To address this, we propose a novel tool -- {\em{label inversion}}. Instead of maximizing logistic loss on the validation set, we construct an inverted validation set $\Sinv$, which contains, for each $(x, y)$ in the validation set, an example $(x, -y)$. We then maximize the negative of the logistic loss on this inverted validation set.

Formally, for an example $(x, y)$ and a weight vector $w$, let $L(w, (x, y))$ denote the negative logistic loss $L(w, (x, y)) = - \log(1 + \exp(-y w^{\top} x))$.  For the semi-online case, the objective~\eqref{eqn:objective} reduces to $\max_{S \in \calF^{T}} L(S) = \max_{S \in \calF^{T}} \sum_{(x,y)\in \Sinv} L(w_T, (x, y))$. Intuitively, this encourages the attacker to modify the data stream to fit a dataset whose labels are inverted with respect to the original data distribution. Again, an analogous objective can be derived for the fully online case.

\paragraph{Idea 2: Compute Gradients via Chain Rule.} The next challenge is how to compute the gradient of the objective function with respect to a training point $(x_t, y_t)$ at position $t$. Unlike the offline case, this is more challenging as the KKT conditions can no longer be used, and the gradient depends on position $t$. Our second observation is that this can still be achieved by judicious use of the chain rule.

We first observe that in the computation of any $w_{\tau}$, replacing a training point $(x_t, y_t)$ with $(-x_t, -y_t)$ leads to exactly the same result; this is due to symmetry of the feasible set as well as the OGD algorithm. We therefore choose to keep the label $y_t$'s fixed in the training set, and only optimize over the feature vector $x_t$s. 

Let $F(w_t)$ denote a (possibly) smoothed version of the objective $f$ evaluated at the $t$-th iterate $w_t$; using Chain Rule we can write:
\begin{equation}
 \frac{\pd F(w_t)}{\pd x_i} = \left\{
	\begin{array}{ll}
	0 & \mbox{if $i\geq t$}\\
         \frac{\pd F(w_t)}{\pd w_{t}}\frac{\pd w_{t}}{\pd x_i}&  \mbox{if $i=t-1$}\\
         \frac{\pd F(w_t)}{\pd w_{t}}
         \frac{\pd w_{t}}{\pd w_{t-1}}\cdots \frac{\pd w_{i+2}}{\pd w_{i+1}}
         \frac{\pd w_{i+1}}{\pd x_i} & \mbox{if $i<t-1$}.
    \end{array}\right\}
\label{eq:gradgeneral}
\end{equation}

Observe that the term $\pd F(w_t)/\pd w_t$ and $\pd w_{i+1}/\pd x_i$ may be calculated directly, and the product 
$\frac{\pd F(w_t)}{\pd w_t} \cdot \frac{\pd w_{t}}{\pd w_{t-1}}\cdots \frac{\pd w_{i+2}}{\pd w_{i+1}}$ 
may be calculated via $t - i$ matrix-vector multiplications. Finally, we observe that since the gradients $\pd F(w_t)/\pd x_i$ and $\pd F(w_t)/\pd x_{i+1}$ share the prefix as shown in Appendix~\ref{sec:prefix} and hence the gradients $\{ \pd F(w_t)/\pd x_{i} \}$ for all $i$ may be computed in $O(Td^2)$ time. A detailed derivation of the general case, as well as specific expressions for the surrogate loss to classification error, is presented in the Appendix~\ref{sec:gradlr}.

\paragraph{Idea 3: Strategic Search over Positions in the Stream.} So far, we have calculated the impact of modifying a single point $(x_t, y_t)$ on the objective function; a remaining question is which data points in the input stream to modify. Recall that in a data stream of length $T$, there are potentially $\binom{T}{K}$ subsequences of $K$ inputs to modify, and a brute force search over all of them is prohibitively expensive. We present below three algorithms for strategically searching over positions in the stream for attack purposes -- the Incremental Attack, the Interval Attack and the Teach-and-Reinforce Attack. 

A first idea is to consider a {\em{greedy approach}}, which iteratively alters the training point in the sequence that has the highest gradient magnitude. While this has been considered by prior work for offline settings, we next derive how to do this in an online setting. This results in what we call the {\em{Incremental Attack}}.

\smallskip\noindent{\textbf{Incremental Attack.} This attack employs an iterative steepest coordinate descent approach. In each iteration, we calculate the gradient of the modified objective with respect to each single training point, and pick the $(x_{t}, y_{t})$ with the largest gradient magnitude. This example is then updated as: 
\[ x_{t} \leftarrow \proj_{\calF}( x_{t} + \epsilon \pd F / \pd x_{t}), \]
where $\epsilon$ is a step size parameter, $F$ is the objective function, and $\proj$ is the projection operator. The process continues until either $K$ distinct points are modified or convergence. The full algorithm is described in Algorithm~\ref{algo:incremental} in the Appendix.

Observe however that the greedy approach has two limitations. The first is that since it does not narrow the search space using the online nature of the problem, it involves a large number of iterations and is computationally expensive. The second is that it is known to be prone to local minima as observed by prior work in the offline setting. \cite{steinhardt2017certified}

\smallskip\noindent{\textbf{Interval Attack.}} To address these limitations, we next propose a novel strategy that is tailored to the online setting, called the {\em{Interval Attack}}. The idea here is to find the  best {\em{consecutive sequence}} $(x_t, y_t), \ldots, (x_{t + K -1}, y_{t+K-1})$ modifying which decreases the objective function the most. 

More specifically, the search process is as follows. For each $t \in \{ 0, \ldots, T - K - 1\}$, the algorithm computes the concatenated gradient $[\pd F/\pd x_{t}, \cdots, \pd F/\pd x_{t+K-1}]$ and carries out the following update until convergence or until a maximum number of iterations is reached:
\[ [x_t, \cdots, x_{t+K-1}] \leftarrow \proj_{\calF^K}([x_t, \cdots, x_{t+K-1}] + \epsilon [\pd F/\pd x_{t}, \cdots, \pd F/\pd x_{t+K-1}]), \]
where $\epsilon$ is again a step-size parameter. The value of $t$ for which the final $[x_t, \ldots, x_{t + K -1}]$ gives the best objective~\eqref{eqn:objective} is then selected. 

\smallskip\noindent{\textbf{Teach-and-Reinforce Attack.}} Even though the Interval attack is faster than the Incremental, it has the highest impact on the performance of the classifiers $w_{t'}$ for $t'$ close to $t$. In particular, it is suboptimal in the fully online case, where the objective is the sum of the losses of all $w_t$. This motivates our third attack strategy -- the {\em{Teach-and-Reinforce Attack}}, which seeks to first {\em{teach}} a new classifier by modifying a set of initial examples, and then {\em{reinforce}} the teaching by modifying examples along the rest of the data stream.

We split $K$, the number of examples to be modified, into two parts $\alpha K$ and $(1 - \alpha)K$. We then modify the first $\alpha K$ points in the stream, followed by every $s$-th remaining point, where $s = \lceil \frac{T-\alpha K}{(1-\alpha) K}\rceil$. As in previous attacks, once the attack positions are determined, the algorithm iteratively finds the best modification through gradient ascent followed by projection to the feasible set. The full algorithm is described in Algorithm~\ref{algo:teachreinforce}. Finally, the optimal value of $\alpha$ is determined through a grid search. Observe that $\alpha=0$ corresponds to modifying points over an uniform grid of positions in the stream, and hence the performance of Teach-and-Reinforce is at least as good as this case. 

\section{Experiments}

We now evaluate the proposed attacks experimentally to determine their practical performance. In particular, we consider the following three questions:

\begin{enumerate}

\item How effective are the proposed gradient ascent-based online adversaries relative to adversaries who are oblivious to the online nature of the learning process, and adversaries that simply invert training labels?

\item Which positions in the data stream are predominantly attacked by the online adversaries?

\item How does attack performance vary with the setting (semi-online vs fully-online) and learning rate?
\end{enumerate}

These questions are investigated in the context of the classification error objective for four data sets, two settings (semi-online and fully-online) as well as three styles of learning rates.

\subsection{Experimental Methodology}

\smallskip\noindent{\textbf{Baselines.}} We implement the three proposed online attacks -- Incremental, Interval and Teach-and-Reinforce -- in both semi-online and fully-online settings. For computational efficiency, in the fully-online setting, we use the objective function $\sum_{t \in G} f(w_t)$ where $G$ is a regular grid containing every $10$-th integer. 

Additionally, we consider two baselines -- the Offline attack and the Label-flip Attack. The Offline attacker represents an adversary who is oblivious to the streaming nature of the input. It uses the method of~\cite{biggio2012poisoning} to generate $K$ additional attack points. Since their method generates points for adding to the training data, and our input stream is fixed-length, we replace a random set of $K$ positions in the stream with the newly generated points. The Label Flip attack selects $K$ points from the input stream, and flips their labels. We use three strategies for selecting the positions of these points -- head-flip (initial $K$ positions in the stream), tail-flip (the final $K$ positions), and random. To keep the figures understandable, we report the best outcome acheived by the three strategies.

\smallskip\noindent{\textbf{Datasets.}} We select four datasets -- a synthetic dataset consisting of a $2$-dimensional mixture of spherical  Gaussians, MNIST, fashion MNIST and UCI Spambase.  To reduce the running time, we project both MNIST and fashion MNIST to $50$ dimensions via random projections. For MNIST, we use the 1 vs. 7 classification task, and for fashion MNIST, the sandals vs. boots task.  

\smallskip\noindent{\textbf{Parameter Choice.}} The online learner has three parameters of interest -- the initial classifier $w_0$, the regularization function, and the learning rate $\eta_t$. We set the initial classifier $w_0$ in Online Gradient Descent to a logistic regression classifier trained offline on a held-out dataset. For all cases, we use $L_2$ regularization $\Omega(w) = \frac{\lambda}{2} \|w\|^2$ with parameter $\lambda = 0.4$. We consider three choices for learning rate -- Constant (where $\eta_t = \eta_0$), Slow Decay (where $\eta_t = \eta_0/\sqrt{t}$) and Fast Decay (where $\eta_t = \eta_0 /\lambda t$) -- in accordance with standard practices. $\eta_0$ is chosen to ensure that in all cases, the online learner has more than $90\%$ test accuracy when run on the clean data stream.

The gradient based attack algorithms also involve a step size parameter $\epsilon$, which is initially set to $\sqrt{d}/100$ where $d$ is the data dimension, and then decays as $\sqrt{d}/[100\sqrt{1+(t/20)}]$. Finally, the hyper-parameter $\alpha$ in the Teach-and-Reinforce attack is chosen by grid search from $\{0, 0.25, 0.5, 0.75\}$; observe that $\alpha=0$ corresponds to modifying points over an uniform grid of positions in the stream. 

\subsection{Results}

Figures~\ref{fig:expres}  show the attack results for synthetic, MNIST and Spambase for the Slow decay learning rate; the remaining figures are in the Appendix~\ref{sec:resplots}. Each plotted point reports average test accuracy as a function of the fraction of modified points, and is an average over $8$ runs.  Table~\ref{tab:positions} provides a qualitative description of the positions in the stream that are attacked by the different methods, with full histograms in the Appendix~\ref{sec:attackpositions}. 

The figures show that the Incremental and Interval attacks are overall the most effective in the semi-online case, while Teach-and-Reinforce is overall the most effective in the fully-online case. Additionally, the gradient ascent-based online attacks are highly effective for all datasets (except for the synthetic 2-d mixture of Gaussians), and have higher performance than the position-oblivious Offline attack as well as the Label Flip attack. Finally, the positions attacked by Incremental and Interval attacks do change as a function of the setting (semi-online vs. fully-online) and learning rate style. In the semi-online case, for Slow Decay and Constant, the attacks modify points towards the end of the stream, while in the fully-online case, the modified points are chosen towards the beginning. In constrast, for Fast Decay, the points modified are chosen from the beginning of the stream in both cases.

\subsubsection{Discussion} 

We now revisit our initial questions in light of these results. 

\smallskip\noindent{\textbf{Comparison with Oblivious Adversaries and Label Flip Adversaries.}} Both Offline and Label Flip fail to perform as well as the best online attack. This implies that an adversary who can exploit the online nature of the problem can produce better attacks than an oblivious one. Additionally, gradient ascent procedures are indeed necessary for high performance, and simple label flipping does not suffice. 

\smallskip\noindent{\textbf{Implications of Attack Positions.}} We find that the positions modified by the Incremental and Interval attacks are highly concentrated in the semi-online case. This implies that we may be able to narrow down the search space of attack positions in this case. However, such short-cuts may not yield high performance in the fully-online case.

\smallskip\noindent{\textbf{Impact of Setting and Learning Rates.}} The setting (semi-online vs. fully-online) and learning rates significantly impact the results. The results there imply that first, the Incremental Attack suffers from local minima which are mitigated by Teach-and-Reinforce, and second, the semi-online setting may be easier to attack than fully-online. Additionally, the efficacy of all three online attacks is higher in the Fast Decay case than Slow Decay or Constant, which suggests that Fast Decay learning rates may be particularly vulnerable to adversarial attacks. 

\subsubsection{Implications for Defenses}

Our experiments show that the semi-online setting is more vulnerable than the fully-online setting, and the Fast Decay learning rate is also more vulnerable than Slow Decay or Constant Decay. We conjecture that this is because the classifier $w_T$ in the semi-online setting depends heavily on a relatively smaller number of training points than the fully-online setting. A similar explanation applies to Fast Decay; because the learning rate decays rapidly, the classifiers produced depend rather critically on the few initial points. 

Based on these results, we therefore recommend the use of online methods where no one classifier depends heavily on a few input points; we conjecture that these methods would be less vulnerable to adversarial attacks. An example of such a method is the averaged stochastic gradient classifier, where the classifier used in iteration $t$ is the average $\frac{2}{t} \sum_{s=t/2}^{t} w_s$. An open question for future work is to investigate the vulnerability of this method. 

\begin{figure}[h]
    \centering
    \resizebox{\columnwidth}{!}{%
    \includegraphics{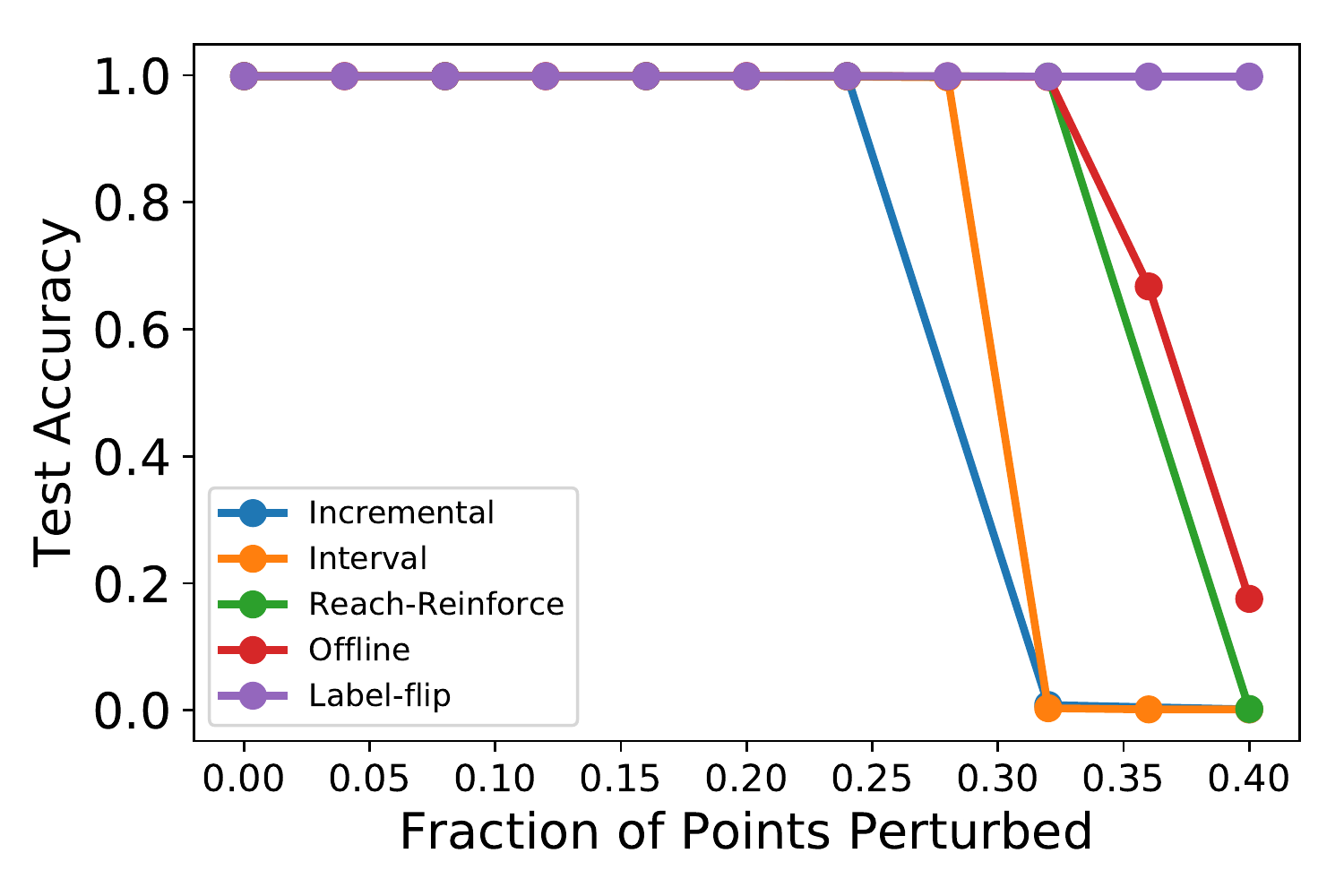}
    \includegraphics{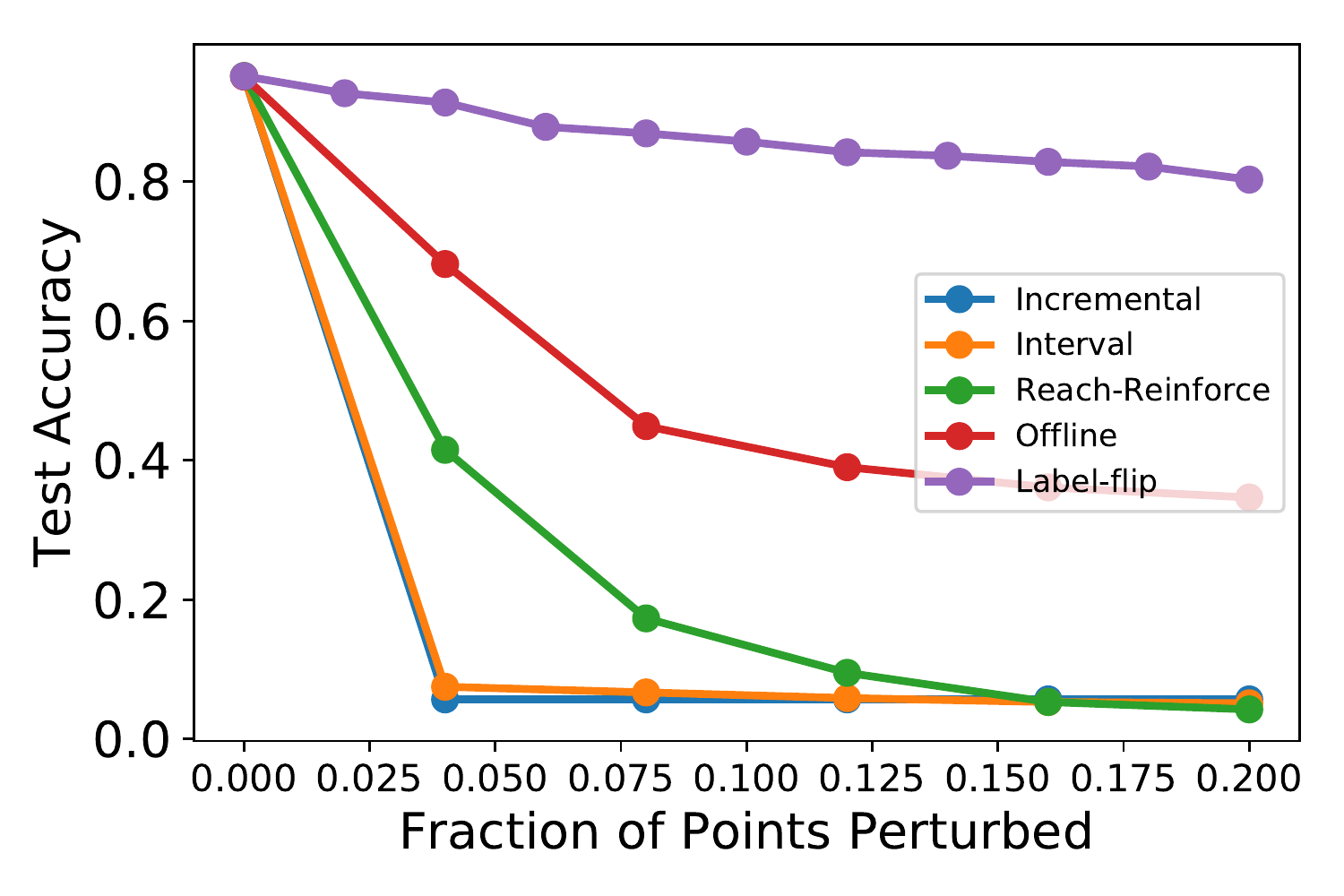}
    \includegraphics{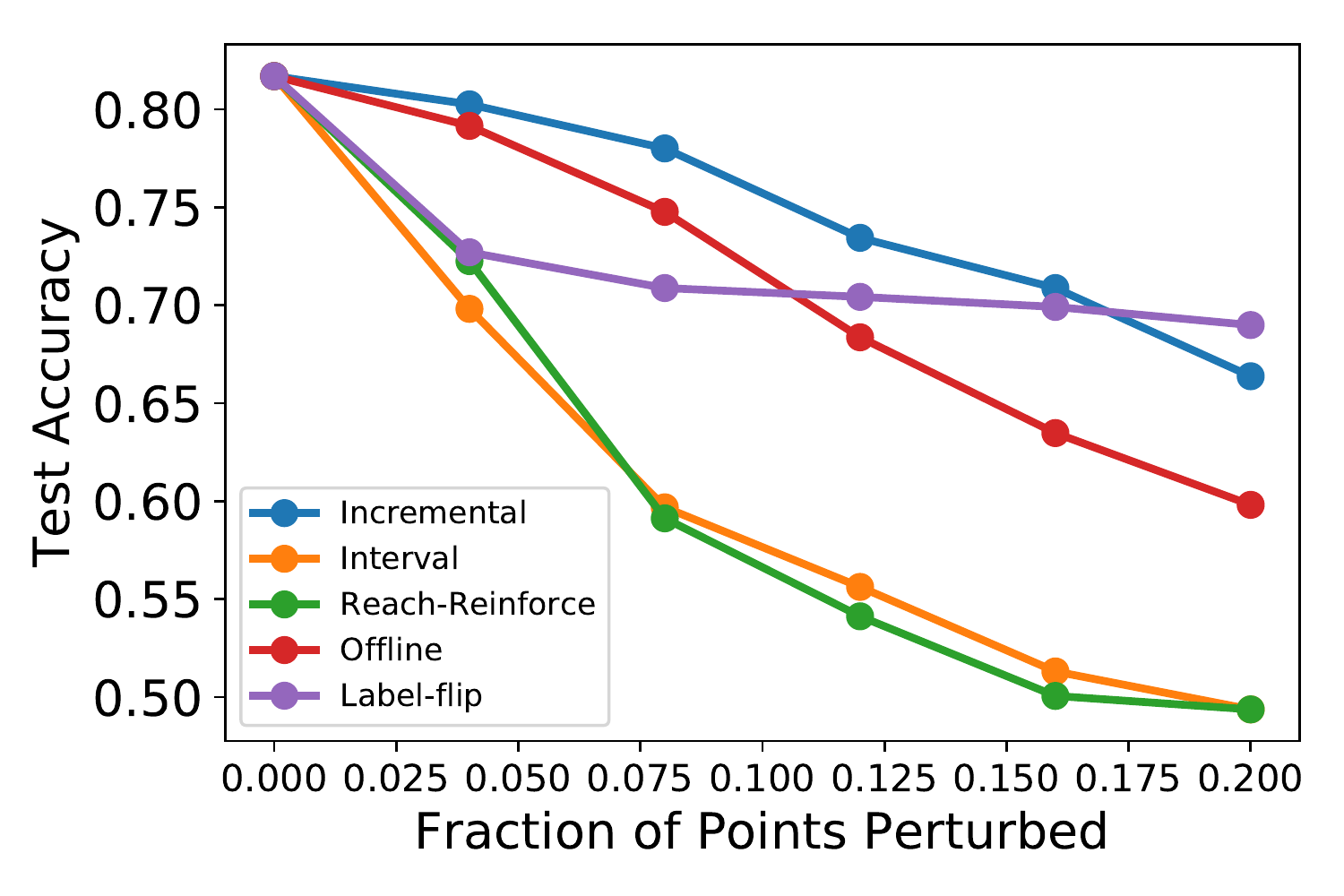}}\\ 
    \resizebox{\columnwidth}{!}{%
    \includegraphics{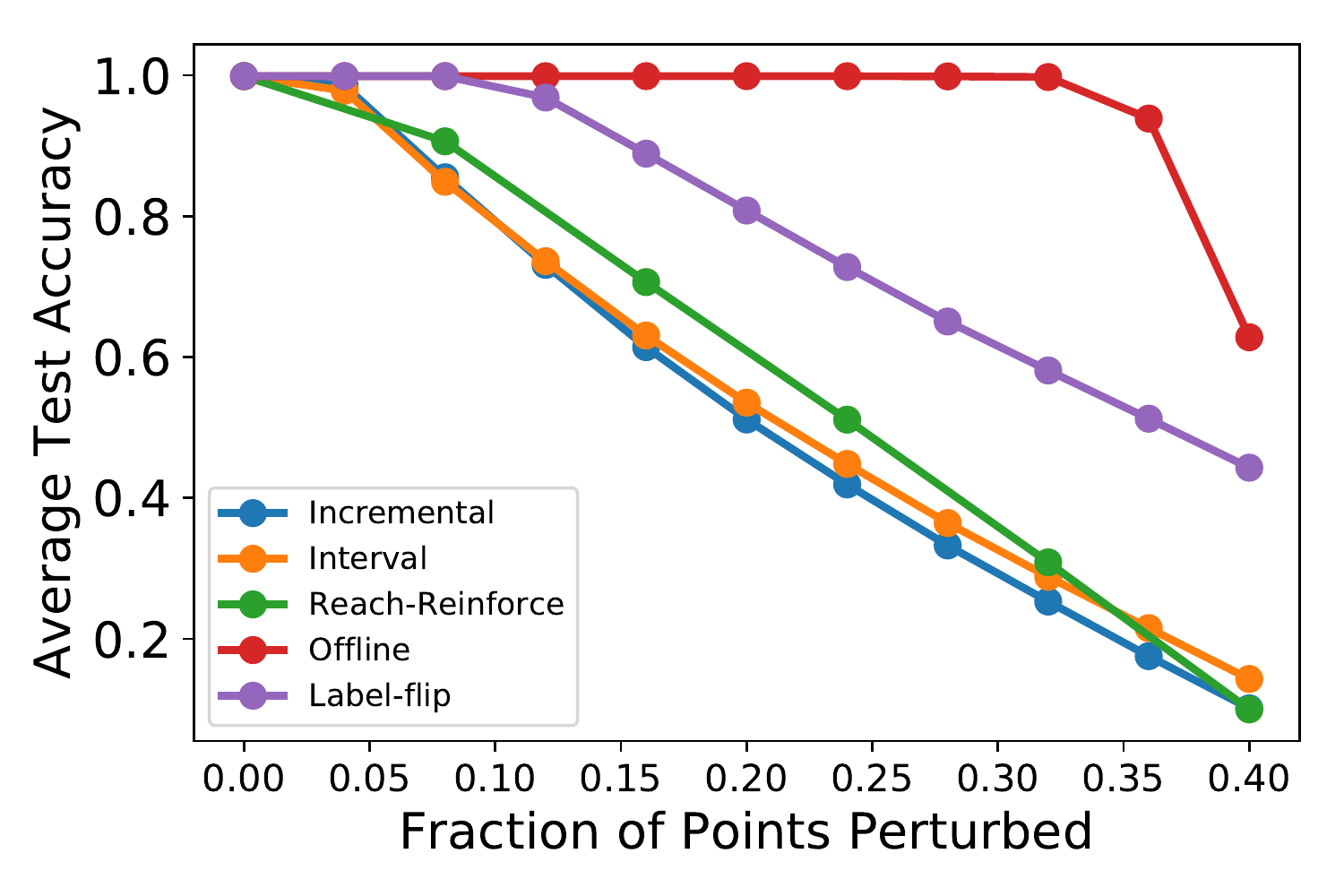}
    \includegraphics{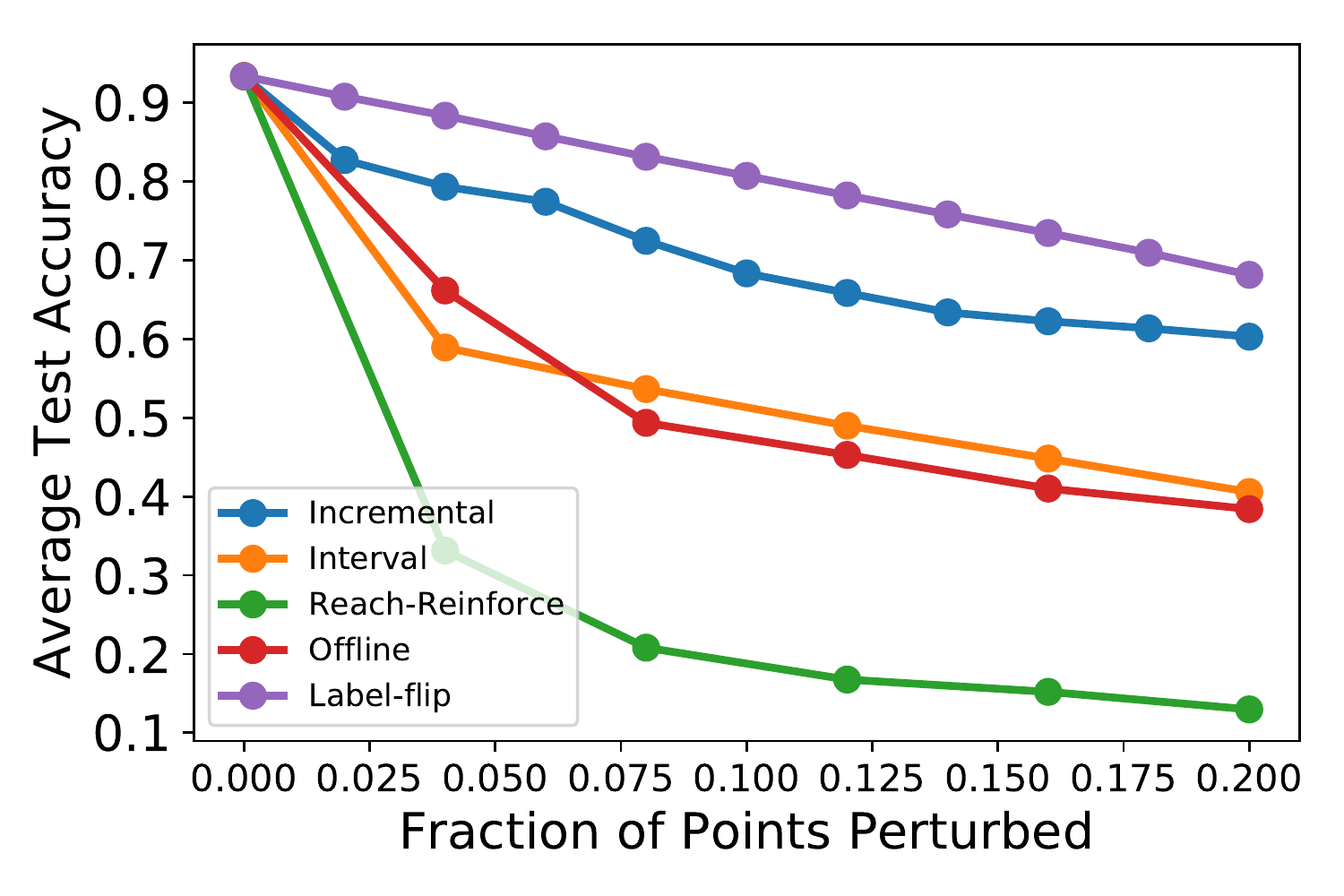}
    \includegraphics{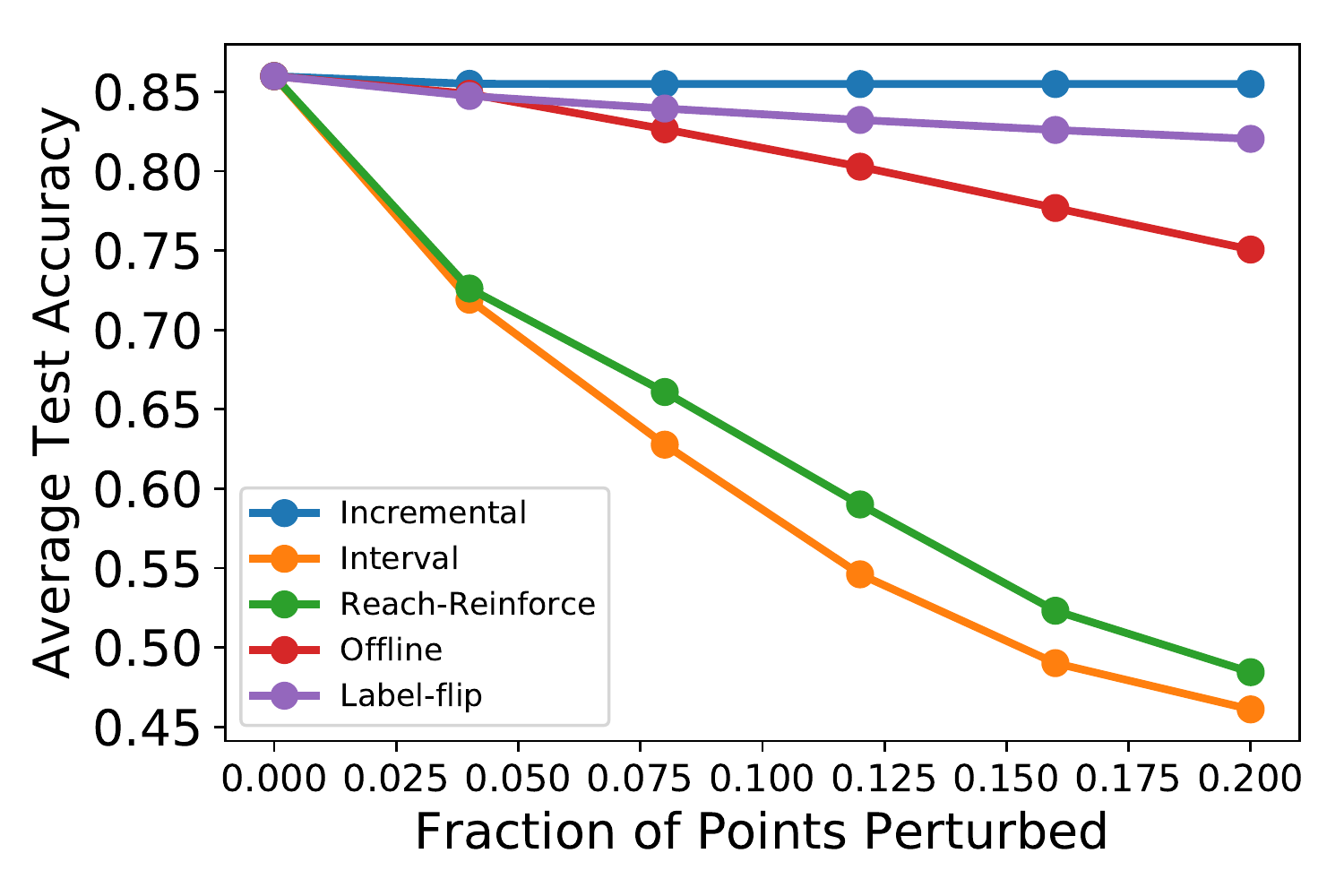}}
    \caption{Test accuracy vs. fraction of modified points for each attack on an online learner with Slow Decay learning rate. \textbf{Top row:} Semi-online, \textbf{Bottom row:} Fully-online. \textbf{Left to right:} 2D Gaussian mixture, MNIST 1 v.s. 7, UCI Spambase.}
    \label{fig:expres}
\end{figure}

\makesavenoteenv{table}
\makesavenoteenv{tabular}
\begin{table}
\begin{center}
\resizebox{\columnwidth}{!}{%
\begin{tabular}{ |c|c|c|c|c|c|c|c|c| } 
\hline
\multirow{2}{4em}{} & \multicolumn{2}{| c |}{Incremental} & \multicolumn{2}{| c |}{Interval} & \multicolumn{2}{| c |}{Teach-and-Reinforce} \\
\hline
& Semi-online & Fully-online & Semi-online & Fully-online & Semi-online & Fully-online\\
\hline 
Fast Decay & Start & Start & Start & Start & 0.75 & 0.75 \\ 
Constant & End & Start/Uniform \footnote{\label{note2} The attack positions concentrate at the beginning for 2D/Spamset, while they are more spreaded for MNIST/fashion MNIST.} & End & Start & 0 & 0.25/0.75 \footnote{\label{note1} The first value is the most frequent $\alpha$ for MNIST and fashion-MNIST; the second for 2-D Gaussian mixtures and UCI Spambase.} \\ 
Slow Decay & End & Start/Uniform\footnotemark[\getrefnumber{note2}] & End & Start & 0 & 0/0.75 \footnotemark[\getrefnumber{note1}] \\ 
\hline
\end{tabular}
}
\caption{Attack positions chosen by each online attack against an online learner in different settings. We use `Start', `End' and `Uniform' to qualitatively represent whether most attack positions are at the beginning, end or uniformly distributed over the stream. The most frequent $\alpha$ is reported for Teach-and-Reinforce.}
\label{tab:positions}
\end{center}
\end{table}

\section{Conclusion}

We initiate the study of data poisoning attacks in the context of online learning. We formalize the problem to abstract out two settings of interest -- semi-online and fully-online. In both cases, we formulate the attacker's strategy as an optimization problem and propose solution strategies. Our experiments show that our attacks perform significantly better than an attacker who is oblivious to the online nature of the input data. 

There are many avenues for further investigation -- such as, extending our attacks to more complex classifiers such as neural networks, and building online defenses. Finally, online learning is closely related to other problems on learning from feedback, such as contextual bandits. We believe that a very interesting open question is to expand our understanding to better understand the role and capabilities of adversaries in these systems.

\medskip
\bibliographystyle{plain}
\bibliography{online}

\begin{thebibliography}{10}

\bibitem{alfeld2016data}
Scott Alfeld, Xiaojin Zhu, and Paul Barford.
\newblock Data poisoning attacks against autoregressive models.
\newblock In {\em AAAI}, pages 1452--1458, 2016.

\bibitem{biggio2012poisoning}
Battista Biggio, Blaine Nelson, and Pavel Laskov.
\newblock Poisoning attacks against support vector machines.
\newblock {\em arXiv preprint arXiv:1206.6389}, 2012.

\bibitem{biggio2013data}
Battista Biggio, Ignazio Pillai, Samuel Rota~Bul{\`o}, Davide Ariu, Marcello
  Pelillo, and Fabio Roli.
\newblock Is data clustering in adversarial settings secure?
\newblock In {\em Proceedings of the 2013 ACM workshop on Artificial
  intelligence and security}, pages 87--98. ACM, 2013.

\bibitem{burkard2017analysis}
Cody Burkard and Brent Lagesse.
\newblock Analysis of causative attacks against svms learning from data
  streams.
\newblock In {\em Proceedings of the 3rd ACM on International Workshop on
  Security And Privacy Analytics}, pages 31--36. ACM, 2017.

\bibitem{cesabianchibook}
Nicolo Cesa-Bianchi and G{\'a}bor Lugosi.
\newblock {\em Prediction, learning, and games}.
\newblock Cambridge university press, 2006.

\bibitem{huang2017adversarial}
Sandy Huang, Nicolas Papernot, Ian Goodfellow, Yan Duan, and Pieter Abbeel.
\newblock Adversarial attacks on neural network policies.
\newblock {\em arXiv preprint arXiv:1702.02284}, 2017.

\bibitem{koh2017understanding}
Pang~Wei Koh and Percy Liang.
\newblock Understanding black-box predictions via influence functions.
\newblock {\em arXiv preprint arXiv:1703.04730}, 2017.

\bibitem{li2016data}
Bo~Li, Yining Wang, Aarti Singh, and Yevgeniy Vorobeychik.
\newblock Data poisoning attacks on factorization-based collaborative
  filtering.
\newblock In {\em Advances in neural information processing systems}, pages
  1885--1893, 2016.

\bibitem{lin2017tactics}
Yen-Chen Lin, Zhang-Wei Hong, Yuan-Hong Liao, Meng-Li Shih, Ming-Yu Liu, and
  Min Sun.
\newblock Tactics of adversarial attack on deep reinforcement learning agents.
\newblock {\em arXiv preprint arXiv:1703.06748}, 2017.

\bibitem{LW88}
Nicholas Littlestone, Philip~M Long, and Manfred~K Warmuth.
\newblock On-line learning of linear functions.
\newblock In {\em Proceedings of the twenty-third annual ACM symposium on
  Theory of computing}, pages 465--475. ACM, 1991.

\bibitem{meizhu}
Shike Mei and Xiaojin Zhu.
\newblock Using machine teaching to identify optimal training-set attacks on
  machine learners.
\newblock In {\em AAAI}, pages 2871--2877, 2015.

\bibitem{munoz2017towards}
Luis Mu{\~n}oz-Gonz{\'a}lez, Battista Biggio, Ambra Demontis, Andrea Paudice,
  Vasin Wongrassamee, Emil~C Lupu, and Fabio Roli.
\newblock Towards poisoning of deep learning algorithms with back-gradient
  optimization.
\newblock In {\em Proceedings of the 10th ACM Workshop on Artificial
  Intelligence and Security}, pages 27--38. ACM, 2017.

\bibitem{nelson2008exploiting}
Blaine Nelson, Marco Barreno, Fuching~Jack Chi, Anthony~D Joseph, Benjamin~IP
  Rubinstein, Udam Saini, Charles~A Sutton, J~Doug Tygar, and Kai Xia.
\newblock Exploiting machine learning to subvert your spam filter.
\newblock {\em LEET}, 8:1--9, 2008.

\bibitem{newell2014practicality}
Andrew Newell, Rahul Potharaju, Luojie Xiang, and Cristina Nita-Rotaru.
\newblock On the practicality of integrity attacks on document-level sentiment
  analysis.
\newblock In {\em Proceedings of the 2014 Workshop on Artificial Intelligent
  and Security Workshop}, pages 83--93. ACM, 2014.

\bibitem{newsome2006paragraph}
James Newsome, Brad Karp, and Dawn Song.
\newblock Paragraph: Thwarting signature learning by training maliciously.
\newblock In {\em International Workshop on Recent Advances in Intrusion
  Detection}, pages 81--105. Springer, 2006.

\bibitem{membershipinference2016}
Reza Shokri, Marco Stronati, Congzheng Song, and Vitaly Shmatikov.
\newblock Membership inference attacks against machine learning models.
\newblock In {\em Security and Privacy (SP), 2017 IEEE Symposium on}, pages
  3--18. IEEE, 2017.

\bibitem{steinhardt2017certified}
Jacob Steinhardt, Pang Wei~W Koh, and Percy~S Liang.
\newblock Certified defenses for data poisoning attacks.
\newblock In {\em Advances in Neural Information Processing Systems}, pages
  3520--3532, 2017.

\bibitem{modelsteal2016}
Florian Tram{\`e}r, Fan Zhang, Ari Juels, Michael~K Reiter, and Thomas
  Ristenpart.
\newblock Stealing machine learning models via prediction apis.
\newblock In {\em USENIX Security Symposium}, pages 601--618, 2016.

\bibitem{xiao2012adversarial}
Han Xiao, Huang Xiao, and Claudia Eckert.
\newblock Adversarial label flips attack on support vector machines.
\newblock In {\em ECAI}, pages 870--875, 2012.

\bibitem{xiao2015feature}
Huang Xiao, Battista Biggio, Gavin Brown, Giorgio Fumera, Claudia Eckert, and
  Fabio Roli.
\newblock Is feature selection secure against training data poisoning?
\newblock In {\em International Conference on Machine Learning}, pages
  1689--1698, 2015.

\bibitem{xiao2015support}
Huang Xiao, Battista Biggio, Blaine Nelson, Han Xiao, Claudia Eckert, and Fabio
  Roli.
\newblock Support vector machines under adversarial label contamination.
\newblock {\em Neurocomputing}, 160:53--62, 2015.

\bibitem{zinkevich2003online}
Martin Zinkevich.
\newblock Online convex programming and generalized infinitesimal gradient
  ascent.
\newblock In {\em Proceedings of the 20th International Conference on Machine
  Learning (ICML-03)}, pages 928--936, 2003.

\end{thebibliography}

\newpage
\appendix
\section{Derivation and Description of Attack Algorithms}
\subsection{Gradient Computation Using an Recurrent Prefix}
\label{sec:prefix}
Equation~\ref{eq:gradgeneral} shows the formula of calculating the exact online attack gradient for a general smooth function $F(w_t)$. In this section, we show how this gradient can be efficiently w.r.t. all points in the data stream using a recurrent prefix. 

The computation for the case of $i\geq t$ and $i=t-1$ in Equation~\ref{eq:gradgeneral} is straightforward once $F(w_t)$ is known, so we focus at computing the gradient for all $i < t-1$. We call $\xi_i$ the prefix of $\pd F(w_t)/\pd x_i$, where
\begin{equation}
\xi_i = \frac{\pd F(w_t)}{\pd w_t}
\frac{\pd w_{t}}{\pd w_{t-1}}\cdots \frac{\pd w_{i+2}}{\pd w_{i+1}}
\end{equation}
and for all $i<t-1$, we have 
\begin{equation}
\frac{\pd F(w_t)}{\pd x_i} = \xi_i \frac{\pd w_{i+1}}{\pd x_i}.
\end{equation}
Notice that the prefix has a simple recurrence relation
\begin{equation}
\xi_{i} = \xi_{i+1}\frac{\pd w_{i+2}}{\pd w_{i+1}},
\end{equation}
so it is more efficient to first find the prefix $\xi_i$ for all $i < t-1$ and then find $\pd F(w_t)/\pd x_i$ from $\xi_i$ than to compute each $\pd F(w_t)/\pd x_i$ from scratch. Notice that $\pd F(w_t)/\pd w_t$ and each $\xi_i$ is a vector of dimension $d$. Finding $\frac{\pd w_{i+2}}{\pd w_{i+1}}$ and $\frac{\pd w_{i+1}}{\pd x_i}$ for a particular $i$ takes $O(d^2)$ time each, and computing $\xi_i$ for all $i$ involves making $O(T)$ vector-matrix multiplications, each taking $O(d^2)$ time. Therefore the computation complexity of finding all gradients is $O(Td^2)$. 

\subsection{Gradient Computation for Online Logistic Regression}
\label{sec:gradlr}
We now instantiate the gradient computation with the objective function for online logistic regression using OGD. Recall that the attacker's objective $L(w_t)$ at time $t$ is the negative logistic loss of the classifier with weight $w_t$ over a label-flipped validation set $S_{validinv}$, i.e.
\begin{equation}
L(w_t) = \sum_{(x,y) \in S_{validinv}} -\log(1+\exp(-yw_t^Tx)).
\end{equation}
The gradient calculation procedure in Equation~\ref{eq:gradgeneral} requires $\frac{\pd L(w_t)}{\pd w_t}$, $\frac{\pd w_{s+1}}{\pd w_s}$ for all $1 \leq s < t$ and $\frac{\pd w_{i+1}}{\pd x_i}$ for all $0 \leq i < t$, which can be found as follows:
\begin{equation}
\frac{\pd L(w_t)}{\pd w_t} = \sum_{(x,y)\in S_{validinv}}\frac{yx}{1+\exp(yw_t^Tx)}, 
\end{equation}
\begin{equation}
\begin{split}
    \frac{\pd w_{s+1}}{\pd w_s} =\ (1-2\lambda\eta_s)\identity - \eta_s\left(\frac{\exp(y_sw_s^Tx_s)}{[1+\exp(y_sw_s^Tx_s)]^2}\right)x_sx_s^T,
\end{split}
\end{equation}
and
\begin{equation}
\begin{split}
    \frac{\pd w_{i+1}}{\pd x_i} =\  -\eta_i\left(
    \frac{-y_i\identity}{1+\exp(y_iw_i^Tx_i)}
    + \frac{\exp(y_iw_i^Tx_i)}{[1+\exp(y_iw_i^Tx_i)]^2}x_iw_i^T
    \right),
\end{split}
\end{equation}
where $\lambda$ is the regularization constant of OGD and $\eta_i$ is the learning rate of OGD at time step $i$.

\subsection{Attack Algorithms Descptions and Pseudocodes.}
In order to simplify the algorithm pseudocode, we introduce the following notations. 

First, we denote the attacker's objective function over a training stream $S$ as $L(S)$. Recall that in Section~{\ref{sec:gradlr}}, we use $L(w_t)$ to represent the attacker's objective function at time $t$, where $w_t$ is the model parameter of the online logistic regression classifier at time $t$ trained on stream $S$. Then we have $L(S) = L(w_T)$ for the semi-online setting and $L(S) = \sum_{t=1}^T L(w_t)$ for the fully-online setting. 

Second, we use $S_{a:b}$ to represent the substream of $S$ from $S_a$ to $S_{b-1}$, and use $S_{a:b:i}$ to represent the substream containing every $i$-th point in $S$ from $S_a$ to $S_{b-1}$. If a substream can be represented by concatenating two above mentioned types of substreams, then its subscript will be a list of its component substream's subscript separated by comma. For example, $S_{a:b, b:c:i}$ means a substream with points from $S_a$ to $S_{b-1}$ and then every $i$-th point from $S_b$ to $S_{c-1}$. 

Lastly, we use $\epsilon_0$ to denote the initial step size of gradient ascent and $\epsilon_{n_{iter}}$ to denote the gradient ascent step size at step $n_{iter}$. The step size $\epsilon_{n_{iter}}$ is a function of $\epsilon_0$ and $n_{iter}$, which decays at the rate of $O(1/\sqrt{n_{iter}})$.
\begin{algorithm}
\caption{Incremental Attack($S$, $K$, $\epsilon_0$, $max_{iter}$)}
\label{algo:incremental}
\begin{algorithmic}
\STATE{\textbf{S: }the original training stream, \textbf{$\mathbf{\epsilon_0}$:} initial gradient ascent step size, \textbf{$\mathbf{K}$:} the attacker's budget, $\mathbf{max_{iter}:}$ maximum number of iterations for gradient ascent.}
\STATE{$n_{iter} = 0$}
\STATE{$T = |S|$}
\STATE{$hasPerturbed = [0]\times T$}
\STATE{$n_{perturbed} = 0$}
\WHILE{($n_{iter} < max_{iter}$) and ($n_{perturbed} <= K$ )}
\STATE{$x_i = \arg\max_{x_k \in S} ||\frac{\pd L(S)}{\pd x_k}||_2$}
\STATE{$x_i = \mbox{proj}_{\calF}(x_i + \epsilon_{n_{iter}} \frac{\pd L(S)}{\pd x_i})$}
\STATE{$hasPerturbed[i] = 1$}
\STATE{$n_{iter} = n_{iter} + 1$}
\STATE{$n_{perturbed} = sum(hasPerturbed)$}
\ENDWHILE
\RETURN{$S$}
\end{algorithmic}
\end{algorithm}

\begin{algorithm}
\caption{Interval Attack($S$, $K$, $\epsilon_0$, $max_{iter}$, $s$)}
\label{algo:interval}
\begin{algorithmic}
\STATE{\textbf{S: }the original training stream, \textbf{$\mathbf{\epsilon_0}$:} initial gradient ascent step size, \textbf{$\mathbf{K}$:} the attacker's budget, which is also the length of the sliding window, $\mathbf{max_{iter}:}$ maximum number of iterations for gradient ascent, \textbf{$\mathbf{s}$:} the step size for grid searching the best attack window.}
\STATE{$T = |S|$}
\STATE{$S_{att} = S$}
\FOR{\mbox{$t \in$ range $(0, T-l, s)$}}
\STATE{$S' = S$}
\STATE{$S'_{t:t+K}$ = Find-Best-Modification-Over-Substream($S'$, $\epsilon_0$, $max_{iter}$, $S'_{t:t+K}$)}
\IF{$L(S') > L(S_{att})$}
\STATE{$S_{att} = S'$}
\ENDIF
\ENDFOR
\RETURN{$S_{att}$}
\end{algorithmic}
\end{algorithm}

\begin{algorithm}
\caption{Teach-and-Reinforce Attack($S$, $K$, $\epsilon_0$, $max_{iter}$, $\alpha$)}
\label{algo:teachreinforce}
\begin{algorithmic}
\STATE{\textbf{S: }the original training stream, \textbf{$\mathbf{\epsilon_0}$:} initial gradient ascent step size, \textbf{$\mathbf{K}$:} the attacker's budget, \textbf{max\_iter:} maximum number of iterations for gradient ascent, \textbf{$\mathbf{\alpha}$:} fraction of attack budget used for teaching.}
\STATE{$T = |S|$}
\STATE{$s = \lceil \frac{T-\alpha K}{(1-\alpha) K}\rceil$}
\STATE{$S_{0:\alpha K, \alpha K:T:s} $ = Find-Best-Modification-Over-Substream($S$, $\epsilon_0$, $max_{iter}$, $S_{0:\alpha K, \alpha K:T:s}$)}
\RETURN{$S$}
\end{algorithmic}
\end{algorithm}

\begin{algorithm}
\caption{Find-Best-Modification-Over-Substream($S$, $\epsilon_0$, $max_{iter}$, $S_{modify}$)}
\begin{algorithmic}
\STATE{\textbf{S: }the original training stream, \textbf{$\mathbf{\epsilon_0}$:} initial gradient ascent step size, \textbf{max\_iter:} maximum number of iterations for gradient ascent, \textbf{$\mathbf{S_{modify}}$:} the substream to be modified to optimize the attacker's objective.}
\STATE{$n_{iter} = 0$}
\WHILE{$n_{iter} < max_{iter}$}
\FOR {$x \in S_{modify}$}
\STATE{$x = \mbox{proj}_{\mathcal{F}}(x + \epsilon_{n_{iter}} \pd L(S) / \pd x)$}
\ENDFOR
\STATE{$n_{iter} = n_{iter} + 1$}
\ENDWHILE
\RETURN{$S_{modify}$}
\end{algorithmic}
\end{algorithm}

\section{Additional Experiment Description and Result Plots}
\subsection{Dataset Size and Initial Learning Rate of Online Learners}

For each data set, we select a training set of size $400$ which is presented to the learner as the input stream, and a separate test set which is used for evaluation. The attacker also has a held-out validation set of size $200$, and the learner has a separate held-out set of the same size as the training set, which is used to initialize the classifier $w_0$. 

For 2D-Gaussian mixtures, MNIST 1 v.s. 7 and fashion MNIST sandals v.s. boots datasets, the initial learning rate of online learner is 
$\eta_0 = 0.005$ for Constant learning rate and $\eta_0 = 0.1$ for Slow Decay and Fast Decay.
For UCI Spamset, the initial learning rate is $\eta_0 = 0.01$ for Constant, $\eta_0 = 0.1$ for Slow Decay and $\eta_0 = 0.2$ for Fast Decay.
These parameter ensure that in all cases, the online learner has more than $90\%$ test accuracy when run on the clean data stream.
\subsection{Efficacy of Attack Plots}
\label{sec:resplots}
Only the result of attacks against Slow Decay online learner is presented in the main text due to the page limit. In this section, we show the plots of test accuracy against fraction of points modified for all datasets and all online learner learning rates.
\begin{center}
    \centering
    \includegraphics[scale=0.3]{./2d/semi-onlineslow-decaytotal-accwarm.pdf}
    \includegraphics[scale=0.3]{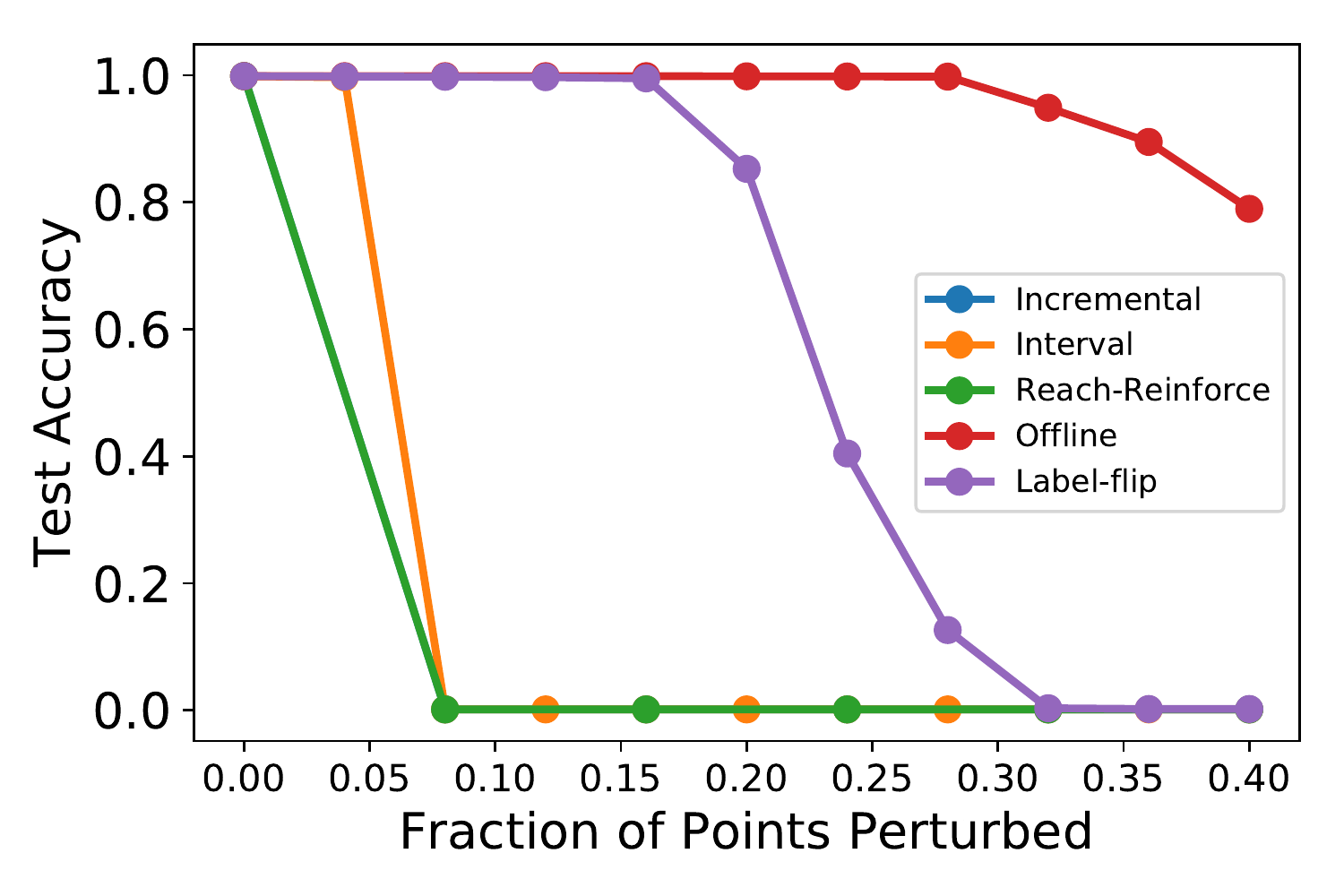}
    \includegraphics[scale=0.3]{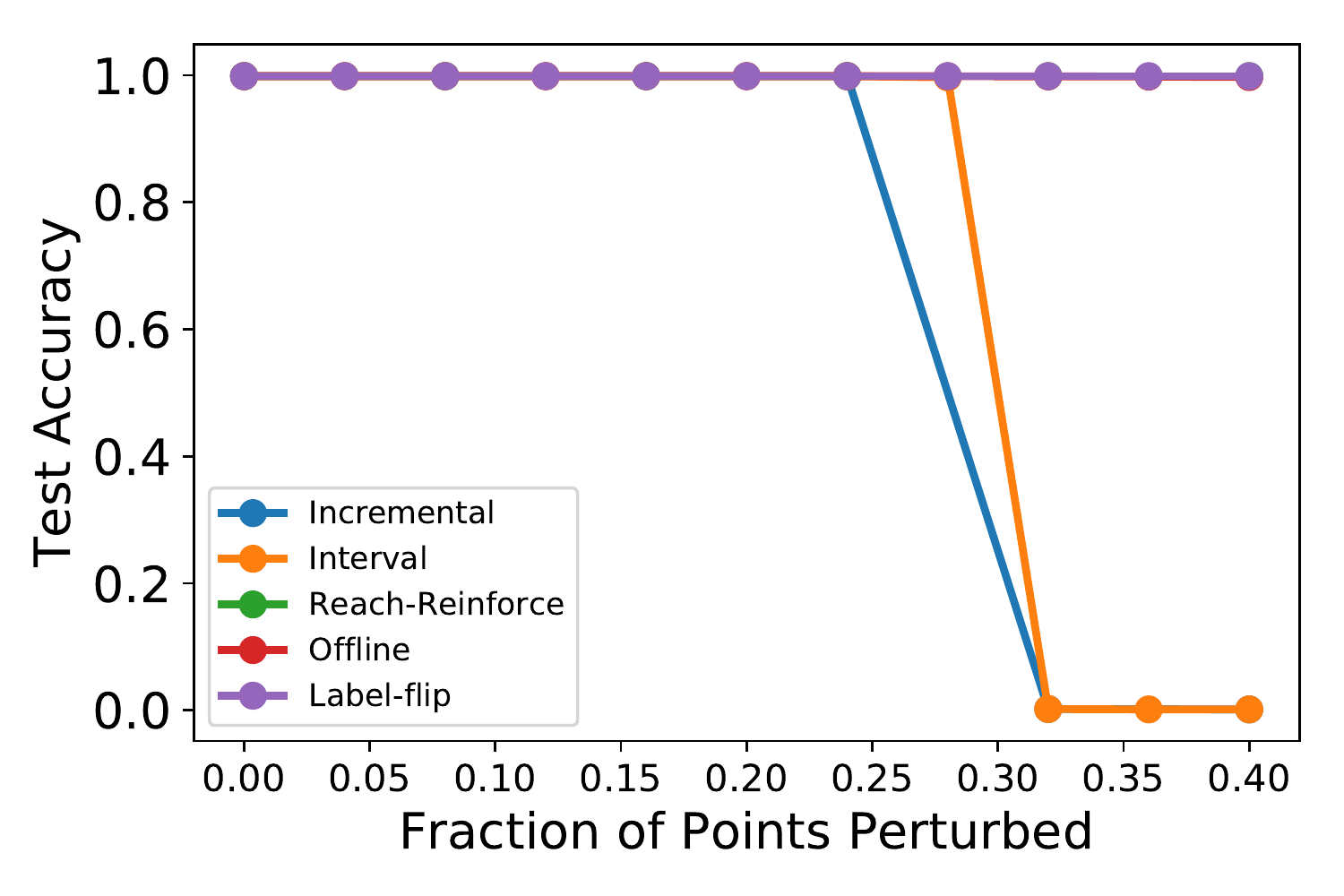}\\ 
    \includegraphics[scale=0.3]{./2d/full-onlineslow-decaytotal-accwarm.pdf}
    \includegraphics[scale=0.3]{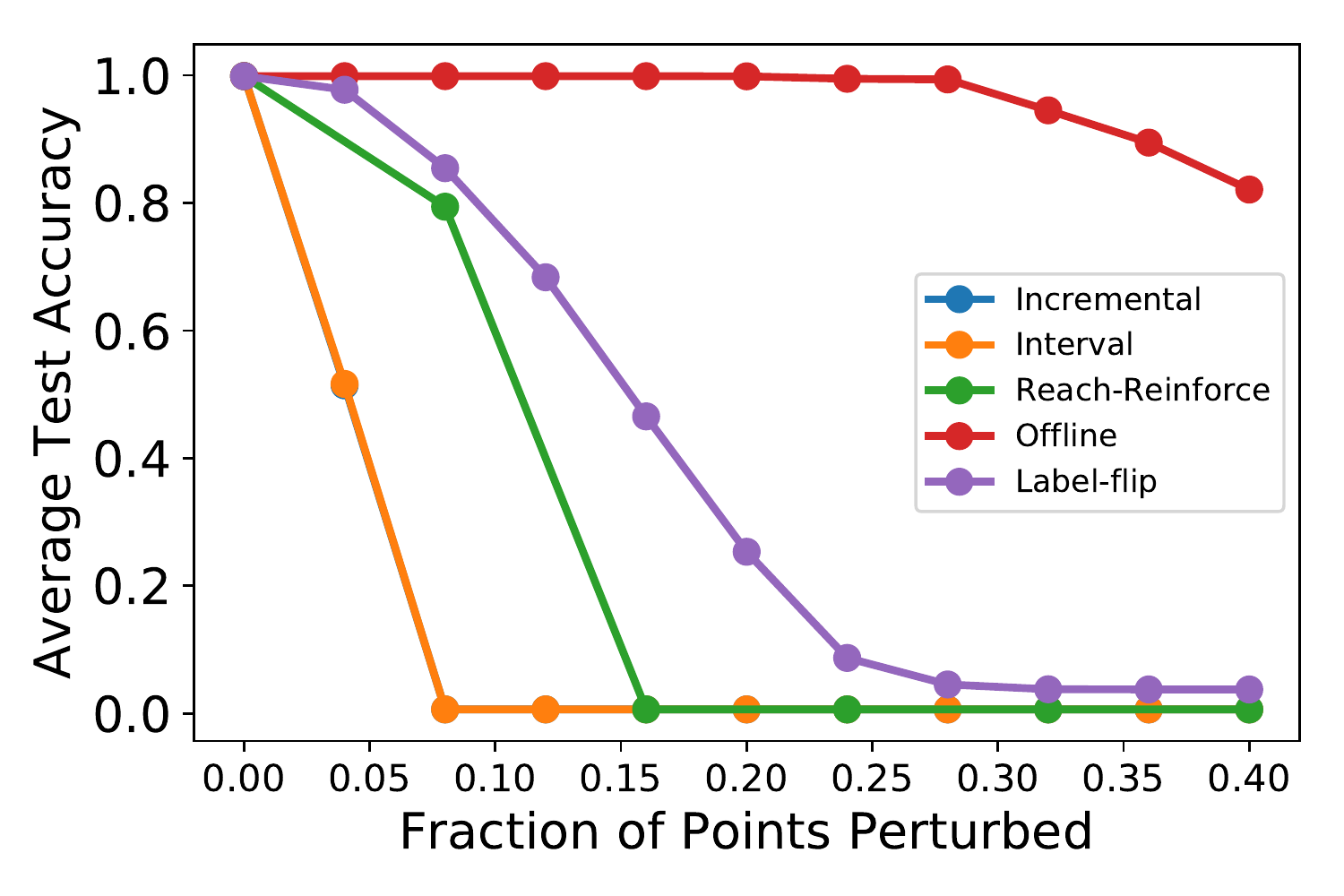}
    \includegraphics[scale=0.3]{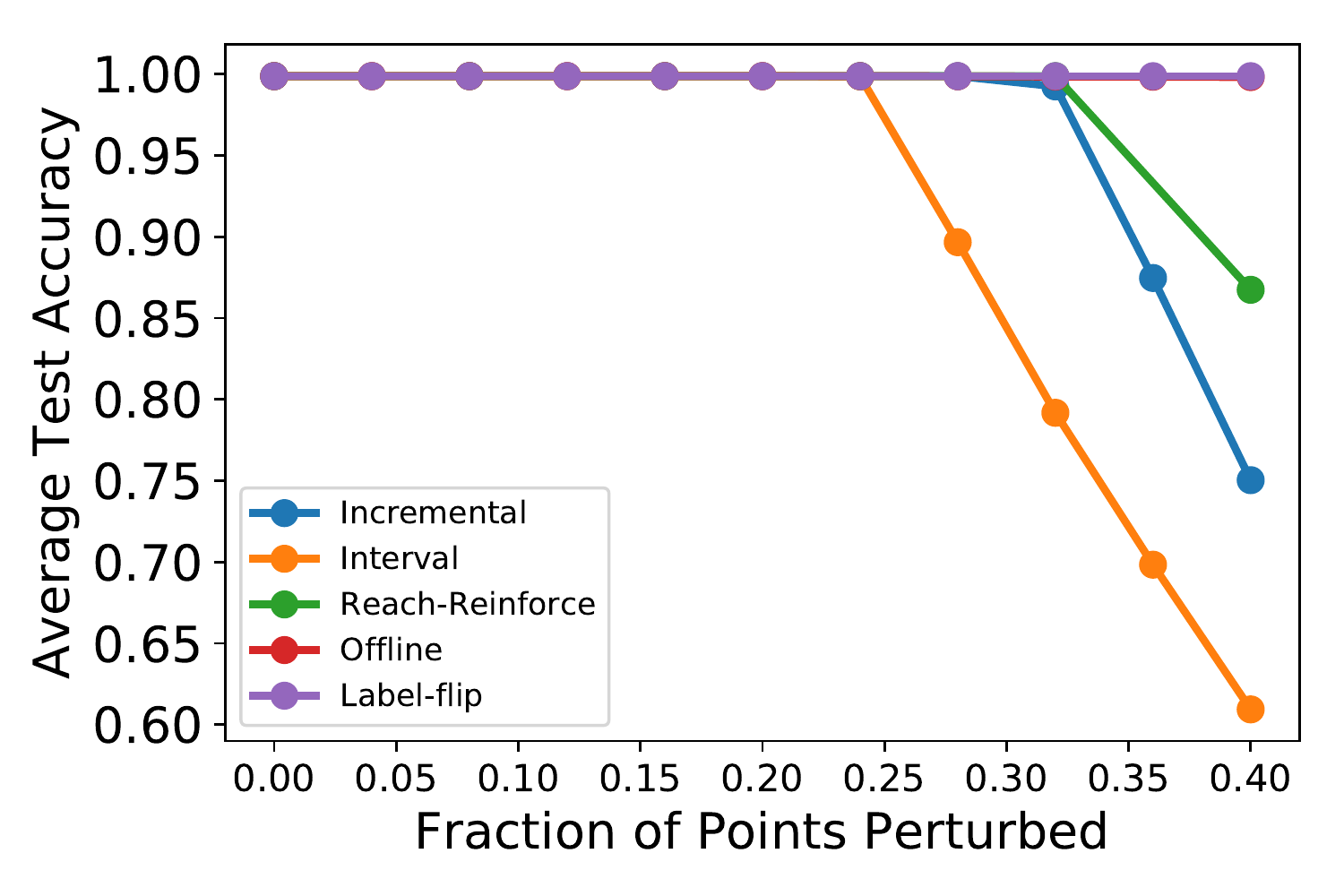}
    \captionof{figure}{Results for 2d Gaussian mixtures. \textbf{Top row:} Semi-online, \textbf{Bottom row:} Fully-online. \textbf{Left to right:} Slow Decay, Fast Decay, Constant}
    \label{fig:2d}
\end{center}

\begin{center}
    \centering
    \includegraphics[scale=0.3]{./MNIST/semi-onlineslow-decaytotal-accwarm.pdf}
    \includegraphics[scale=0.3]{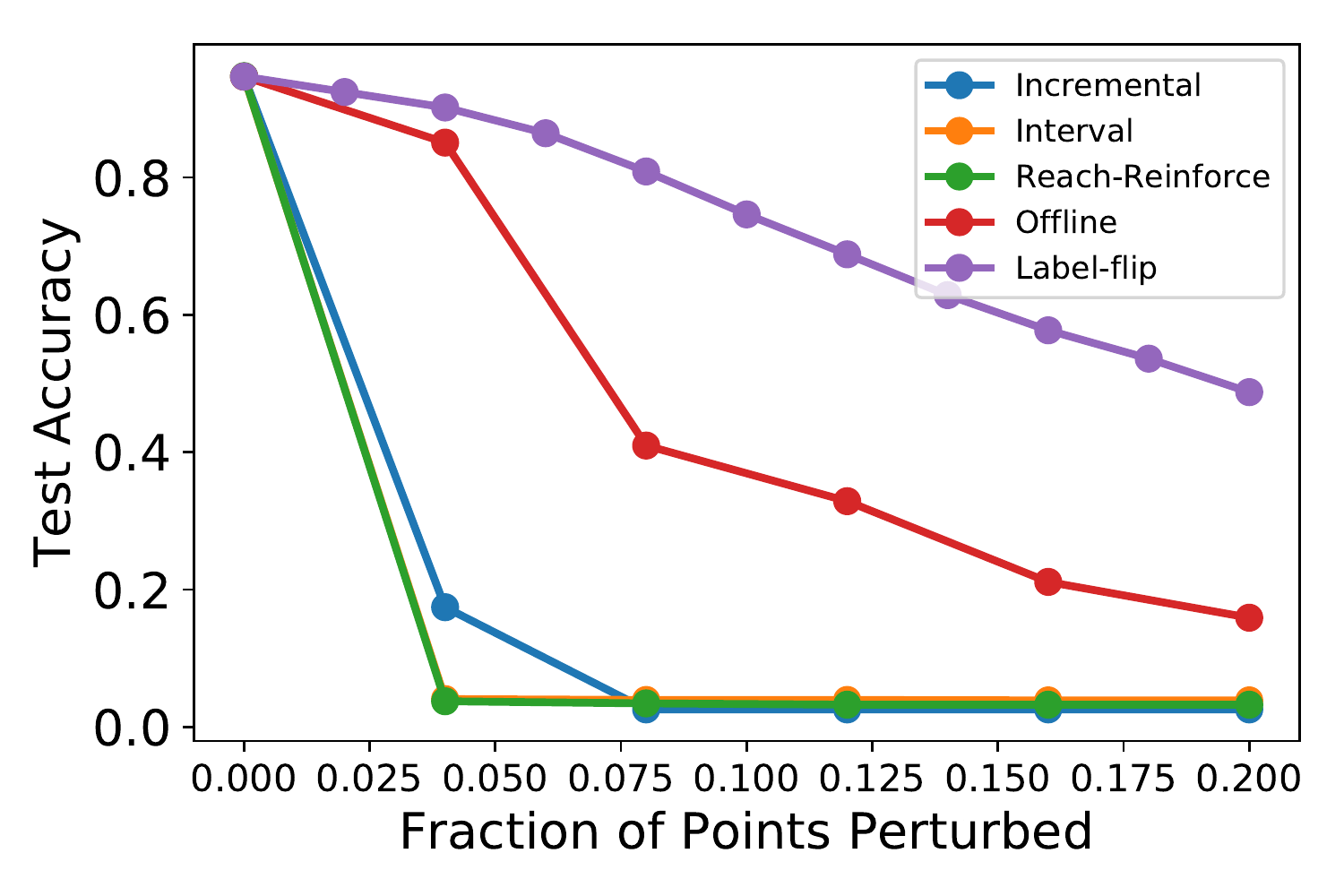}
    \includegraphics[scale=0.3]{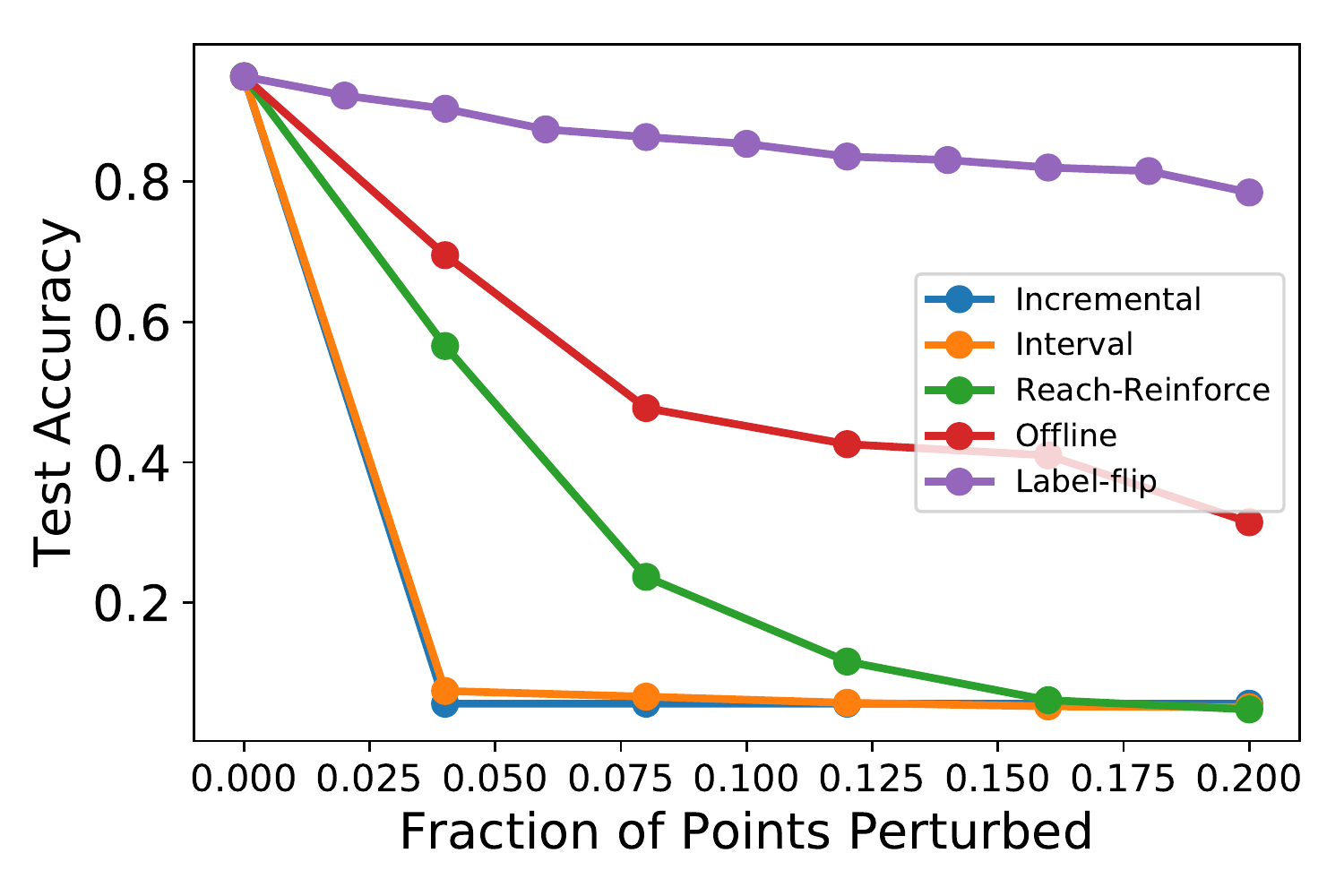}\\ 
    \includegraphics[scale=0.3]{./MNIST/full-onlineslow-decaytotal-accwarm.pdf}
    \includegraphics[scale=0.3]{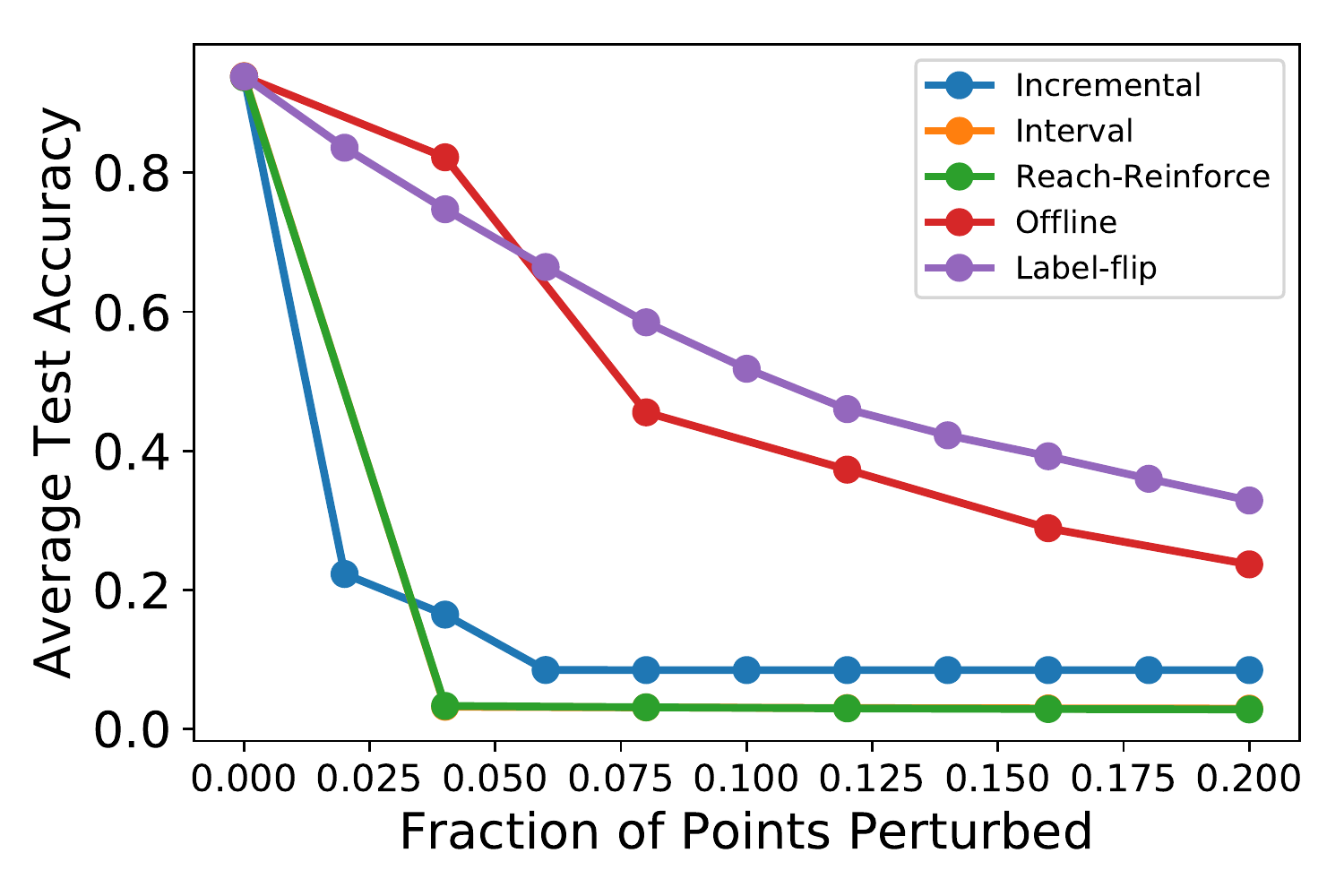}
    \includegraphics[scale=0.3]{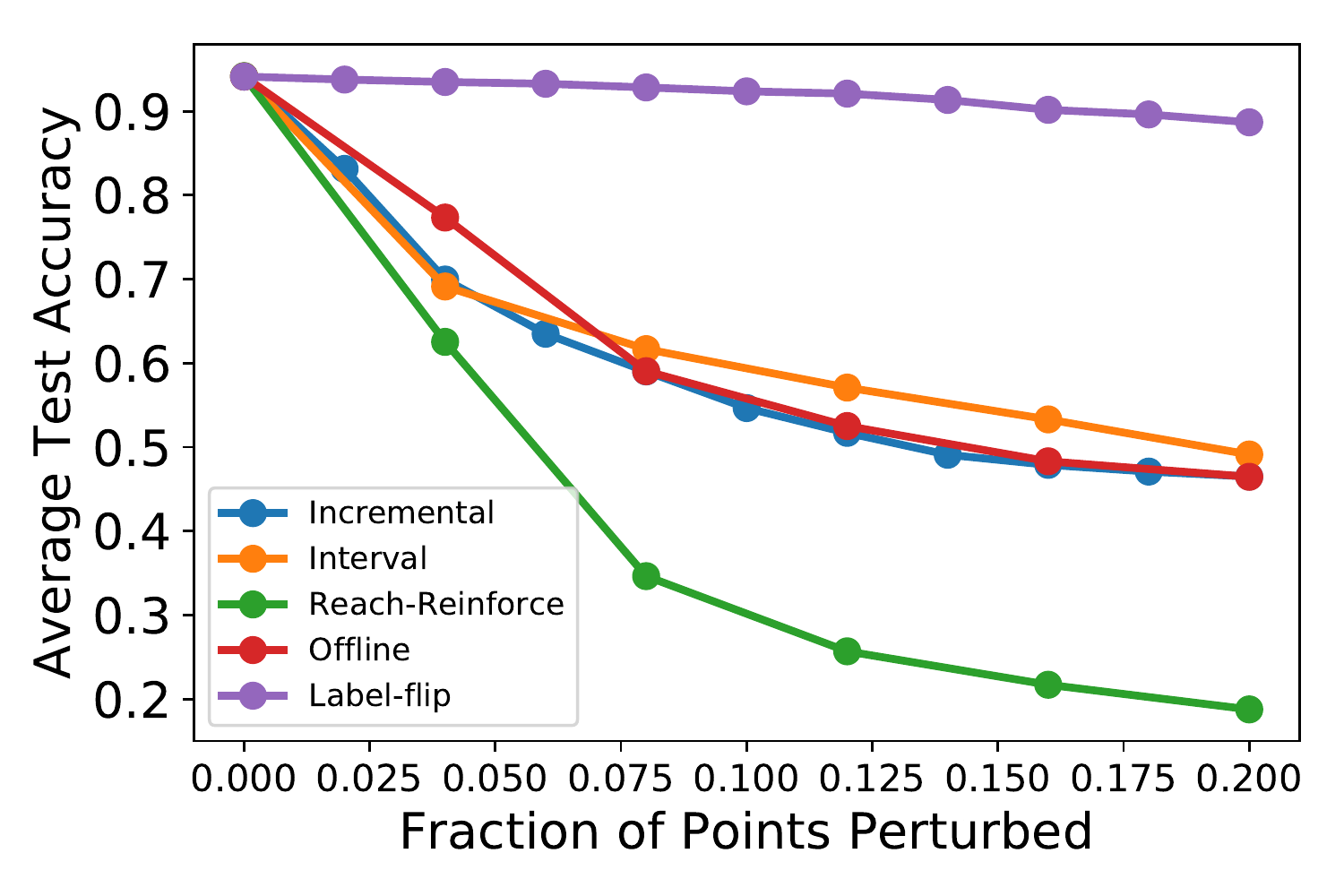}
    \captionof{figure}{Results for MNIST. \textbf{Top row:} Semi-online, \textbf{Bottom row:} Fully-online. \textbf{Left to right:} Slow Decay, Fast Decay, Constant}
    \label{fig:MNIST}
\end{center}

\begin{center}
    \centering
    \includegraphics[scale=0.3]{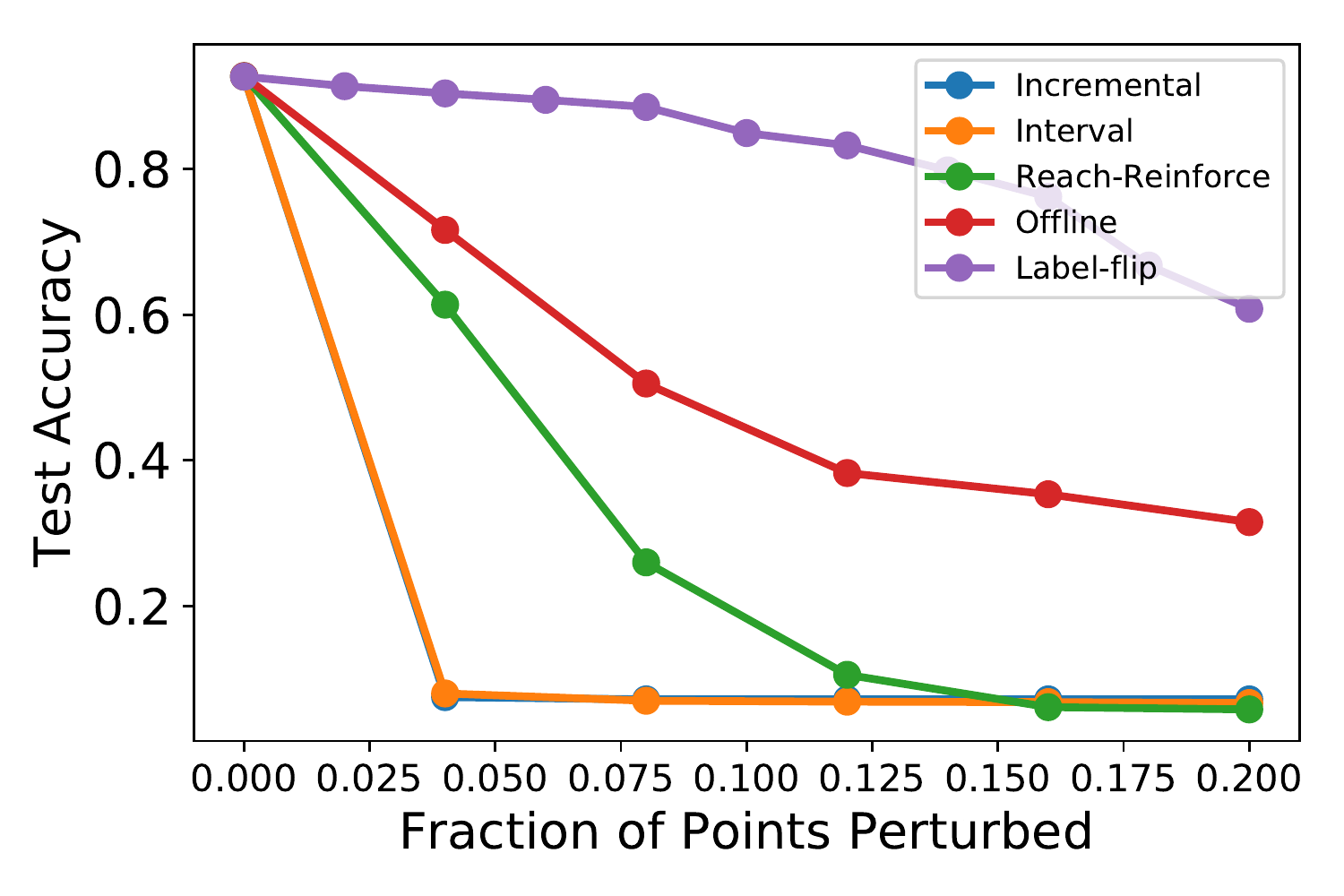}
    \includegraphics[scale=0.3]{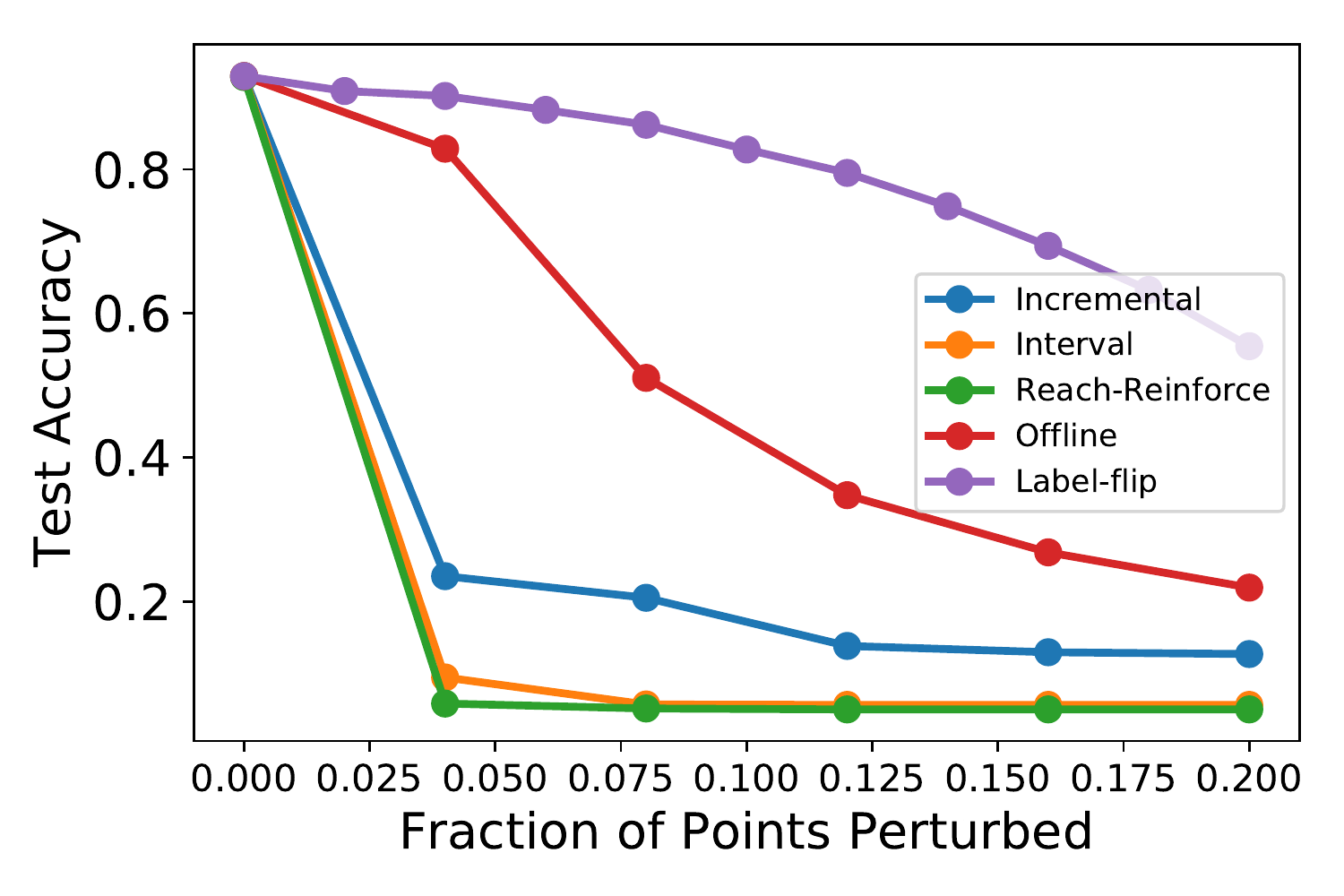} 
    \includegraphics[scale=0.3]{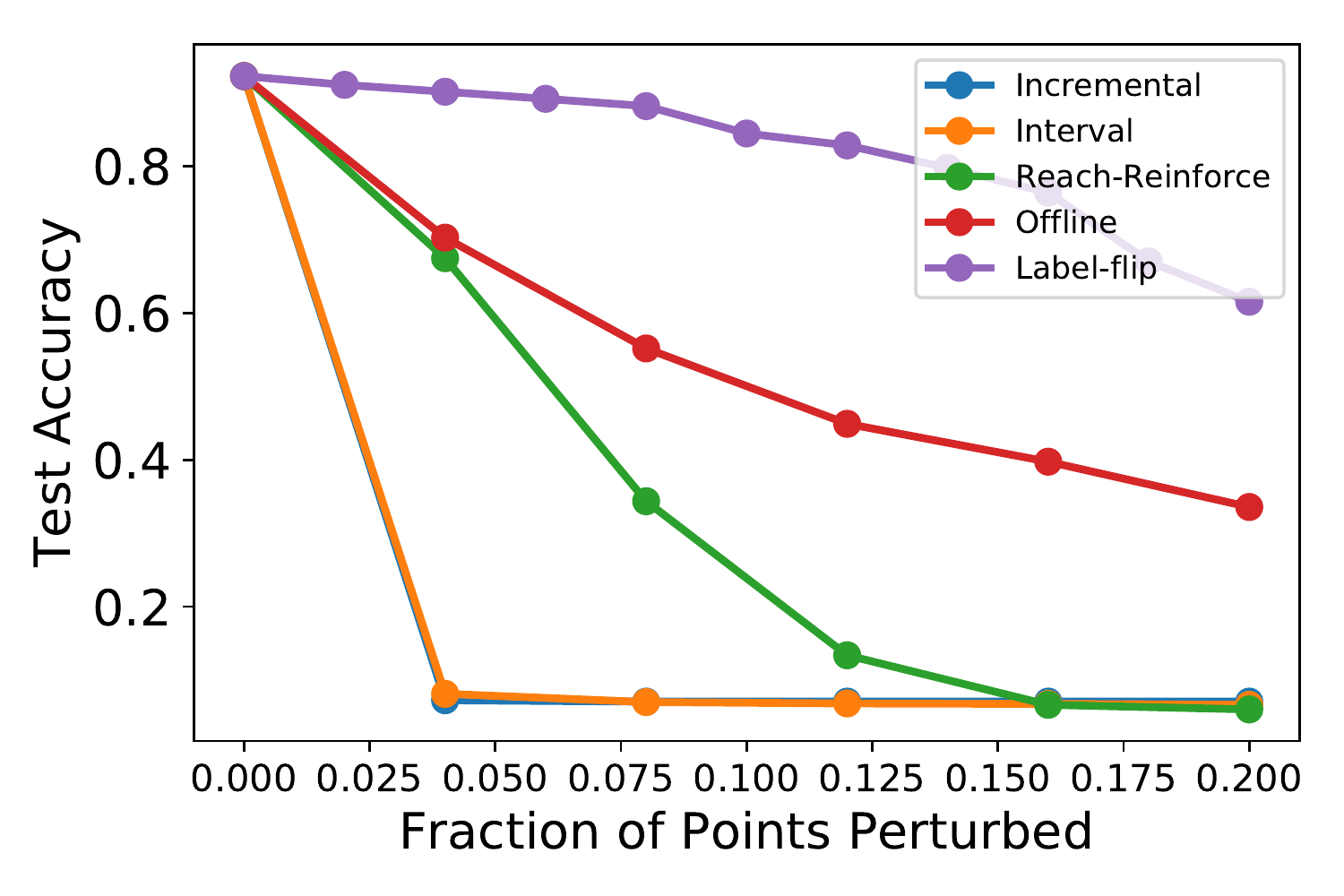}\\
    \includegraphics[scale=0.3]{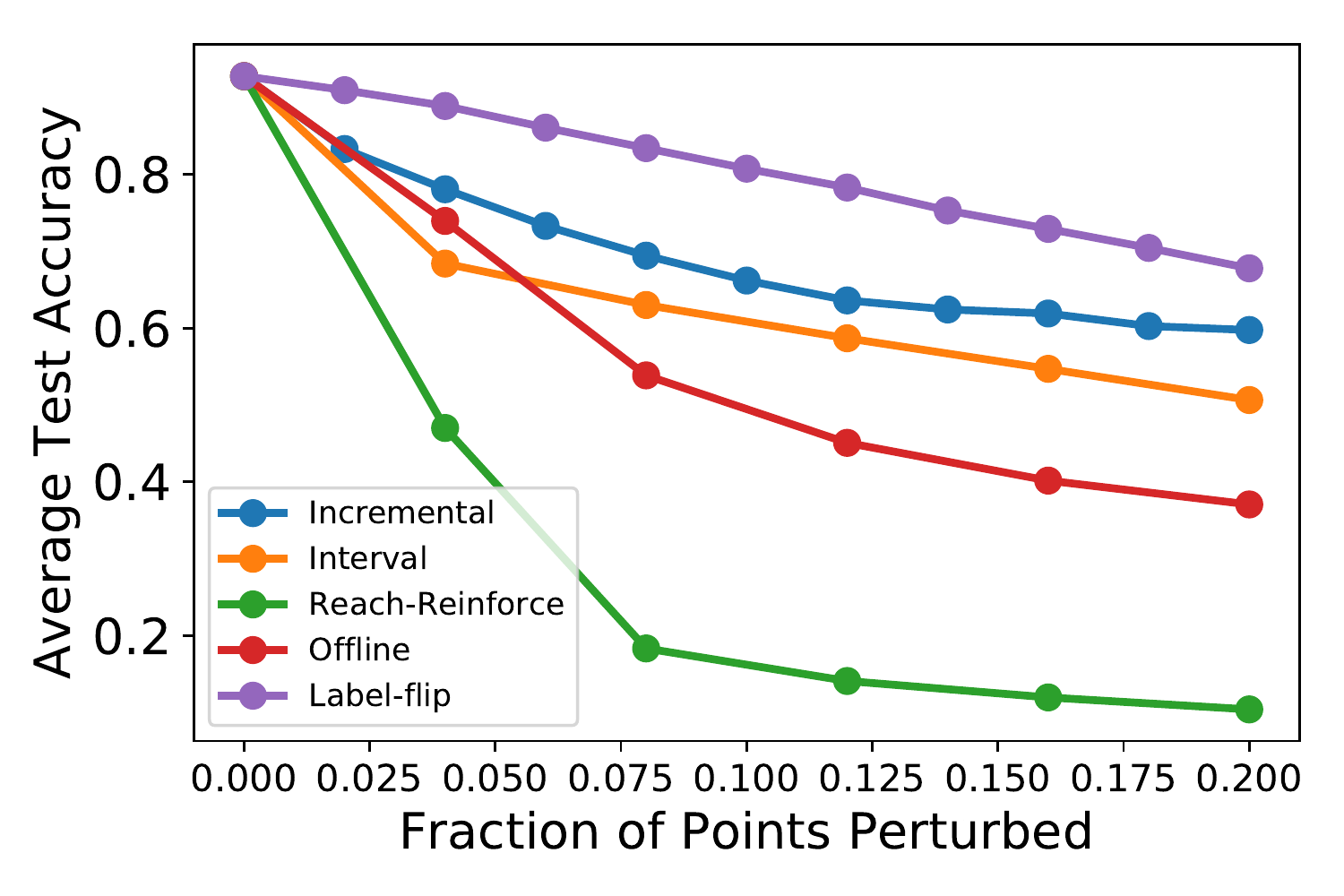}
    \includegraphics[scale=0.3]{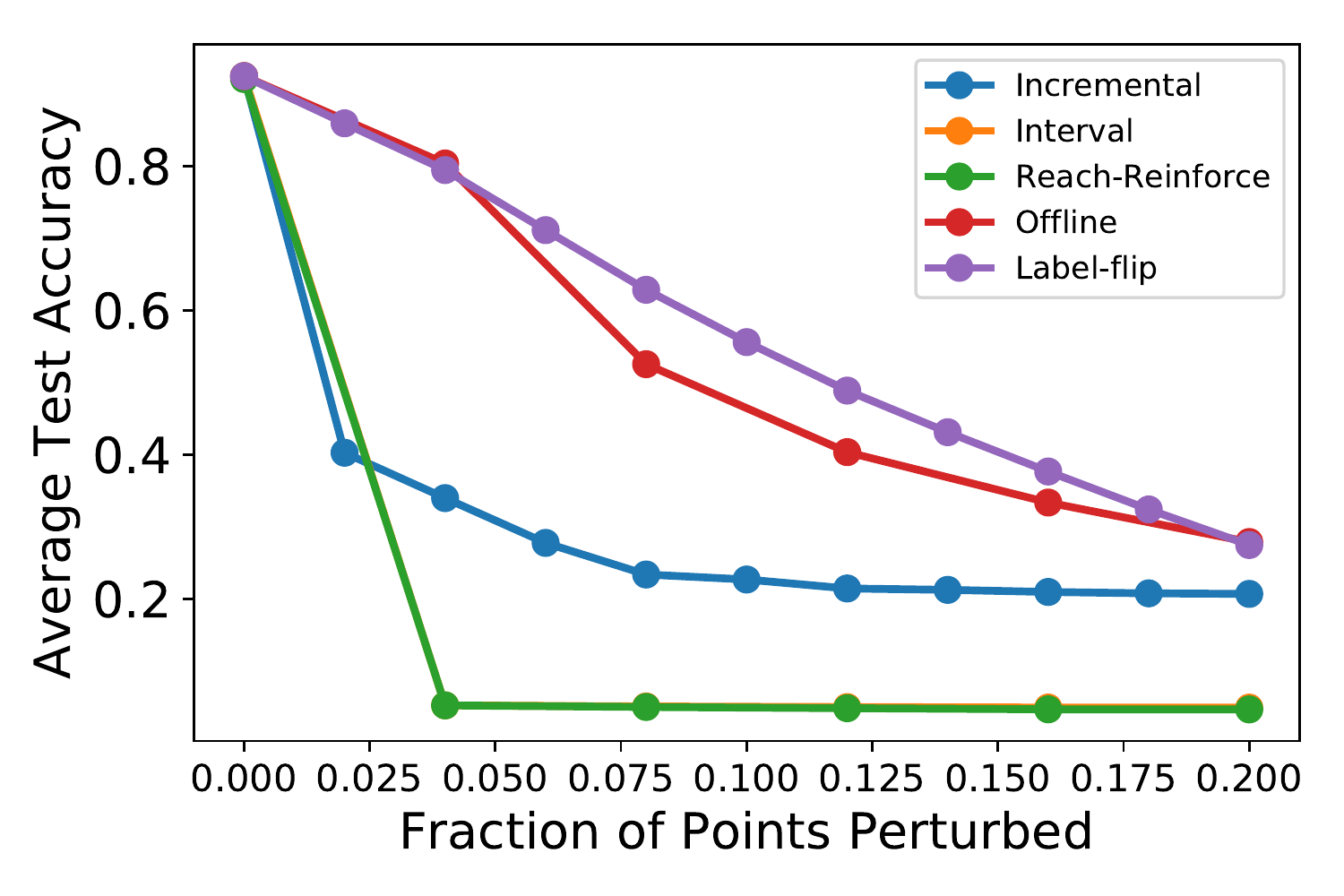} 
    \includegraphics[scale=0.3]{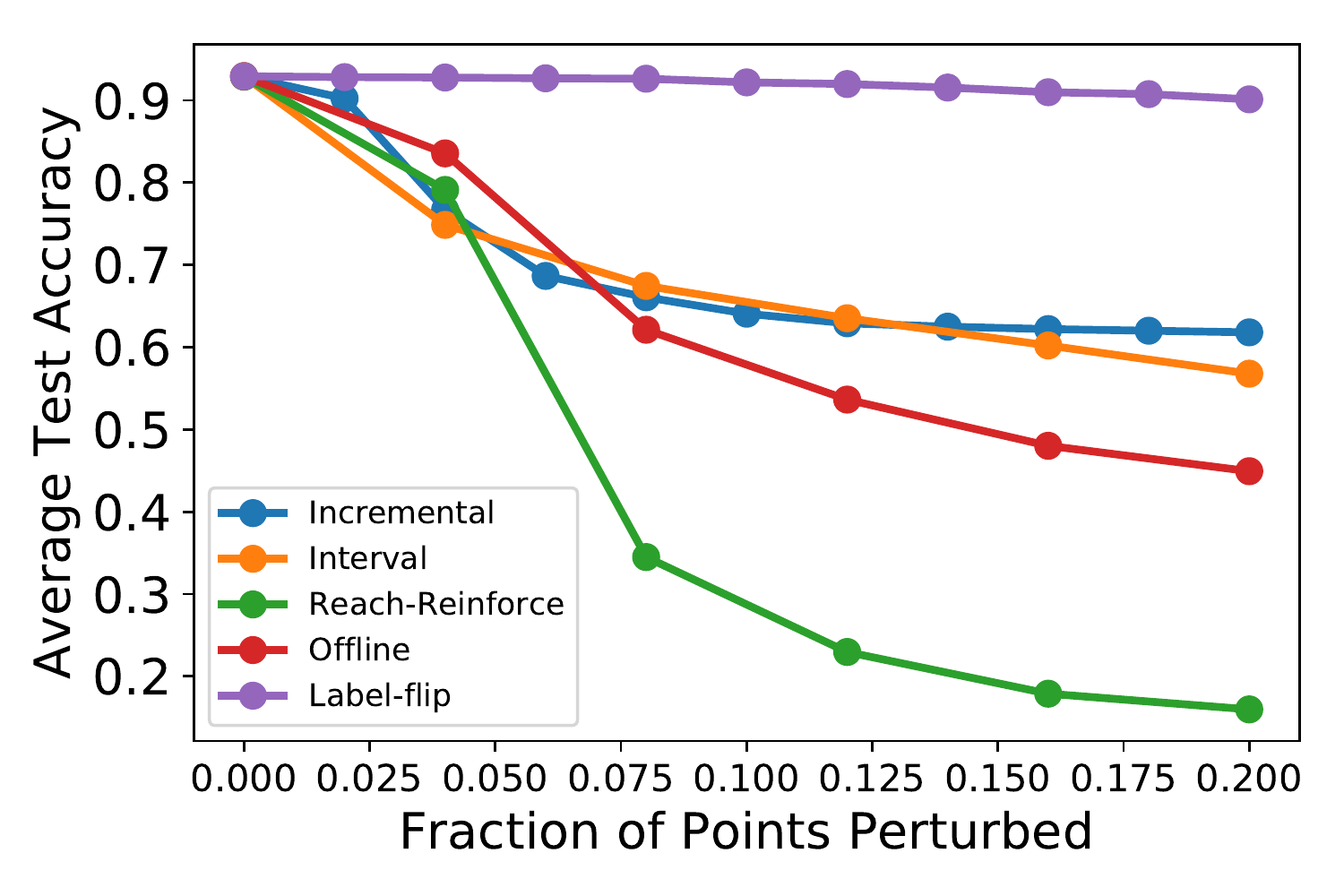}
    \captionof{figure}{Fashion MNIST Sandals v.s. Boots. \textbf{Top row:} Semi-online, \textbf{Bottom row:} Fully-online. \textbf{Left to right:} Slow Decay, Fast Decay, Constant}
    \label{fig:fashion_MNIST}
\end{center}

\begin{center}
    \centering
    \includegraphics[scale=0.3]{./spam/semi-onlineslow-decaytotal-accwarm.pdf}
    \includegraphics[scale=0.3]{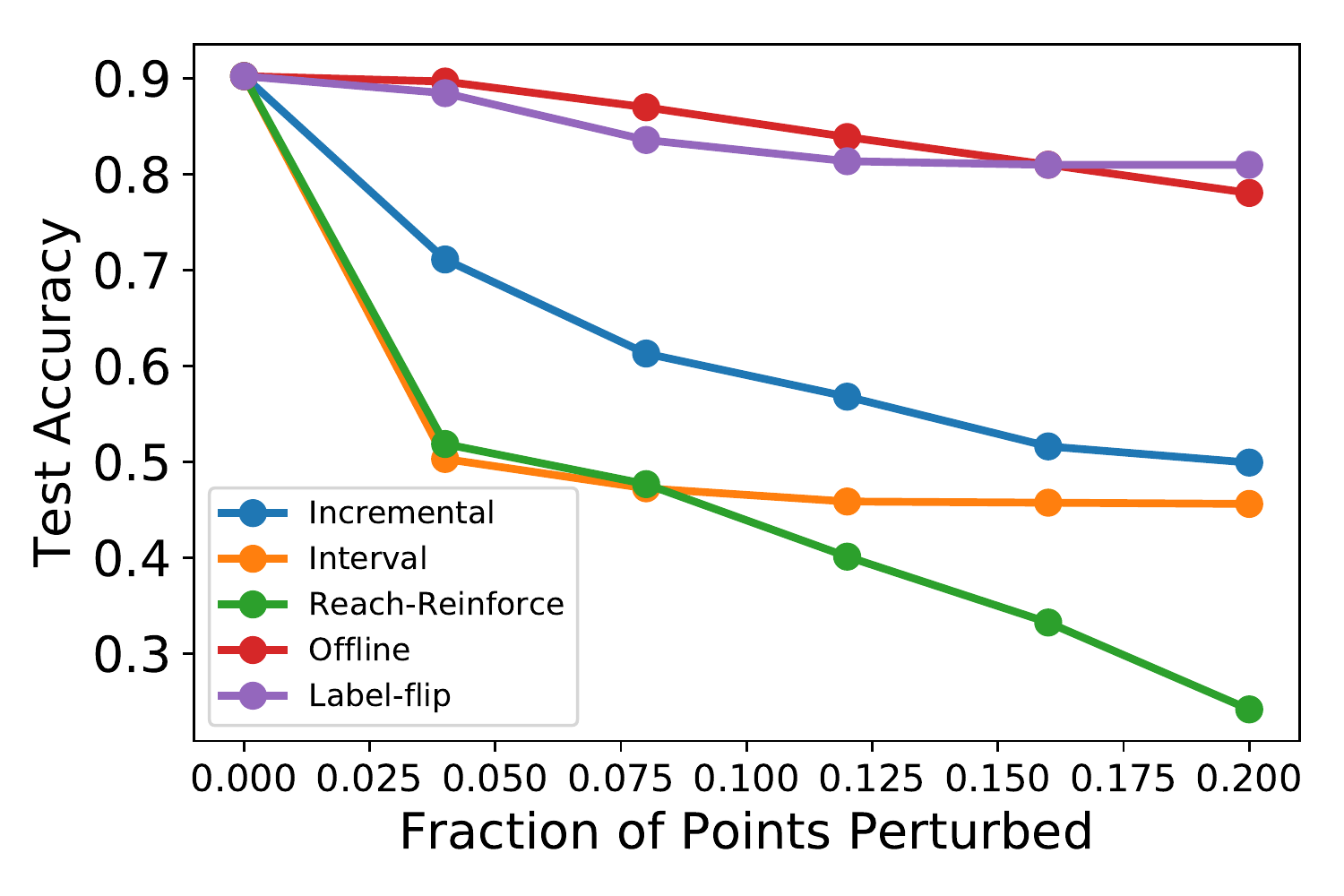} 
    \includegraphics[scale=0.3]{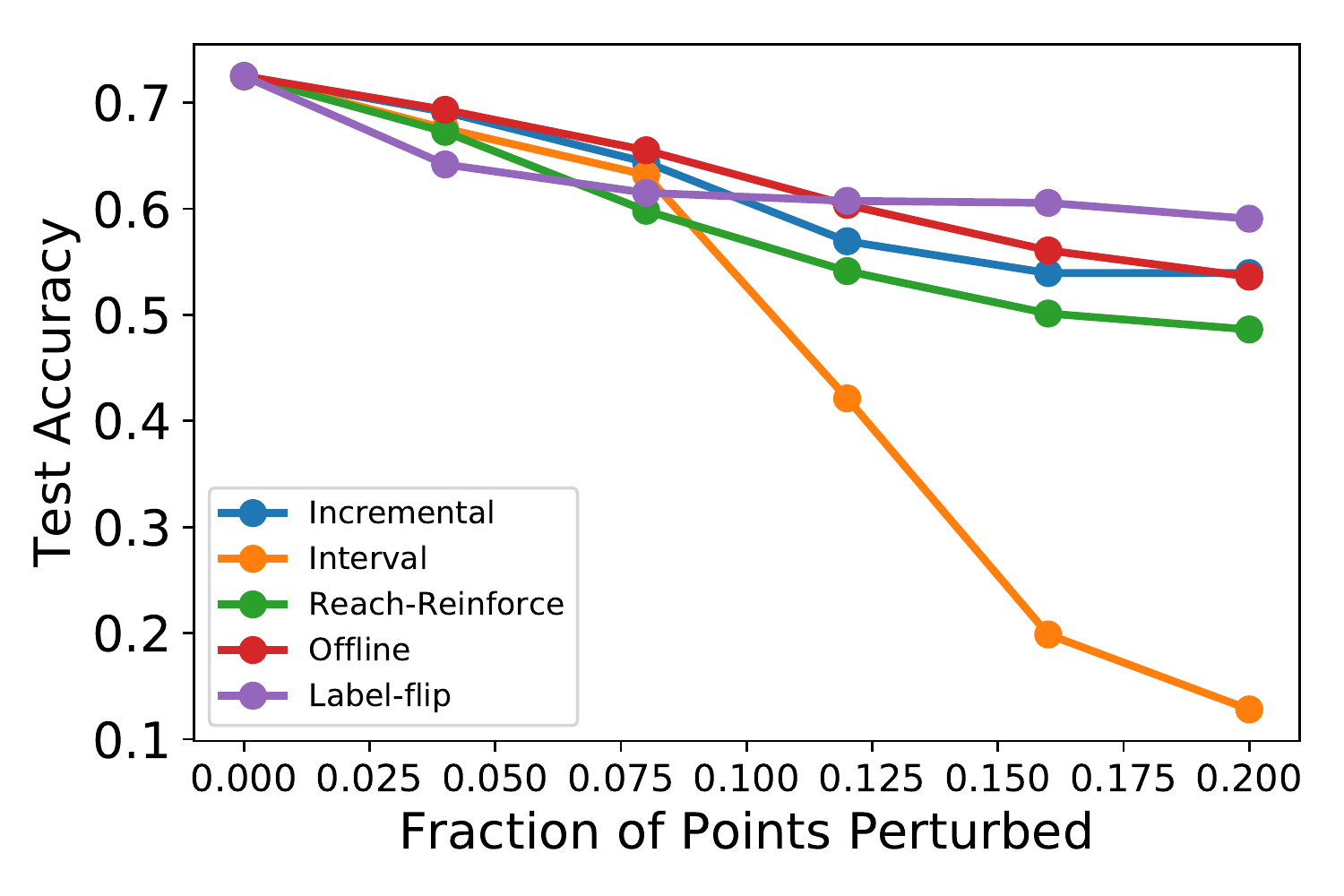}\\
    \includegraphics[scale=0.3]{./spam/full-onlineslow-decaytotal-accwarm.pdf}
    \includegraphics[scale=0.3]{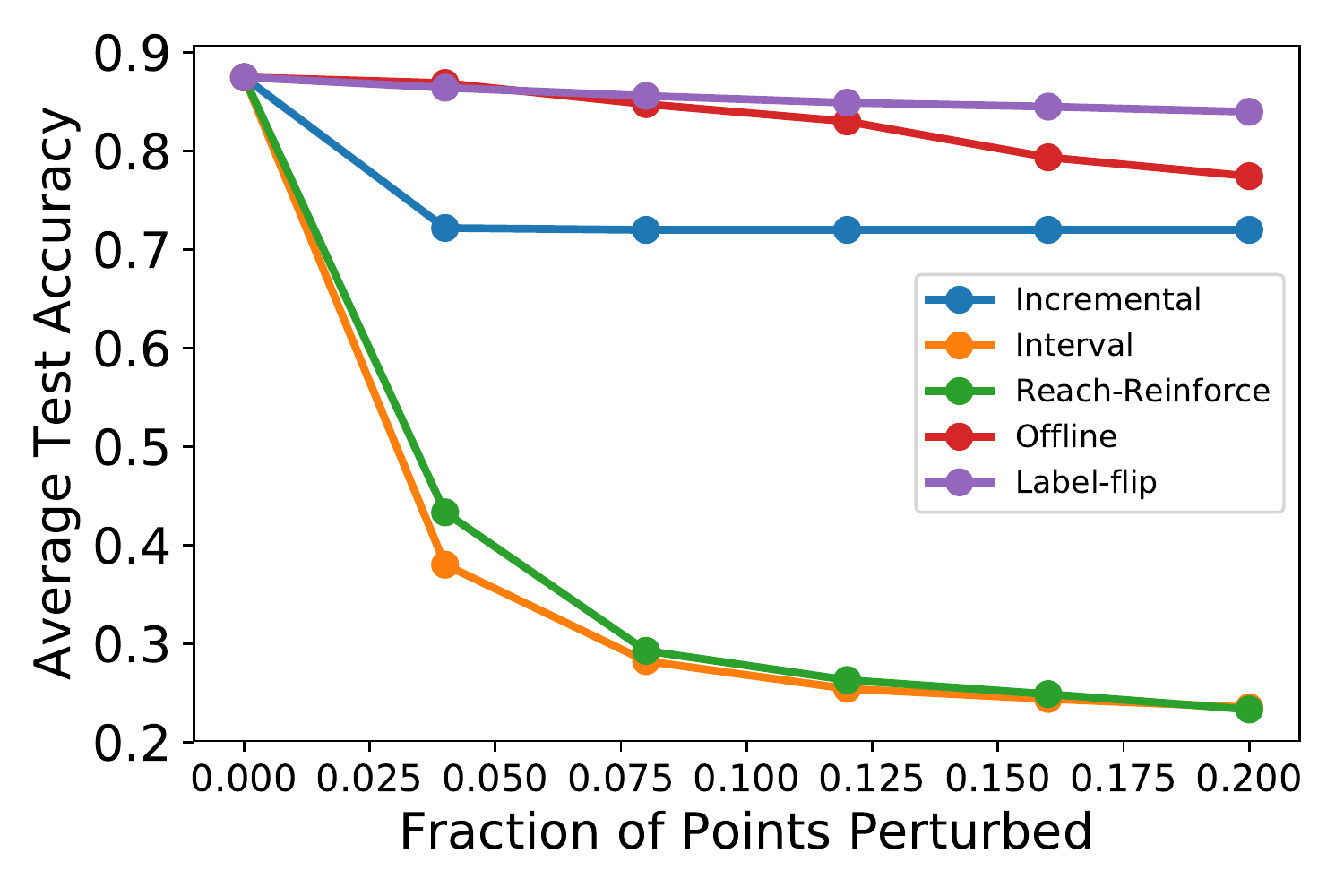} 
    \includegraphics[scale=0.3]{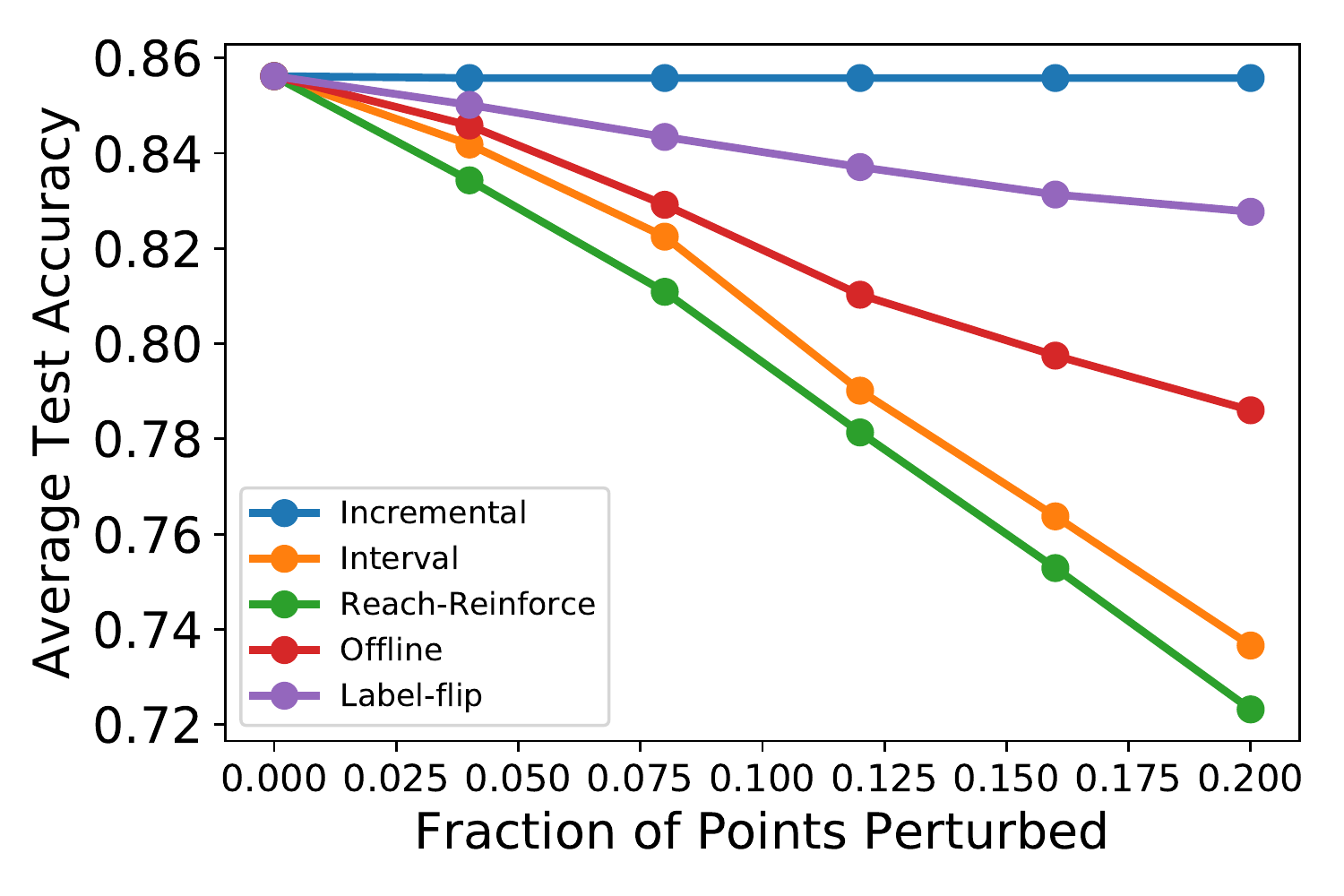}
    \captionof{figure}{UCI Spambase. \textbf{Top row:} Semi-online, \textbf{Bottom row:} Fully-online. \textbf{Left to right:} Slow Decay, Fast Decay, Constant}
    \label{fig:spam}
\end{center}

\subsection{Distribution of Attack Positions}
\label{sec:attackpositions}
In Table~\ref{tab:positions} in the experiment section, we qualitatively summarize the most frequent attack positions chosen by each online attack. In this section, we show the histogram plots of attack positions chosen by the Incremental and Interval attacks.

The entire length $T$ window is divided into $20$ equal-size bins. Each bar represents the frequency of modified points lying in the bin, normalized by the total number of points attacked in all repeated experiment. For Incremental attack, all modified points before reaching maximum attack budget of each series or maximum number of iteration are counted. For interval attack, we count the modified points when the attack budget is $160$ for 2-D Gaussian mixtures, $40$ for MNIST and fashion MNIST, and $80$ for UCI Spamset.

\subsubsection{2D Gaussian Mixtures}
\begin{center}
    \centering
    \includegraphics[scale=0.3]{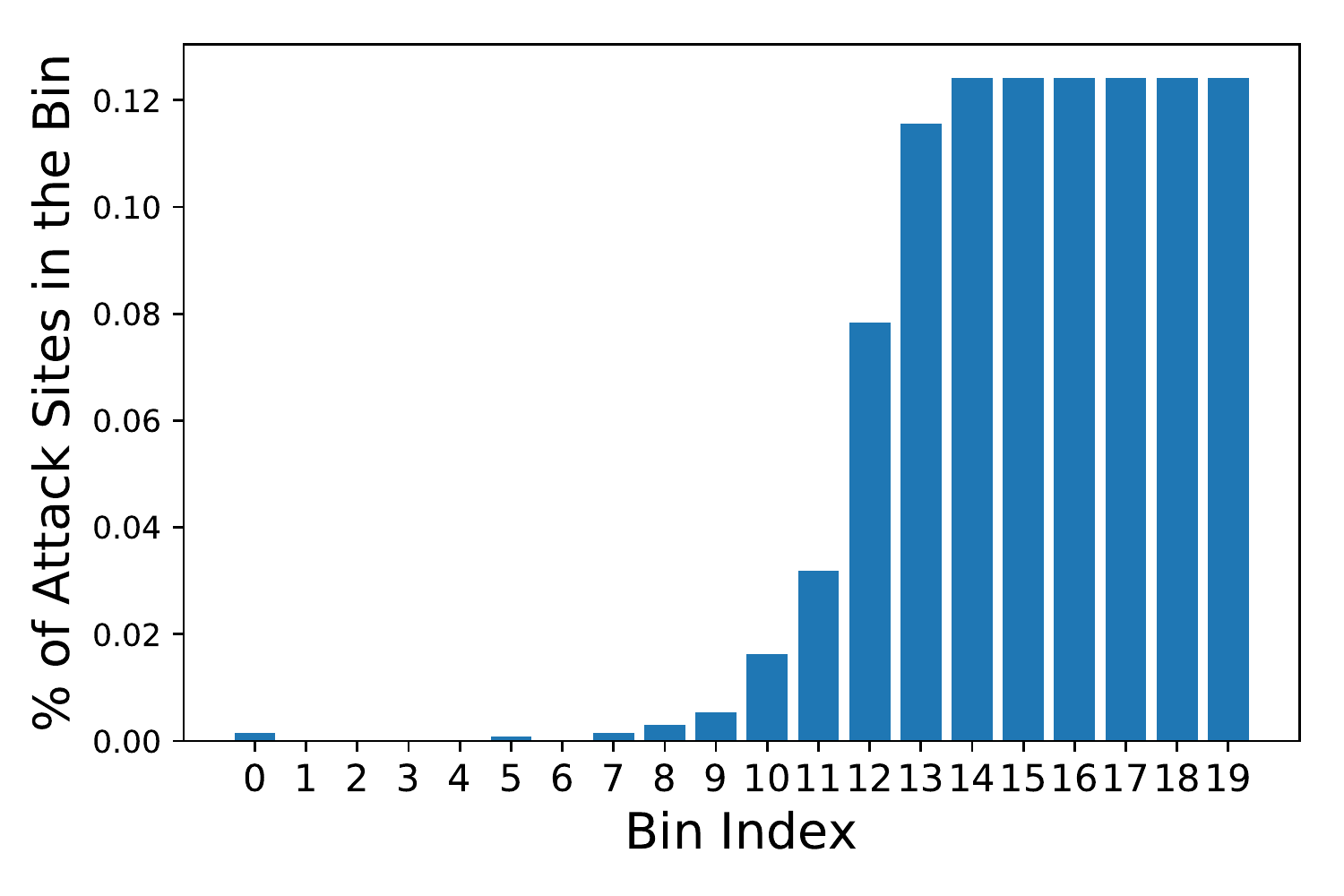}
    \includegraphics[scale=0.3]{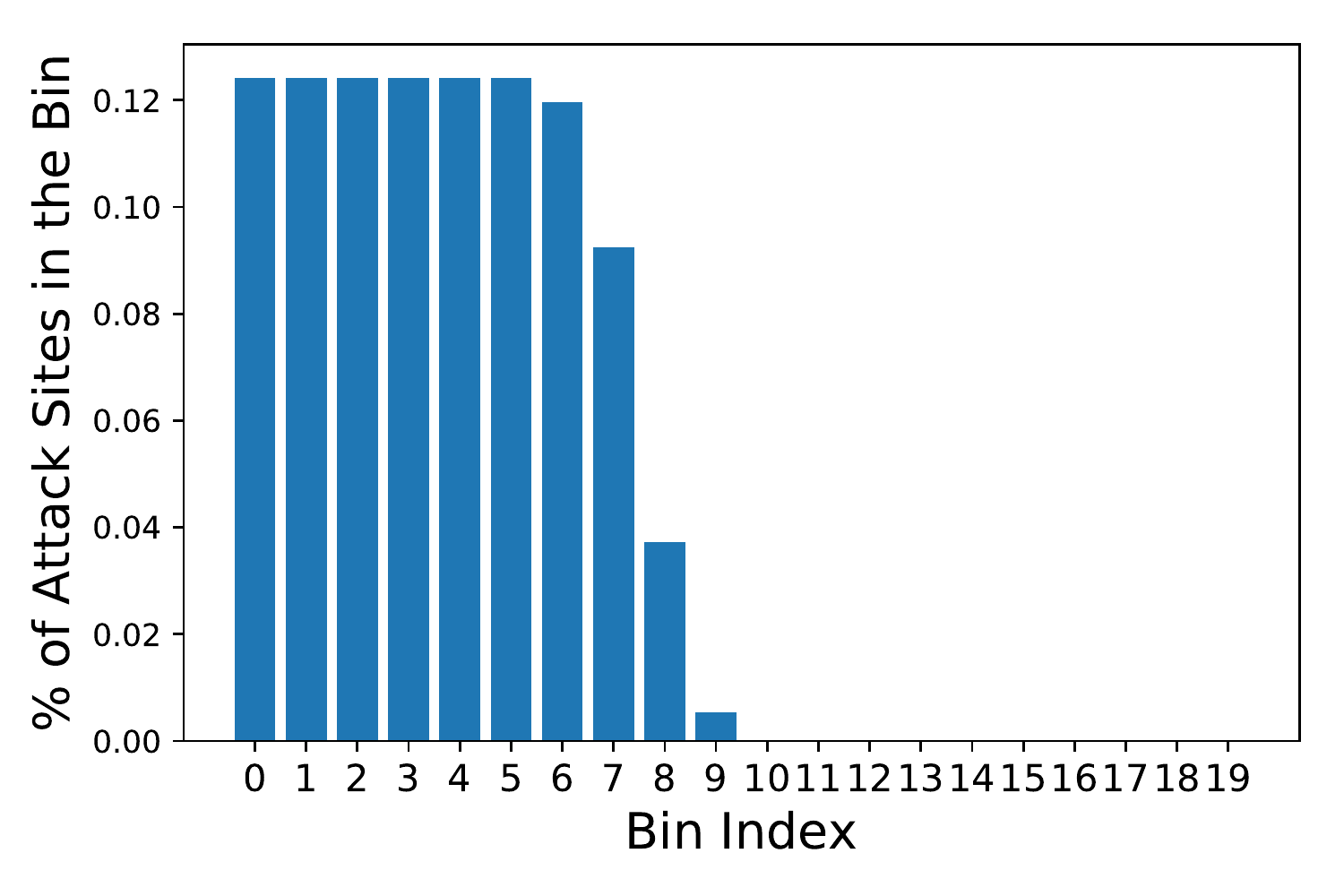}
    \includegraphics[scale=0.3]{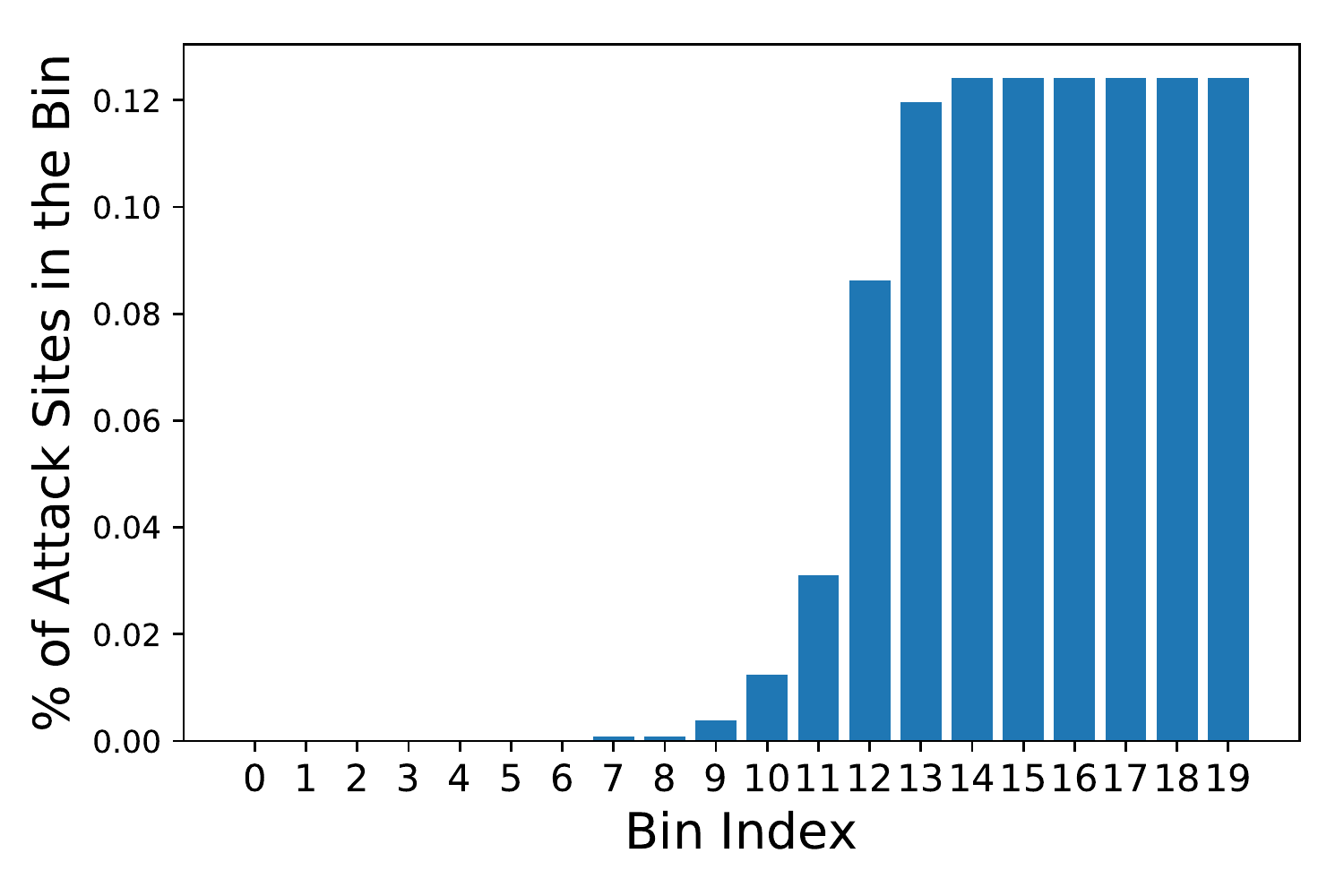}\\ 
    \includegraphics[scale=0.3]{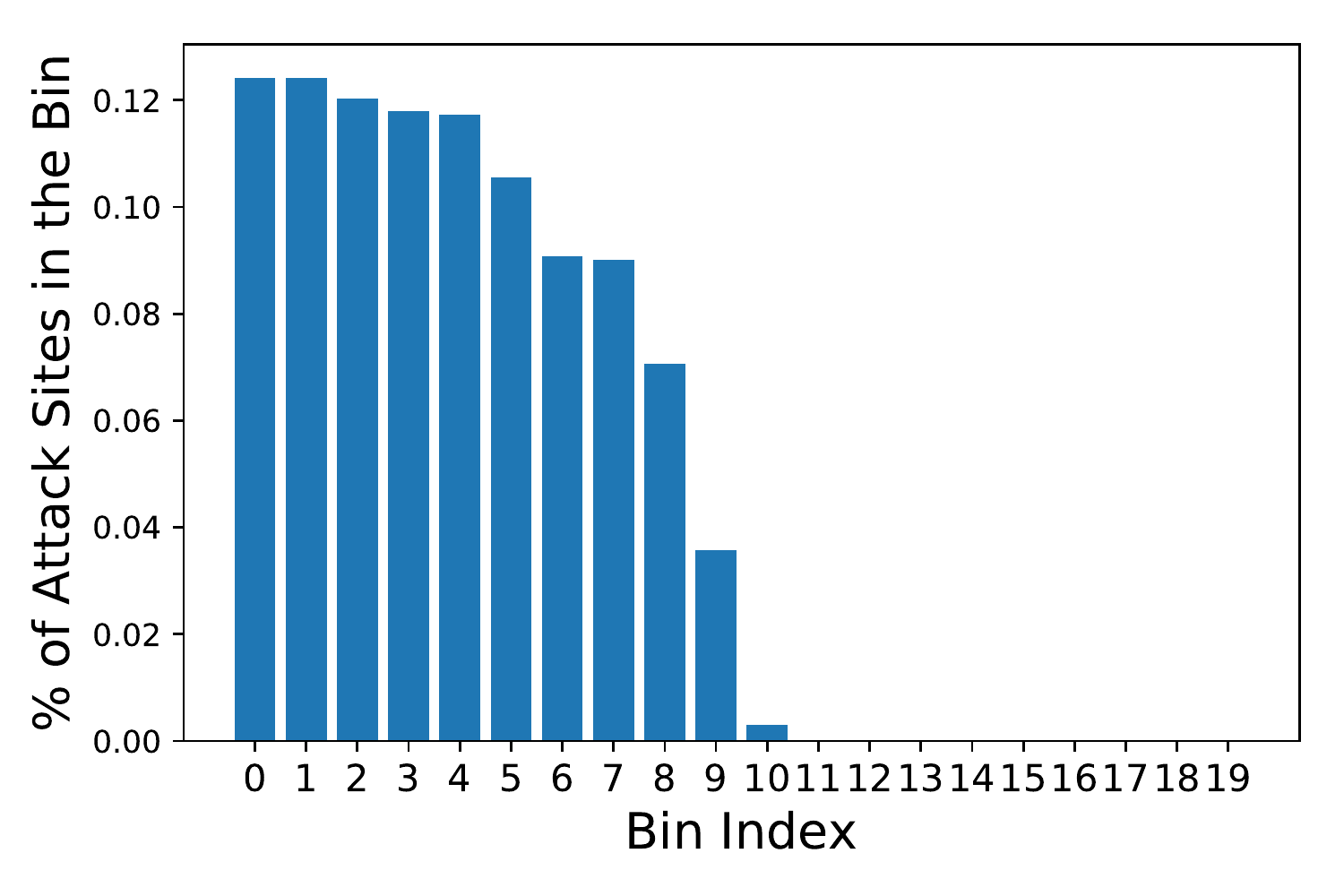}
    \includegraphics[scale=0.3]{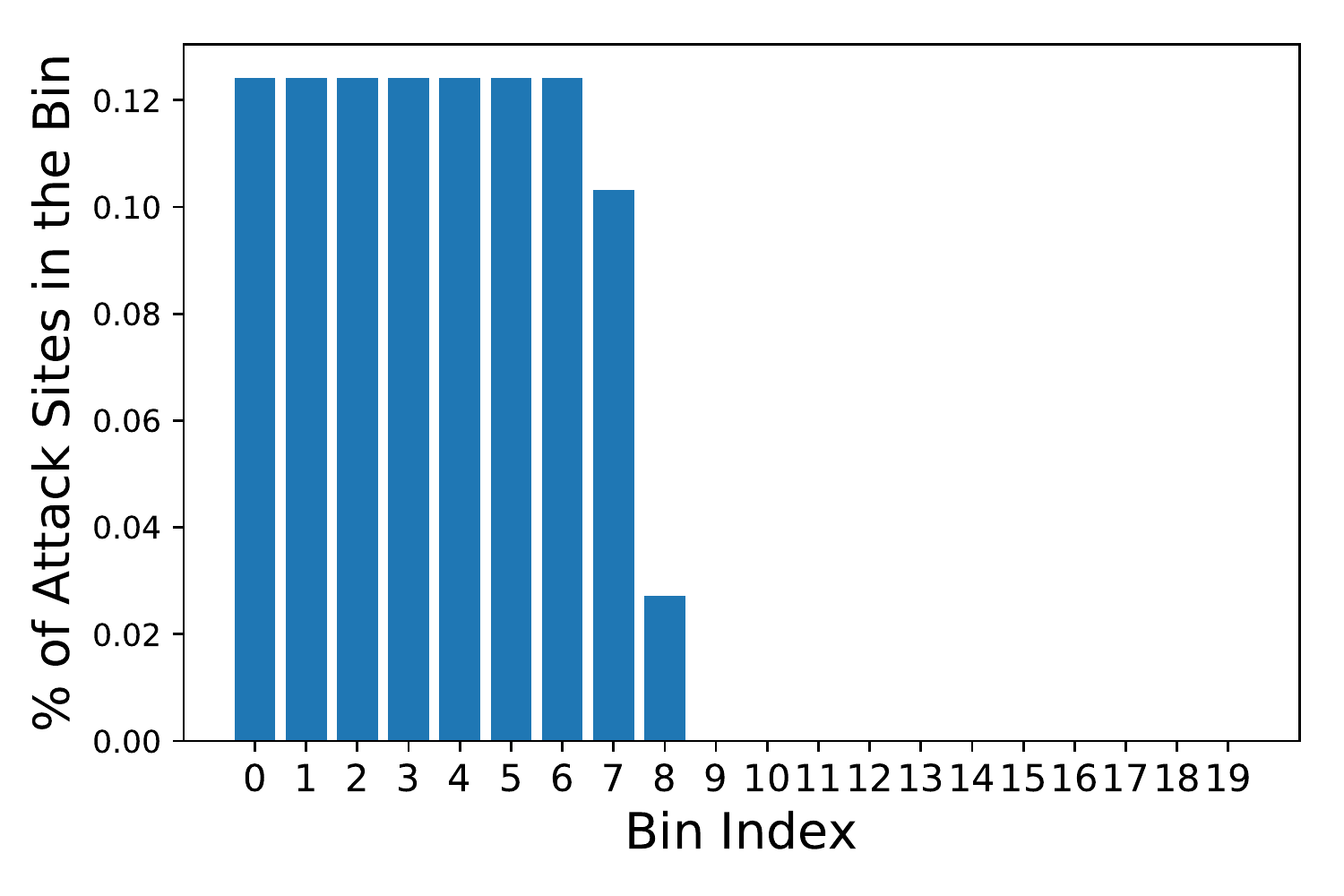}
    \includegraphics[scale=0.3]{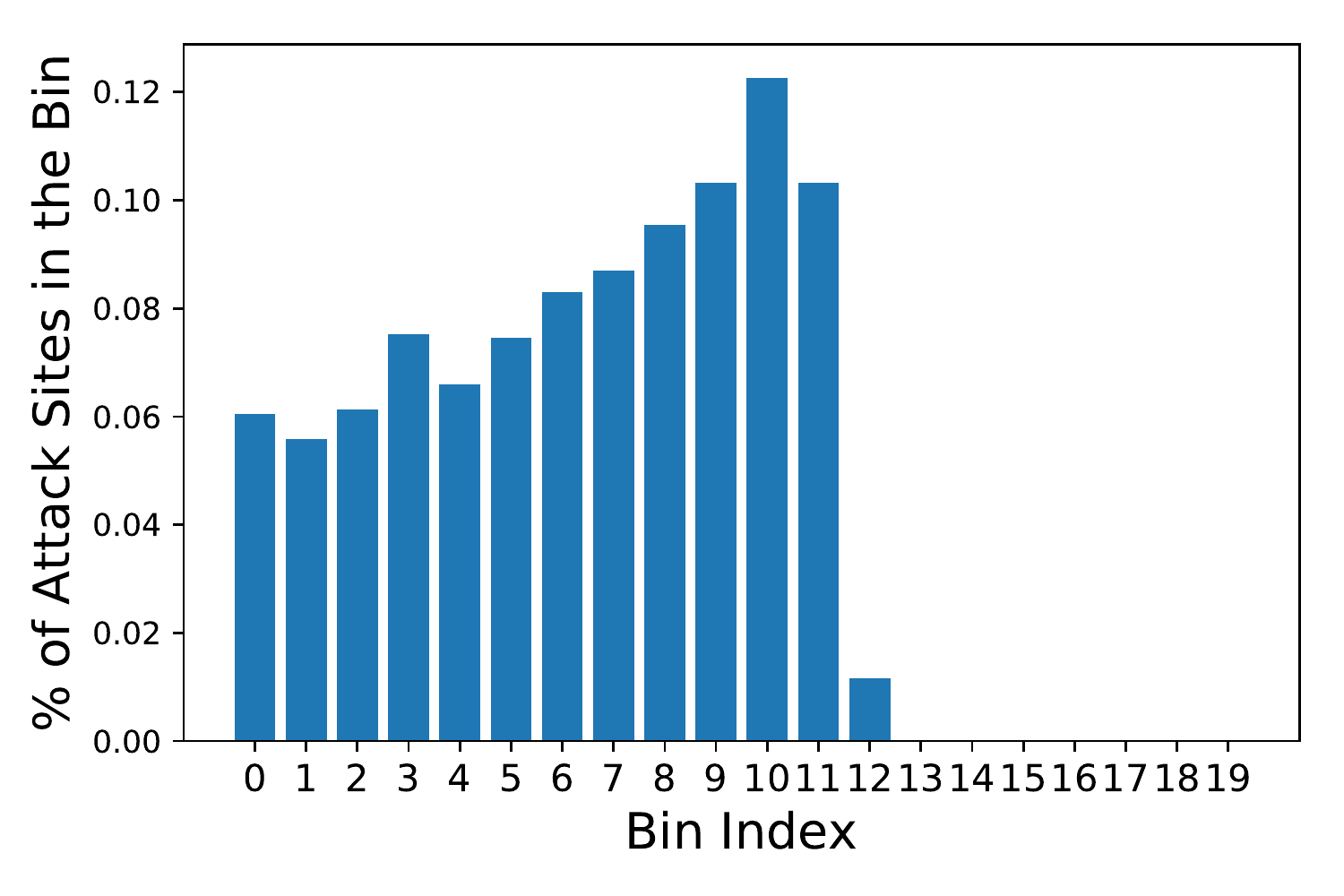}
    \captionof{figure}{Results for 2d Gaussian mixtures, Incremental \textbf{Top row:} Semi-online, \textbf{Bottom row:} Full-online. \textbf{Left to right:} Slow Decay, Fast Decay, Constant}
    \label{fig:2dpos}
\end{center}

\begin{center}
    \centering
    \includegraphics[scale=0.3]{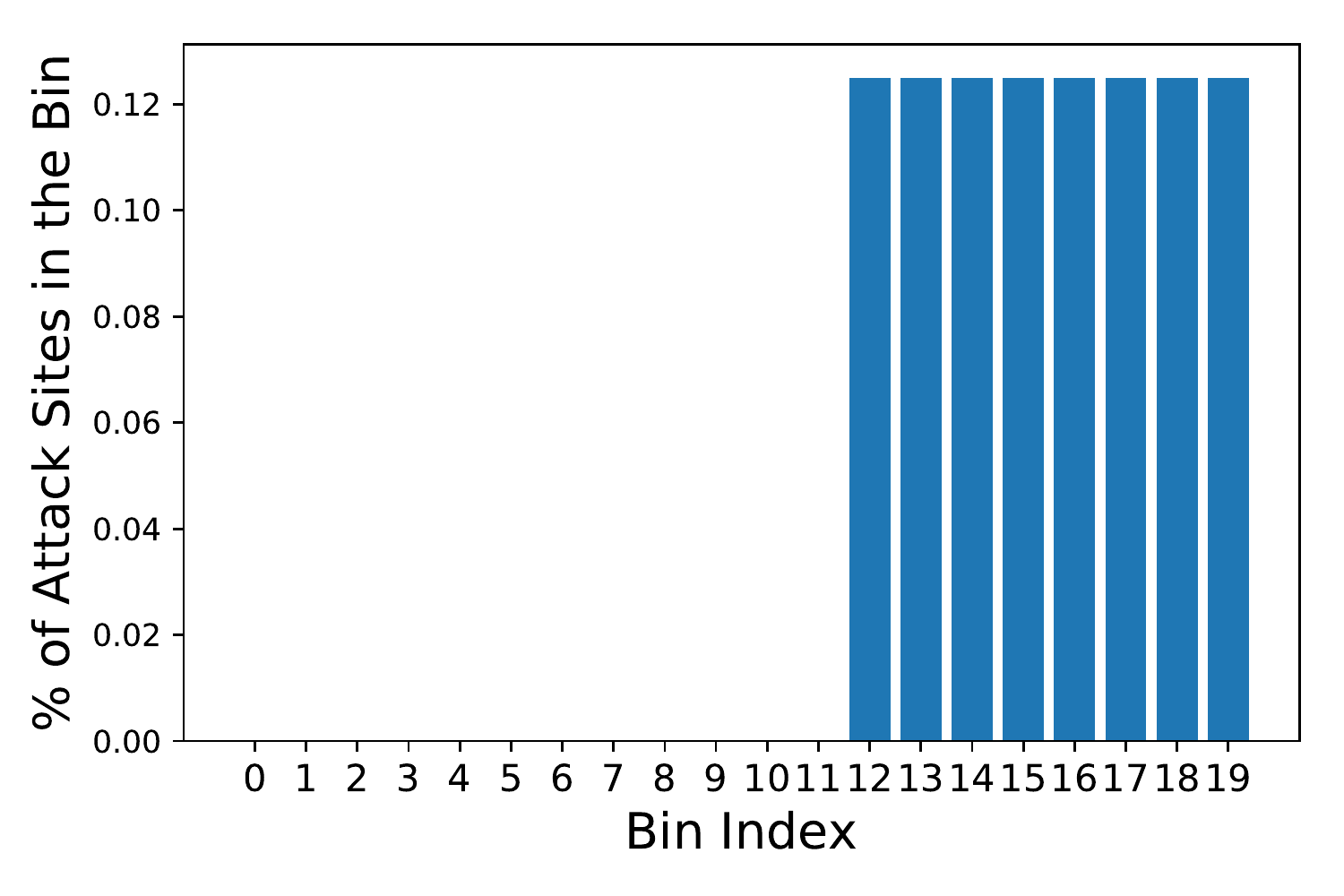}
    \includegraphics[scale=0.3]{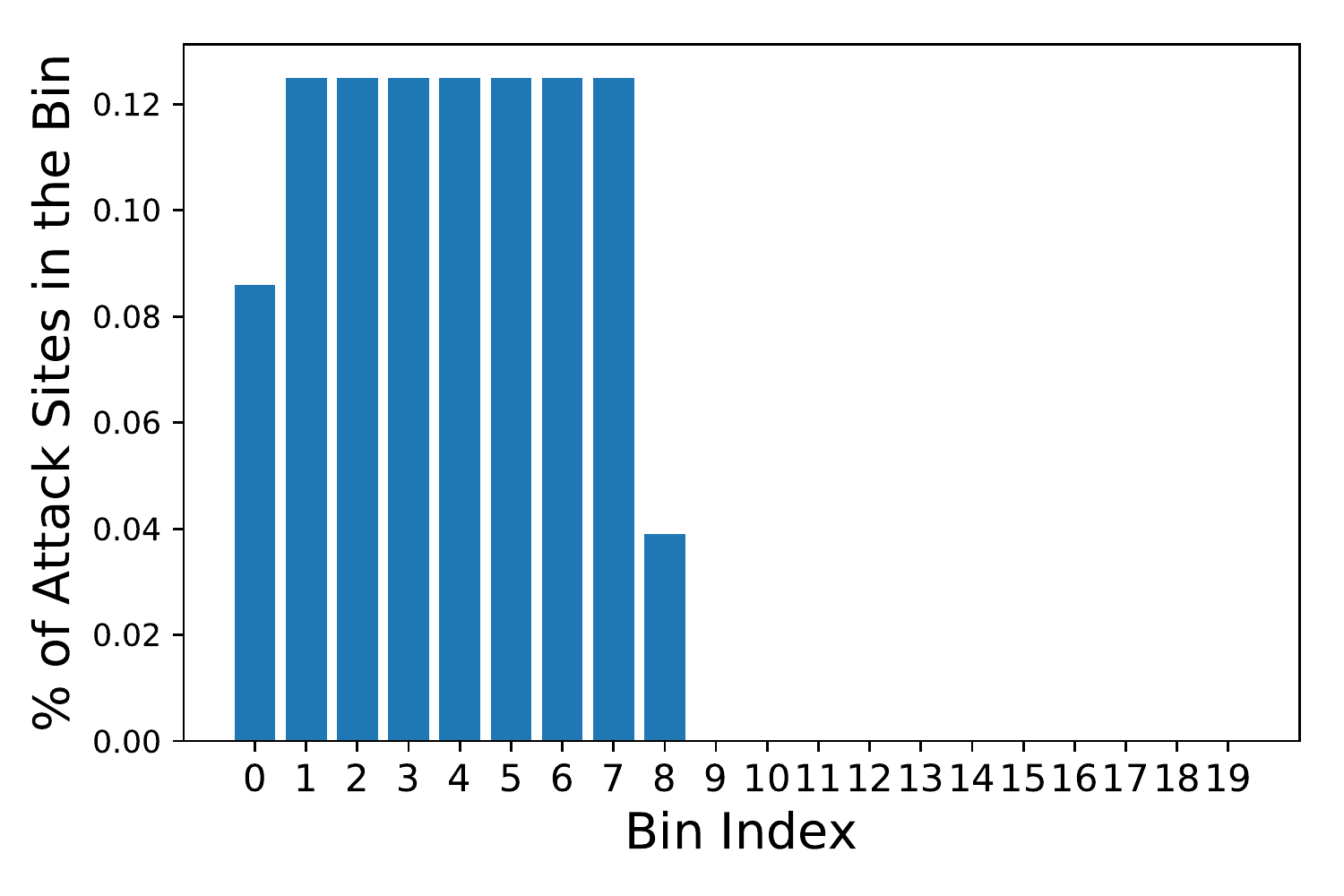}
    \includegraphics[scale=0.3]{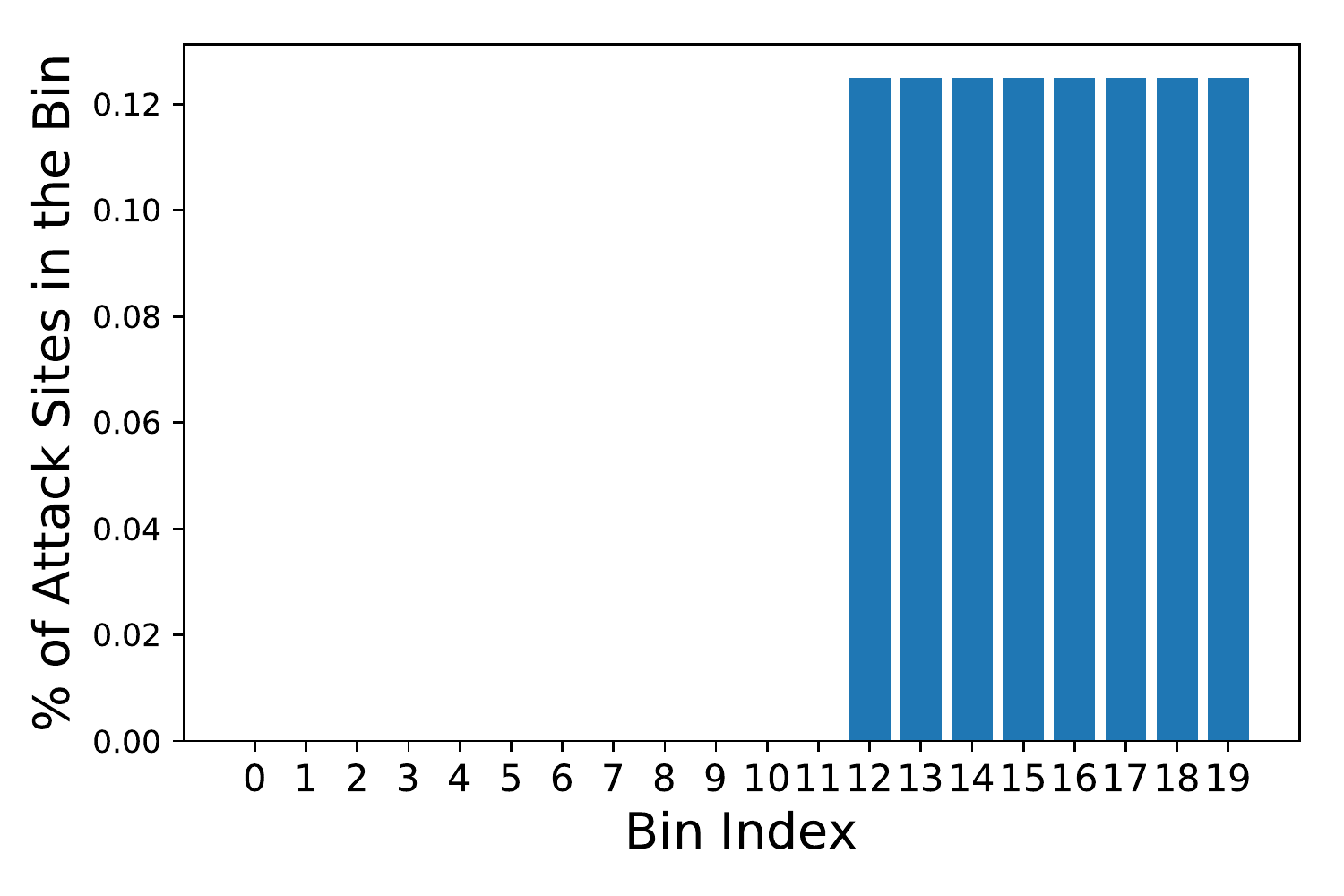}\\ 
    \includegraphics[scale=0.3]{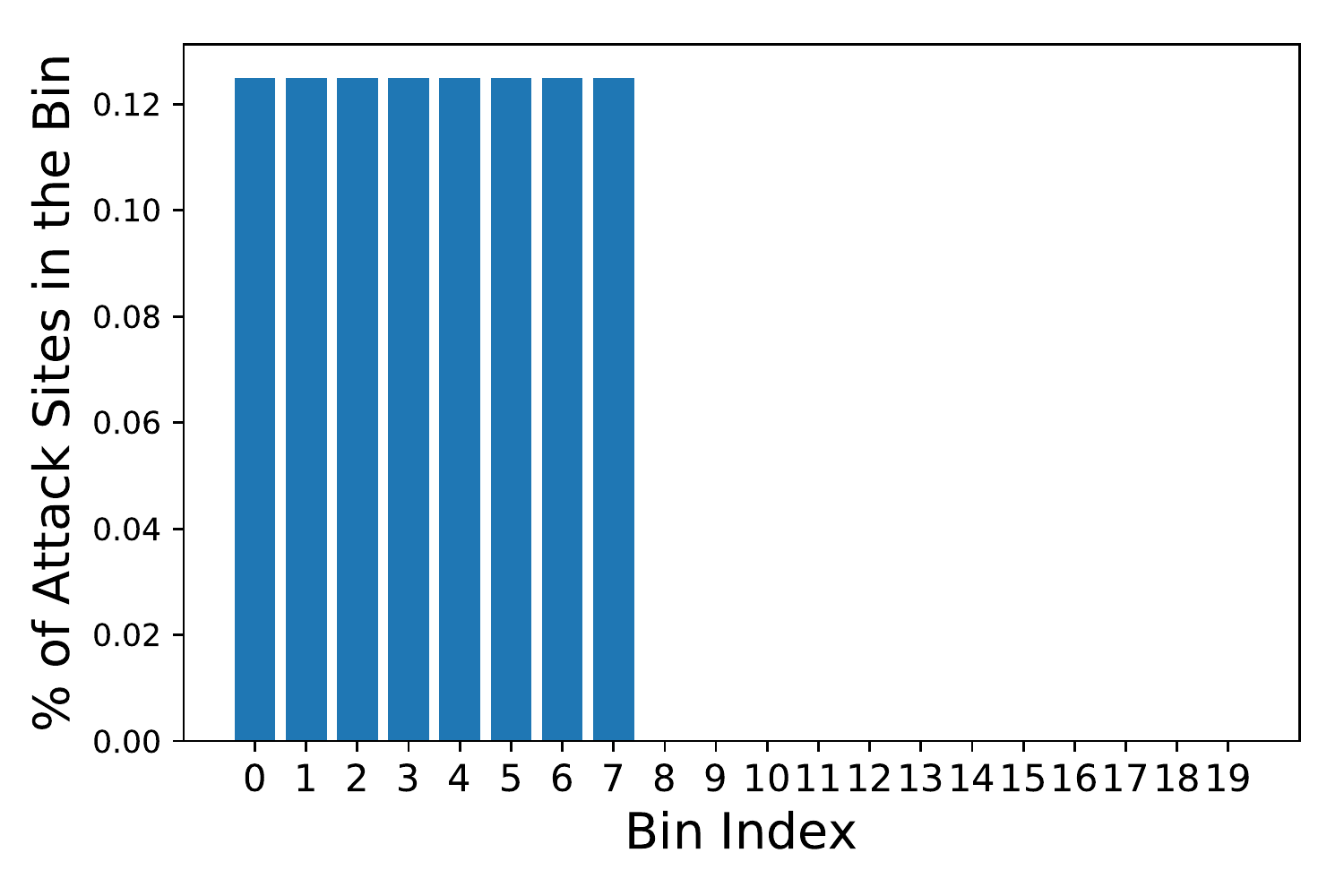}
    \includegraphics[scale=0.3]{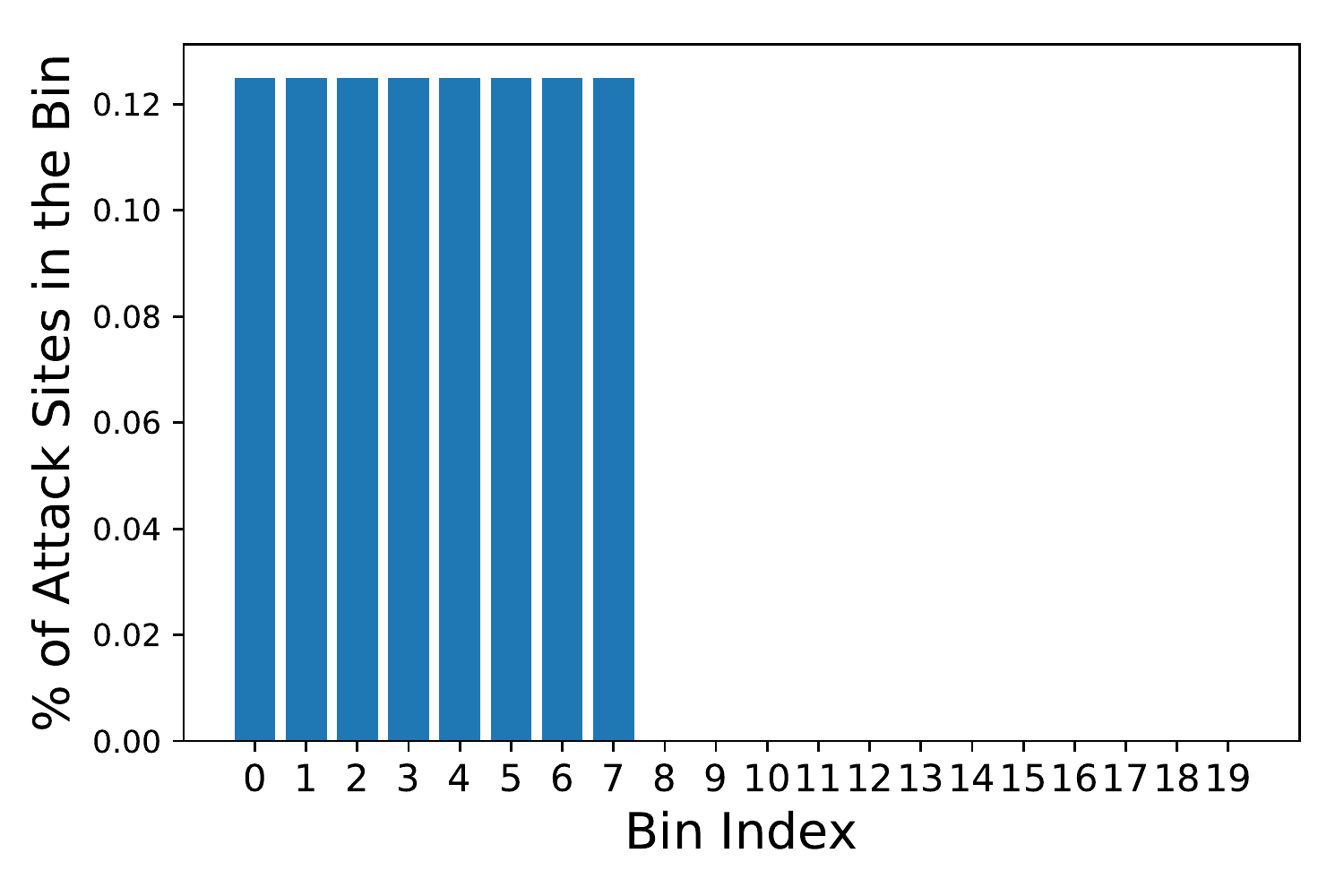}
    \includegraphics[scale=0.3]{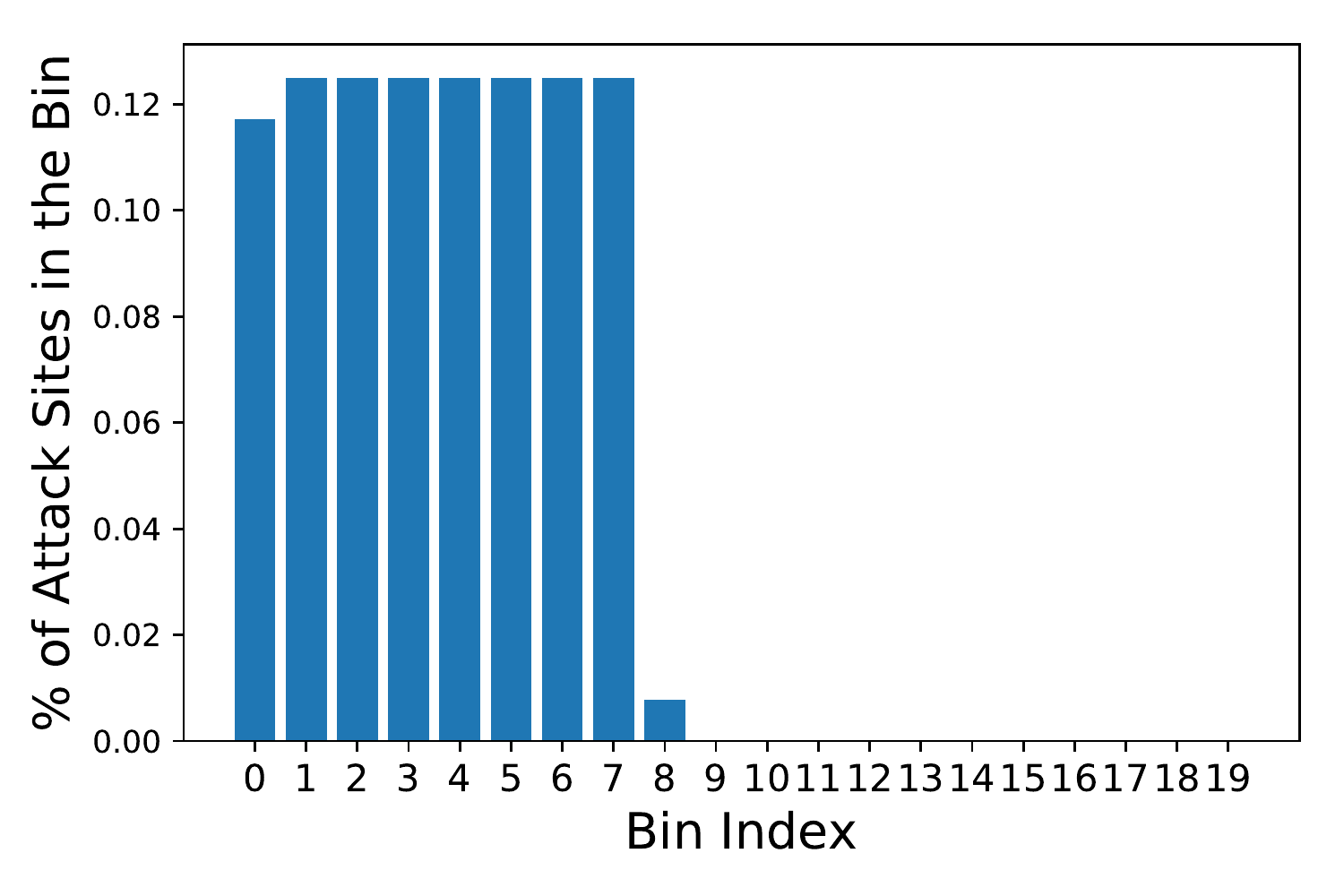}
    \captionof{figure}{Results for 2d Gaussian mixtures, Interval \textbf{Top row:} Semi-online, \textbf{Bottom row:} Full-online. \textbf{Left to right:} Slow Decay, Fast Decay, Constant}
    \label{fig:2dpos}
\end{center}

\subsubsection{MNIST 1 v.s. 7}
\begin{center}
    \centering
    \includegraphics[scale=0.3]{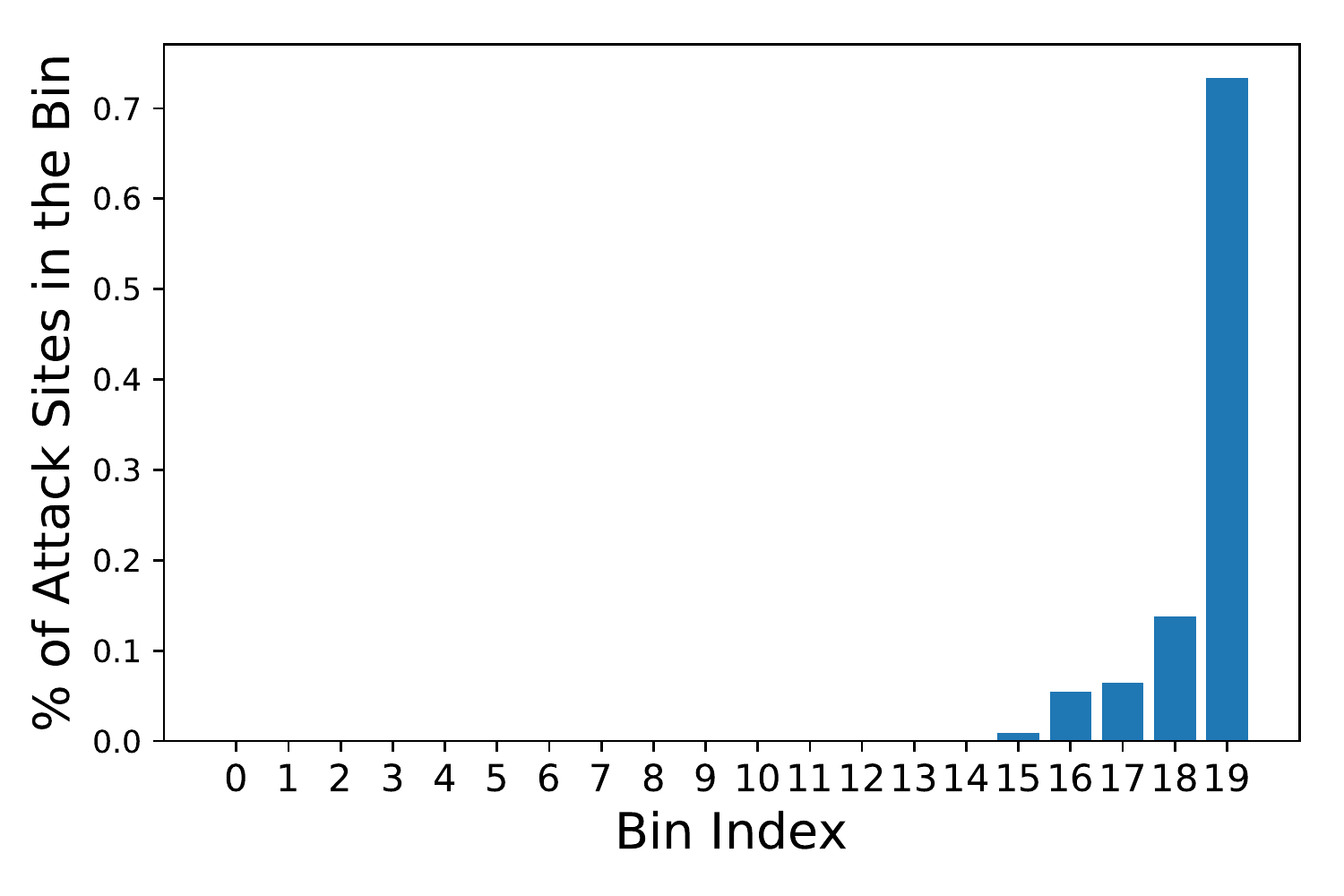}
    \includegraphics[scale=0.3]{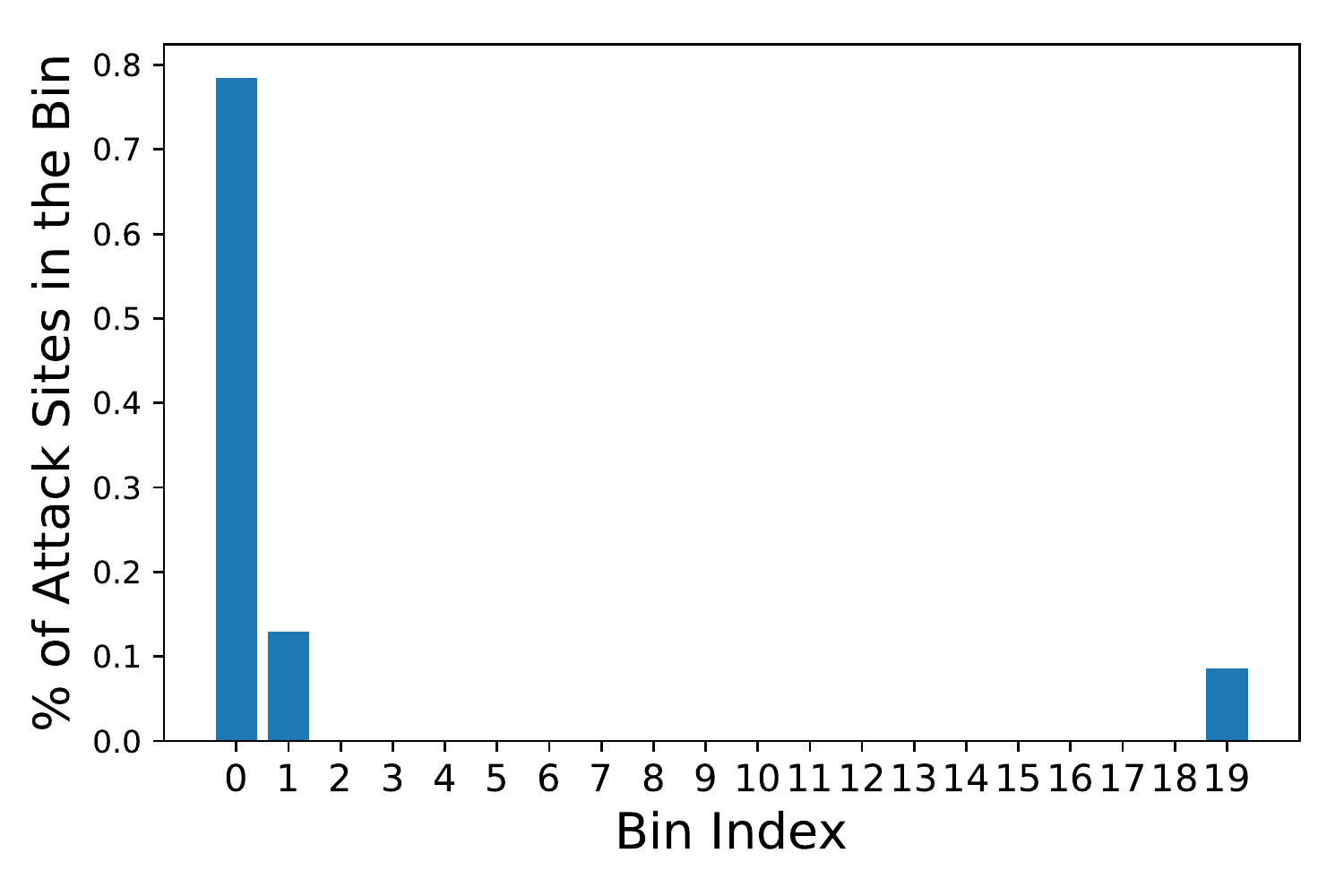}
    \includegraphics[scale=0.3]{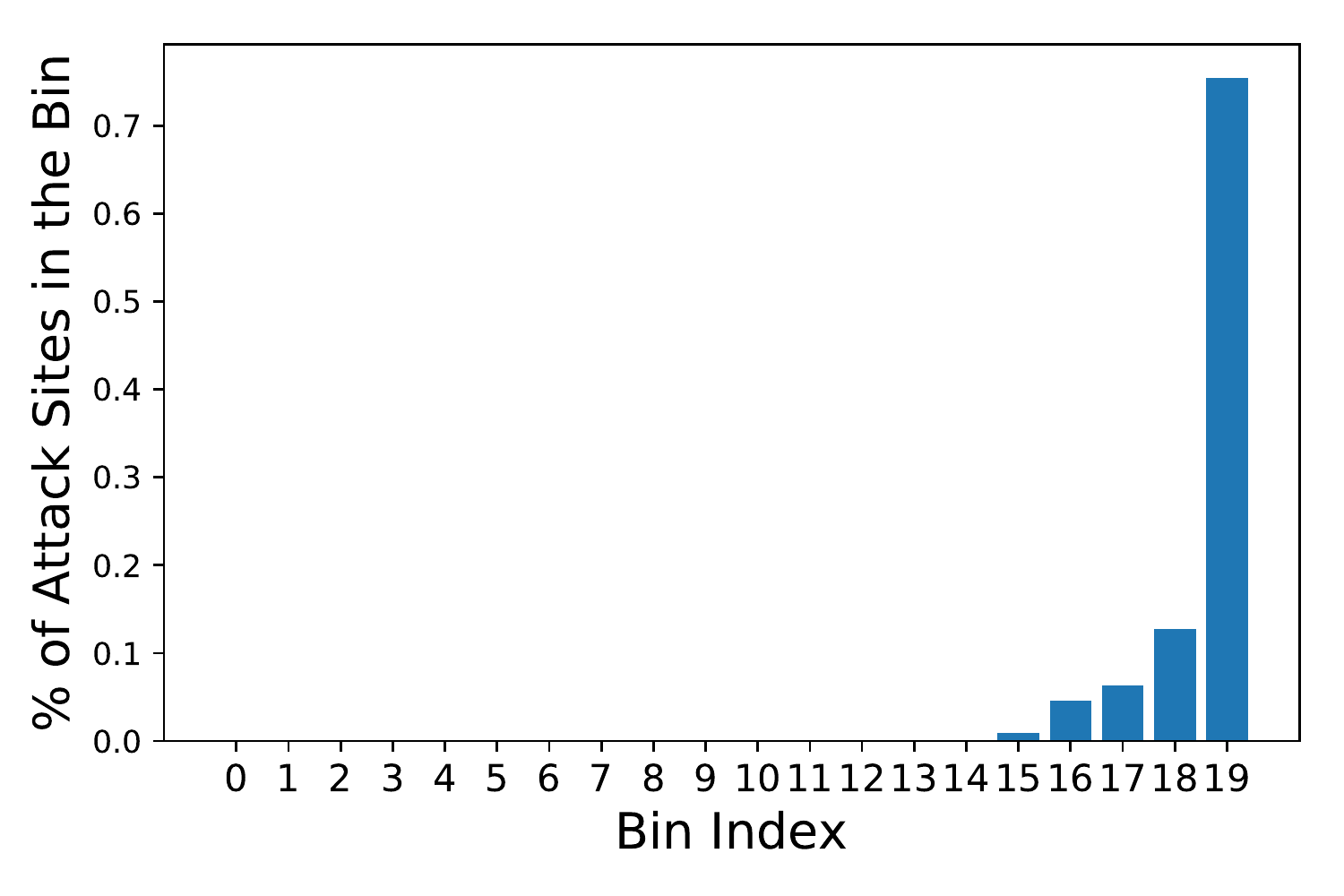}\\ 
    \includegraphics[scale=0.3]{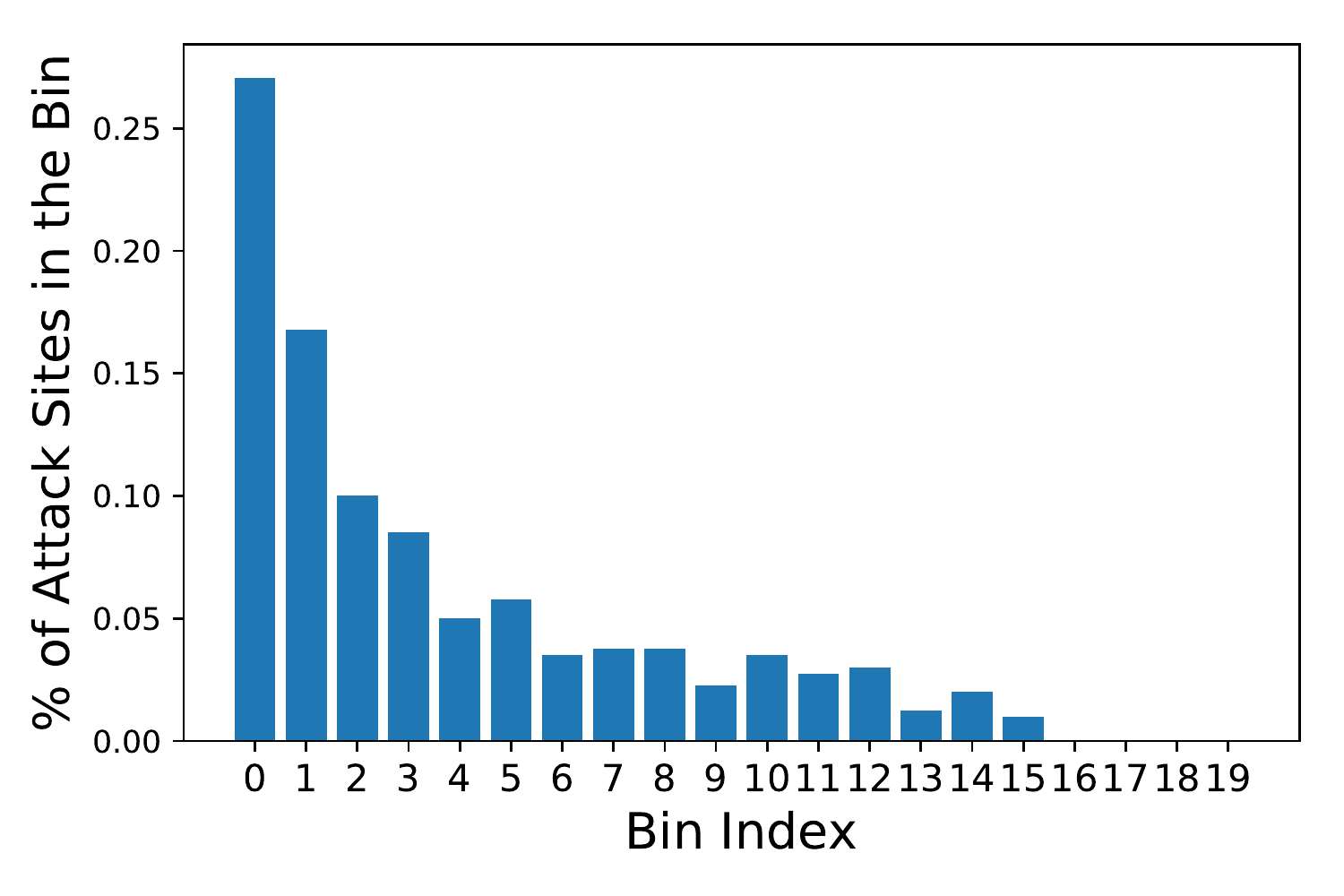}
    \includegraphics[scale=0.3]{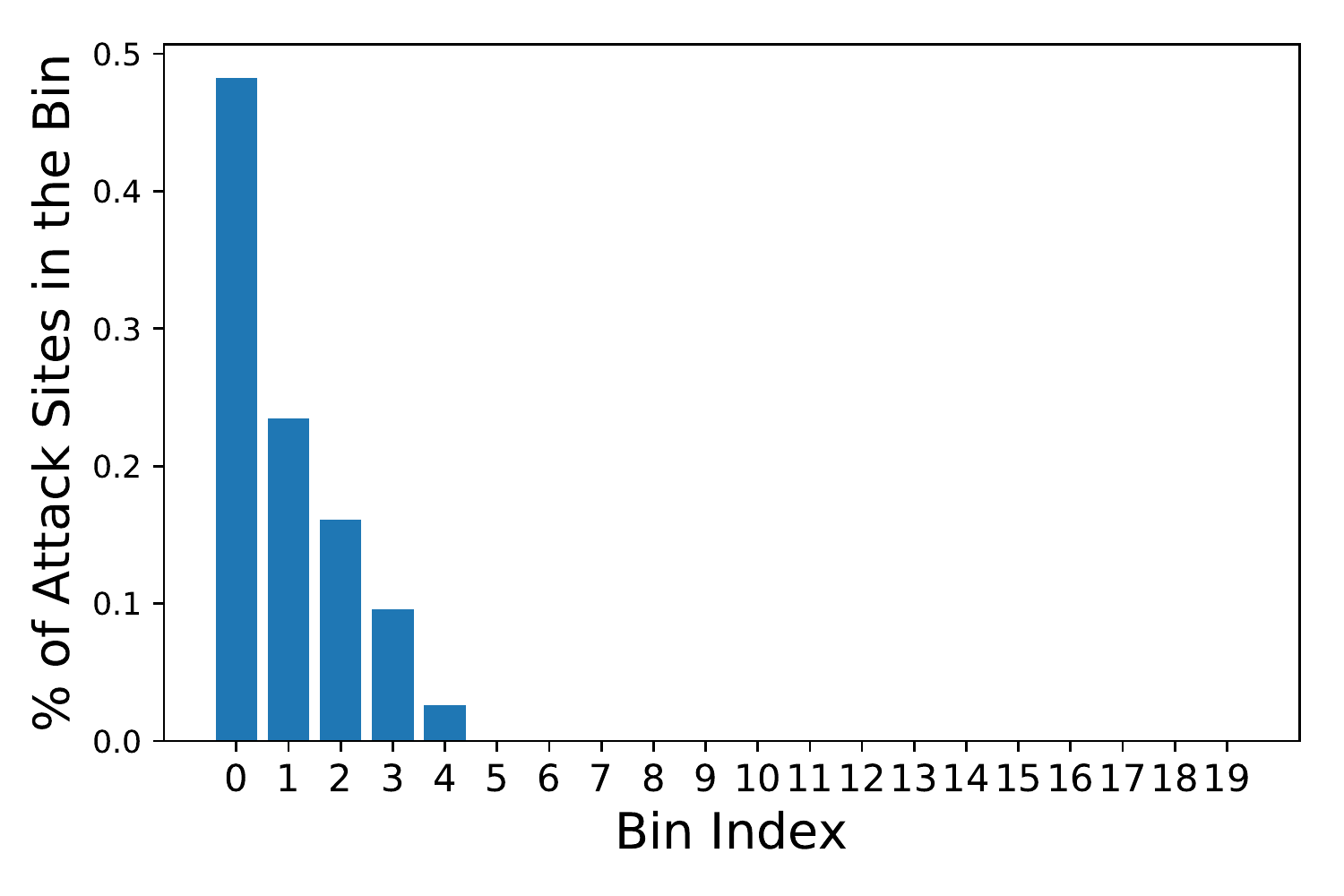}
    \includegraphics[scale=0.3]{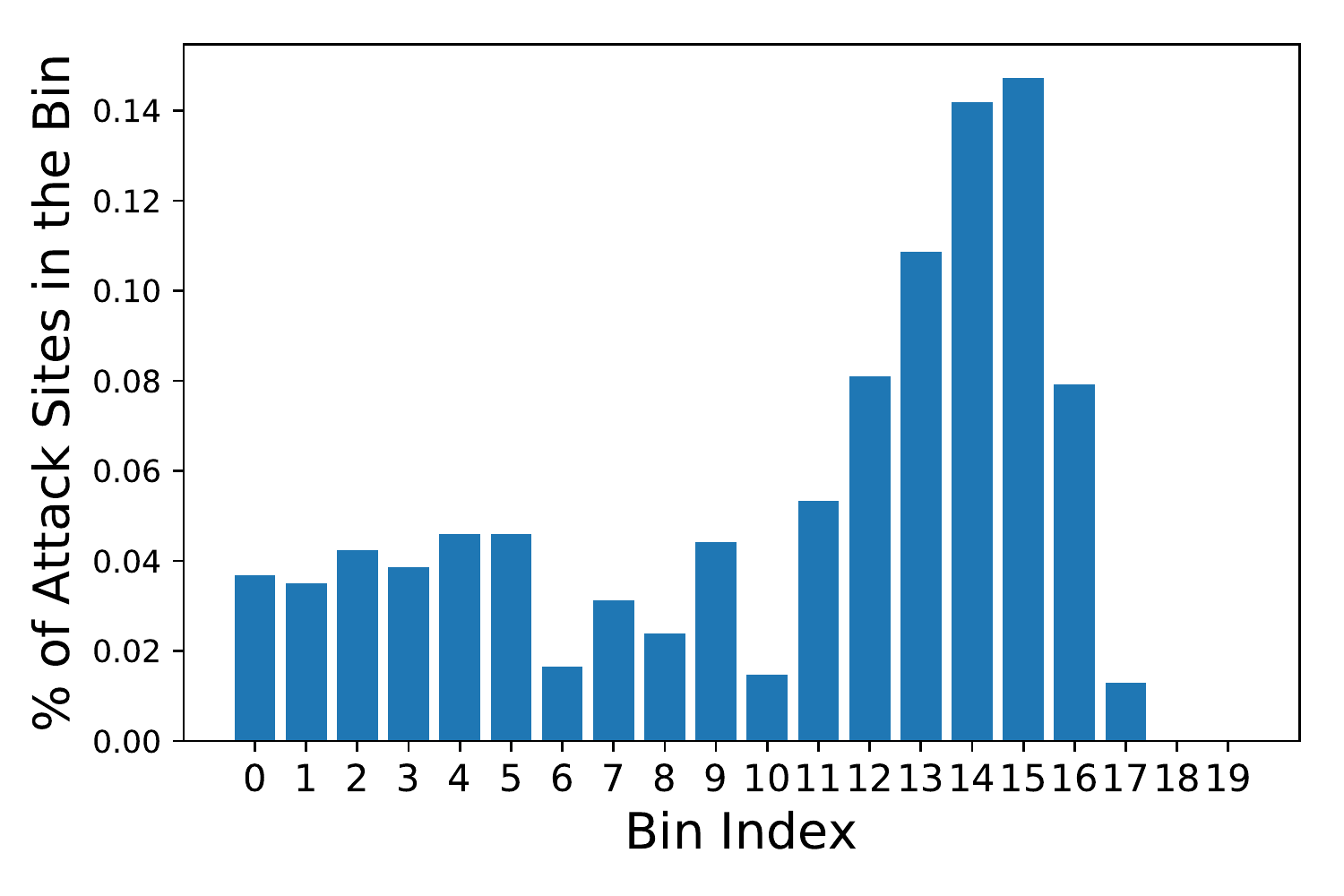}
    \captionof{figure}{Results for MNIST, Incremental. \textbf{Top row:} Semi-online, \textbf{Bottom row:} Full-online. \textbf{Left to right:} Slow Decay, Fast Decay, Constant}
    \label{fig:MNISTpos}
\end{center}

\begin{center}
    \centering
    \includegraphics[scale=0.3]{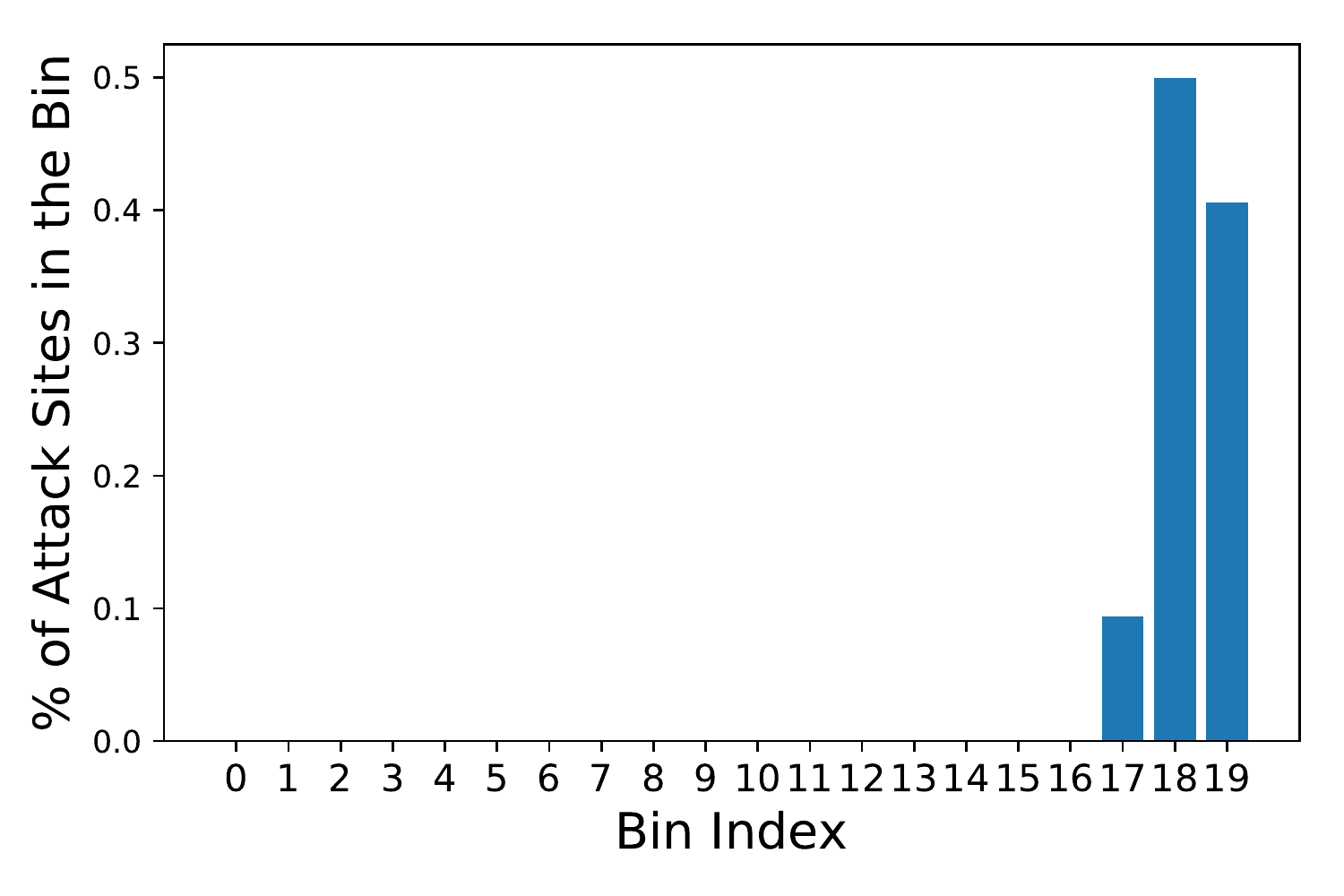}
    \includegraphics[scale=0.3]{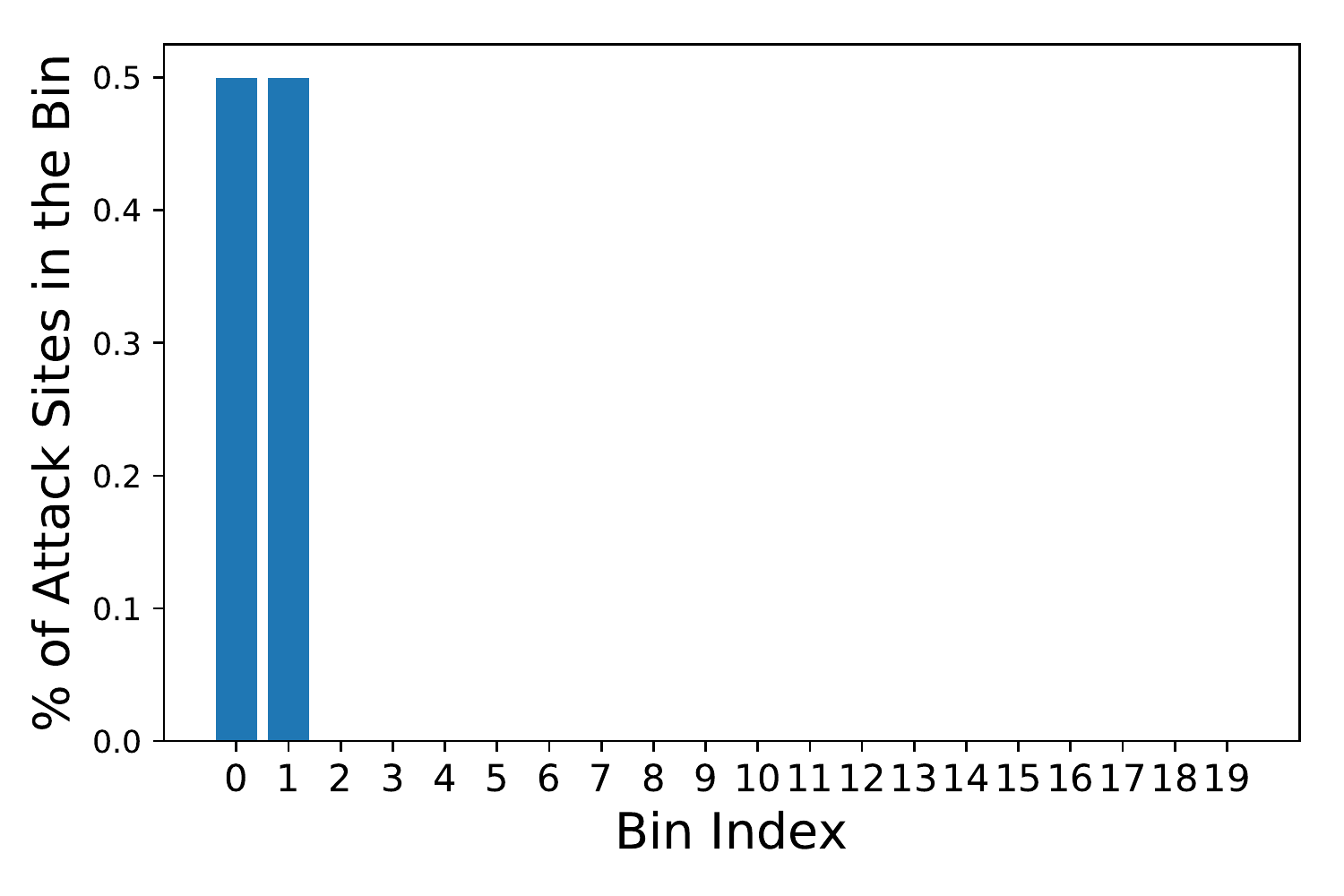}
    \includegraphics[scale=0.3]{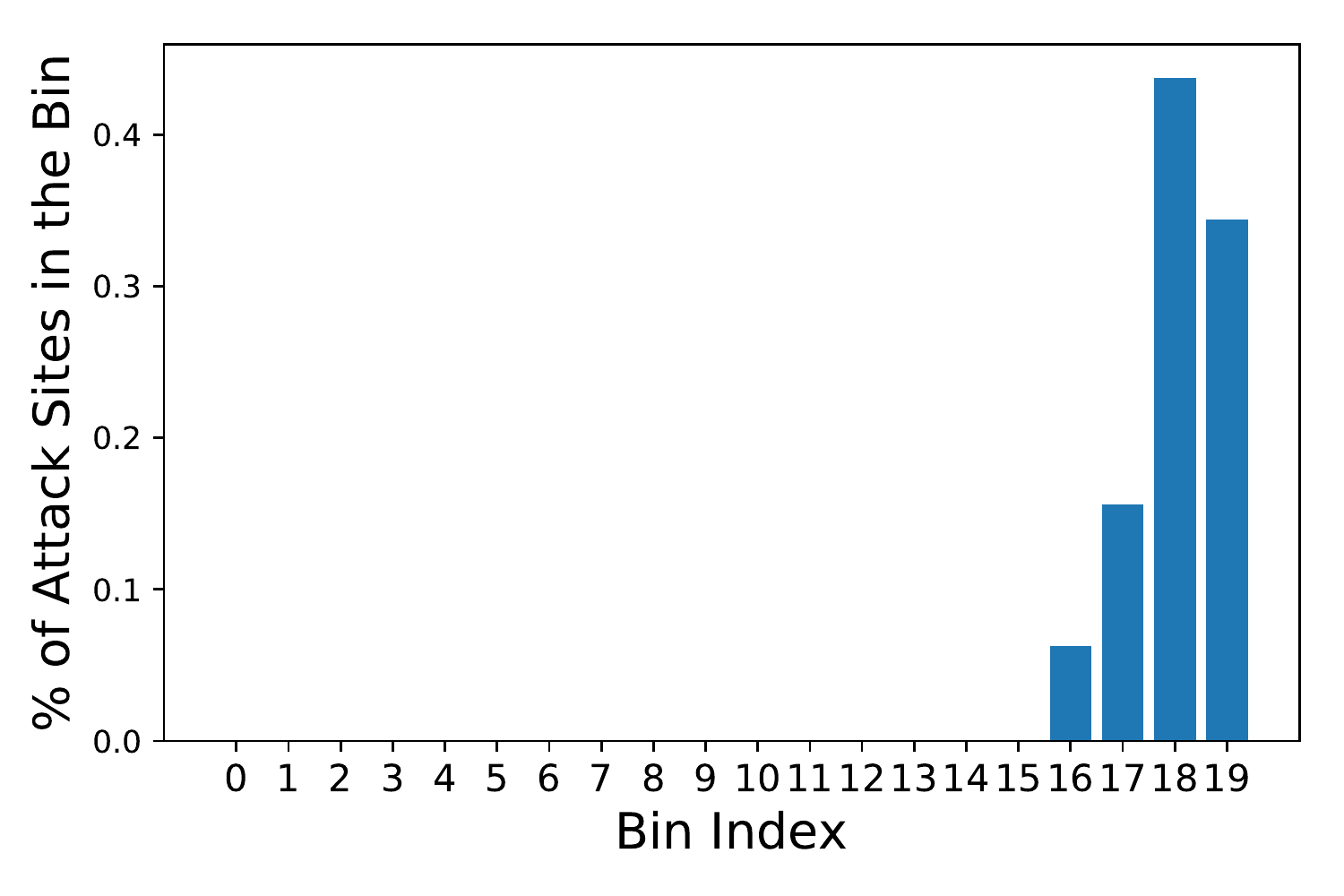}\\ 
    \includegraphics[scale=0.3]{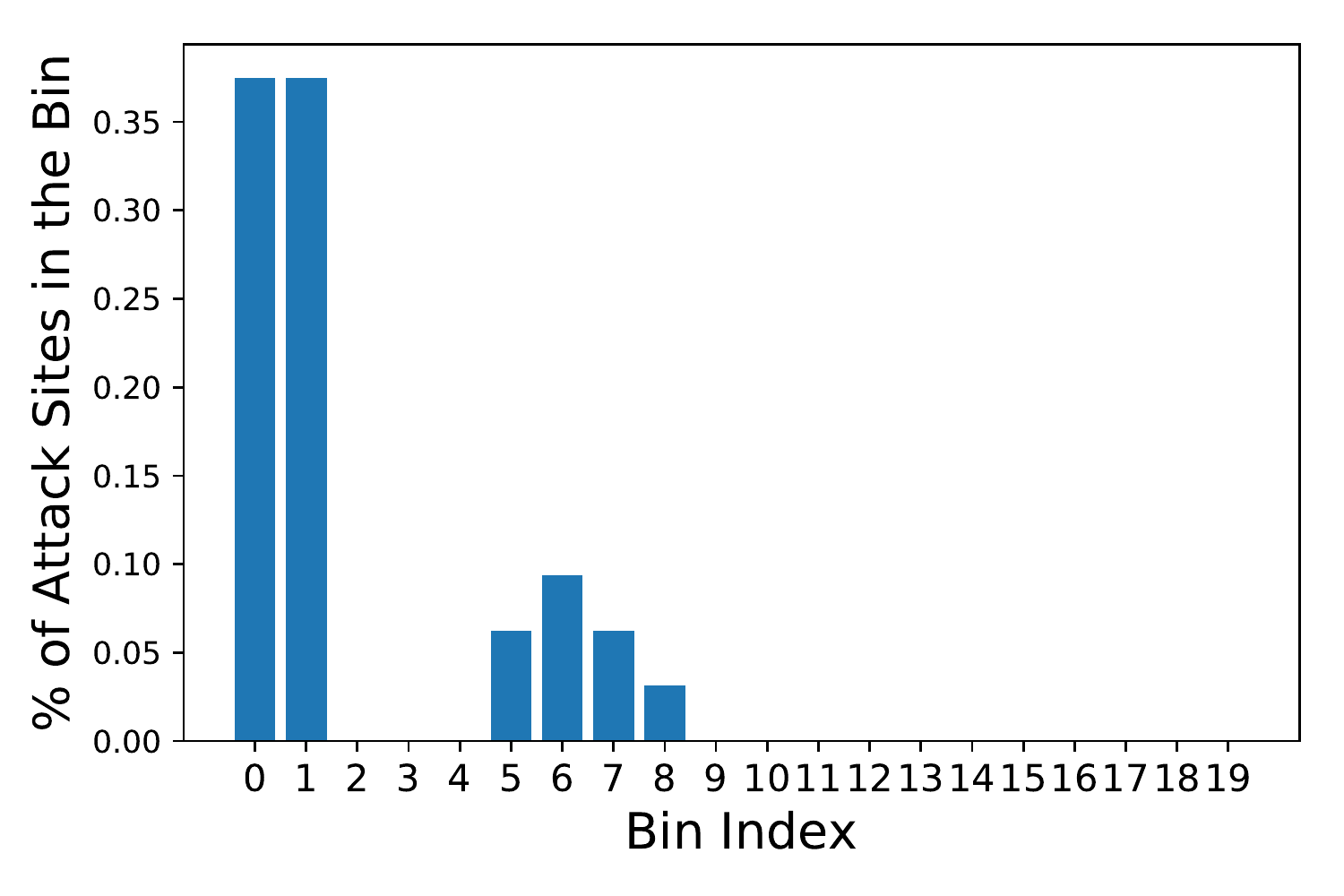}
    \includegraphics[scale=0.3]{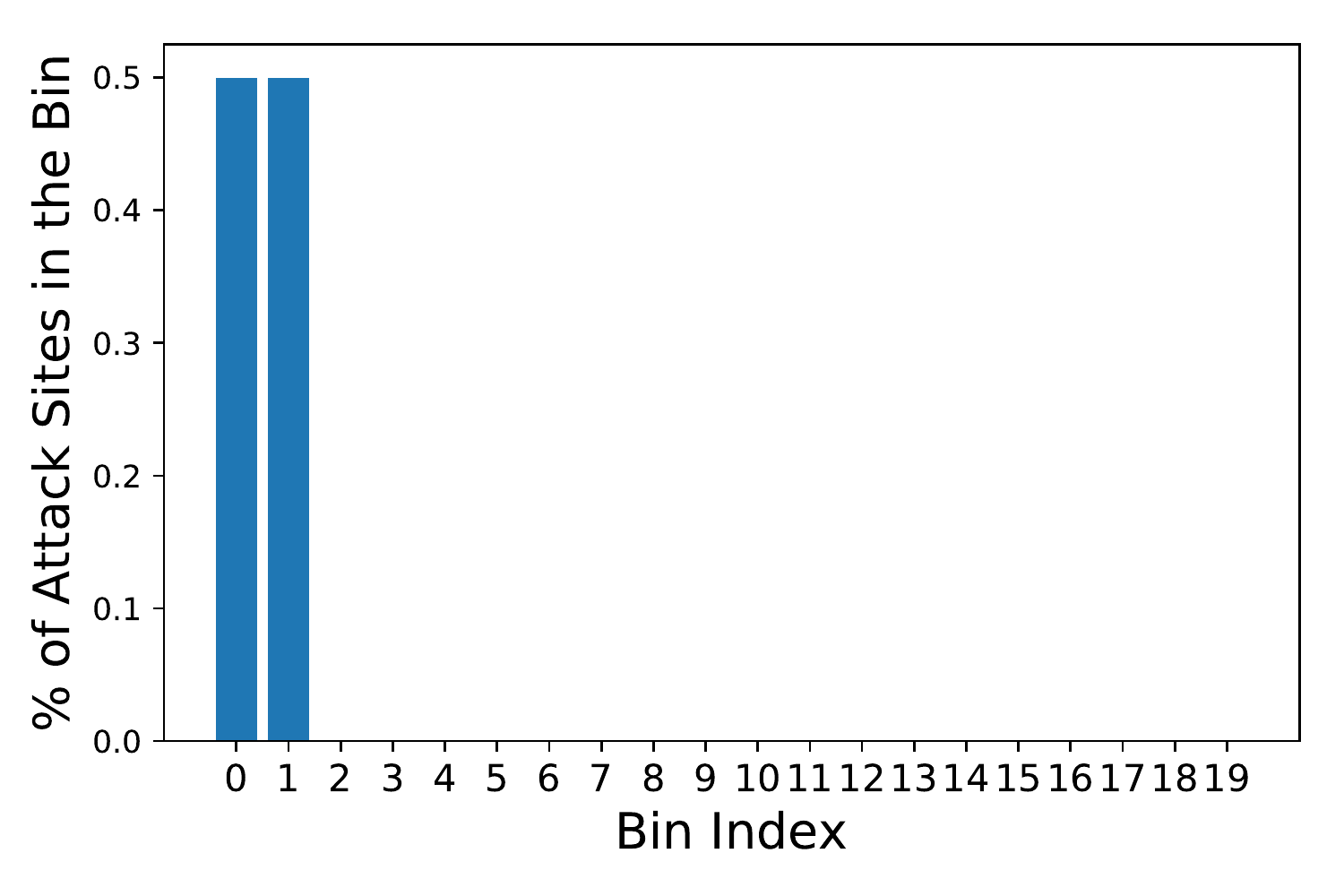}
    \includegraphics[scale=0.3]{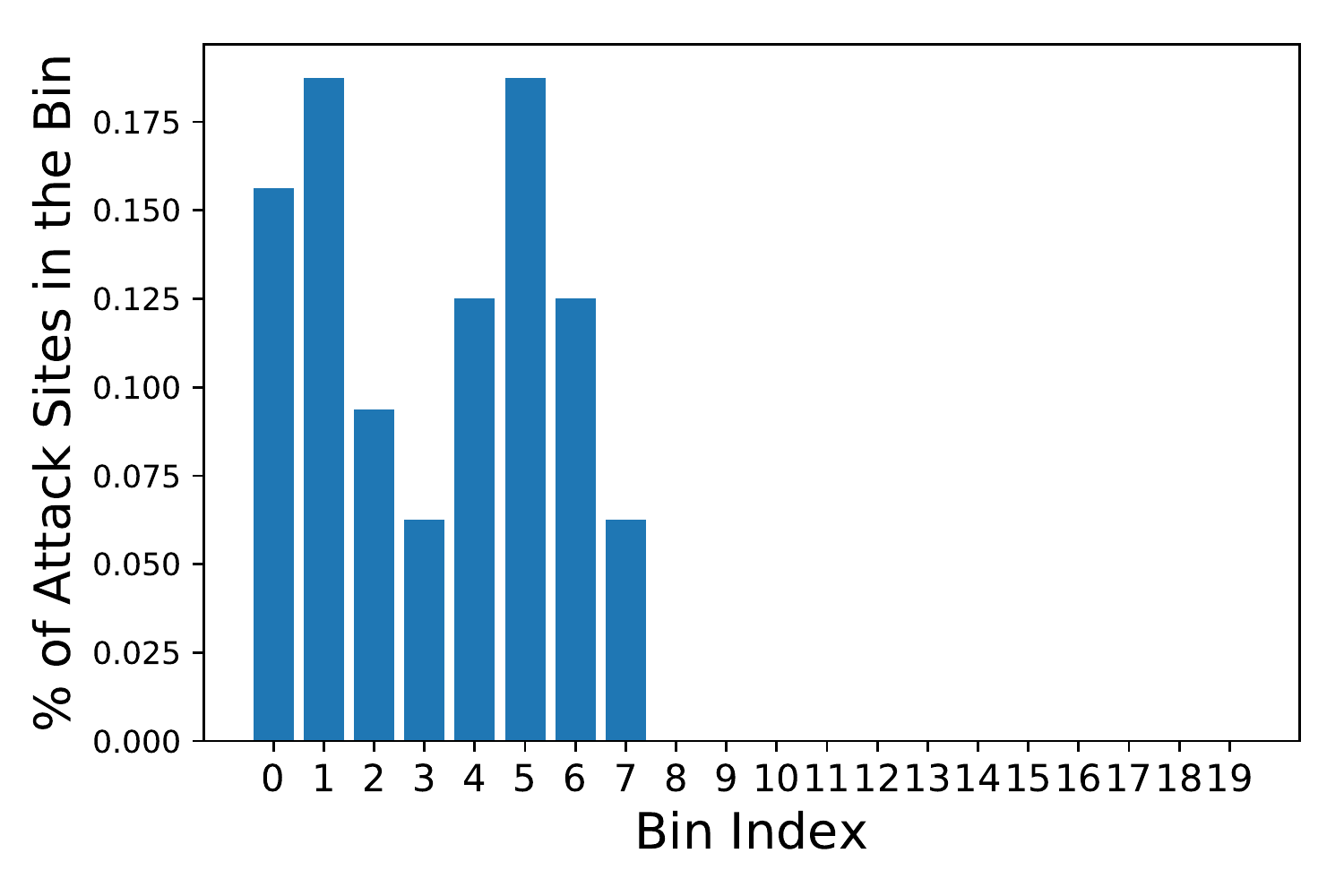}
    \captionof{figure}{Results for MNIST, Interval \textbf{Top row:} Semi-online, \textbf{Bottom row:} Full-online. \textbf{Left to right:} Slow Decay, Fast Decay, Constant}
    \label{fig:MNISTpos}
\end{center}

\subsubsection{Fashion MNIST Sandals v.s. Boots}
\begin{center}
    \centering
    \includegraphics[scale=0.3]{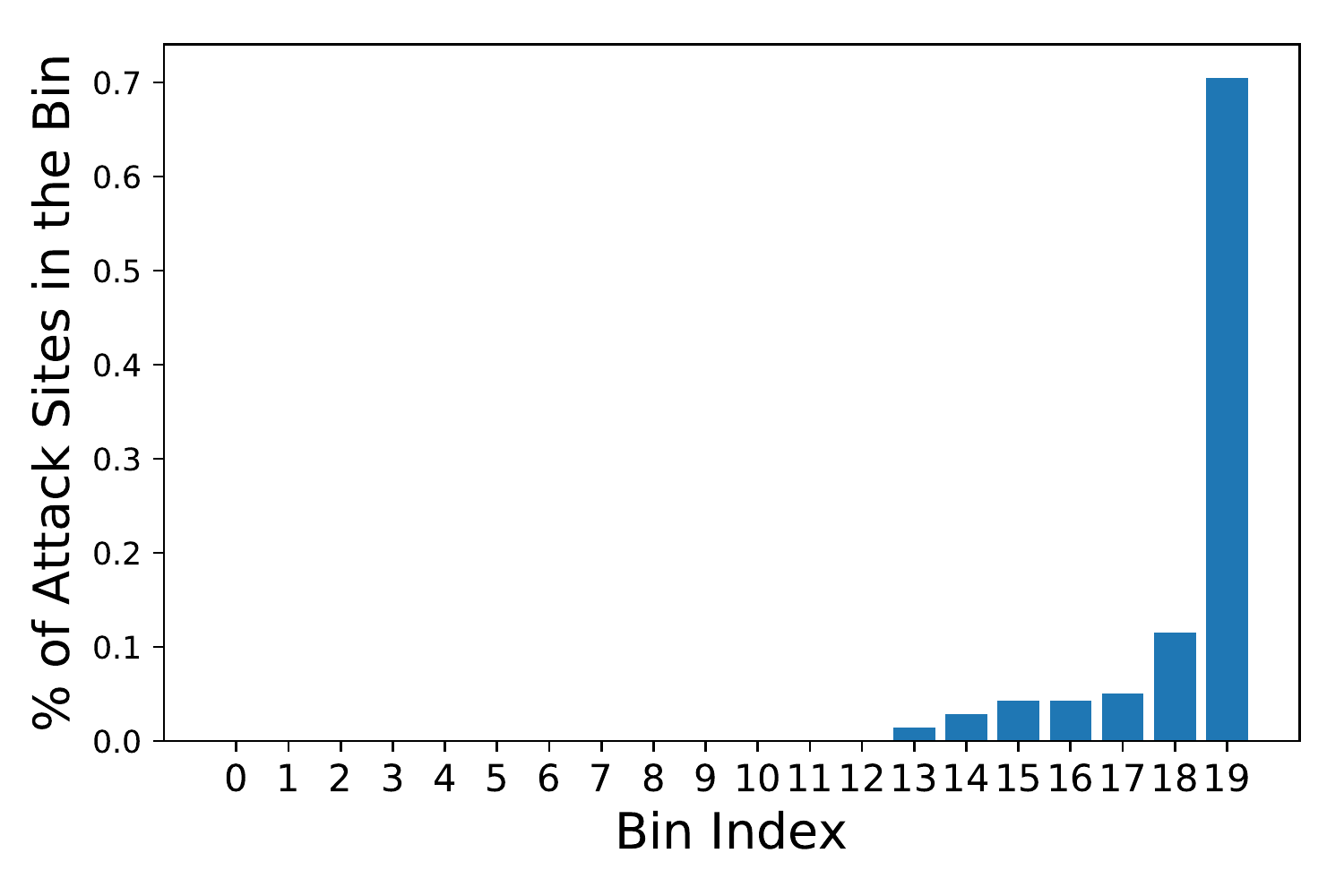}
    \includegraphics[scale=0.3]{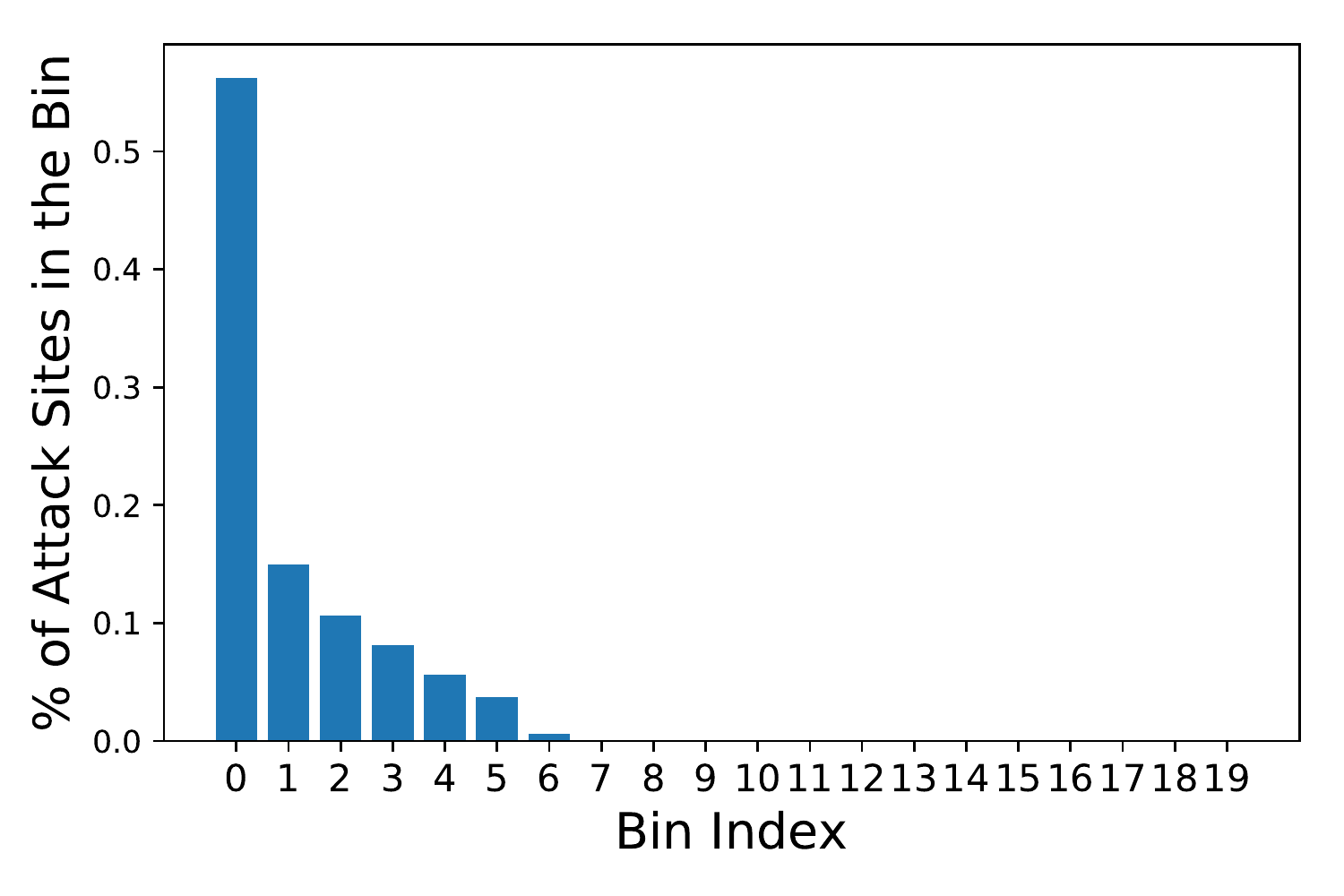} 
    \includegraphics[scale=0.3]{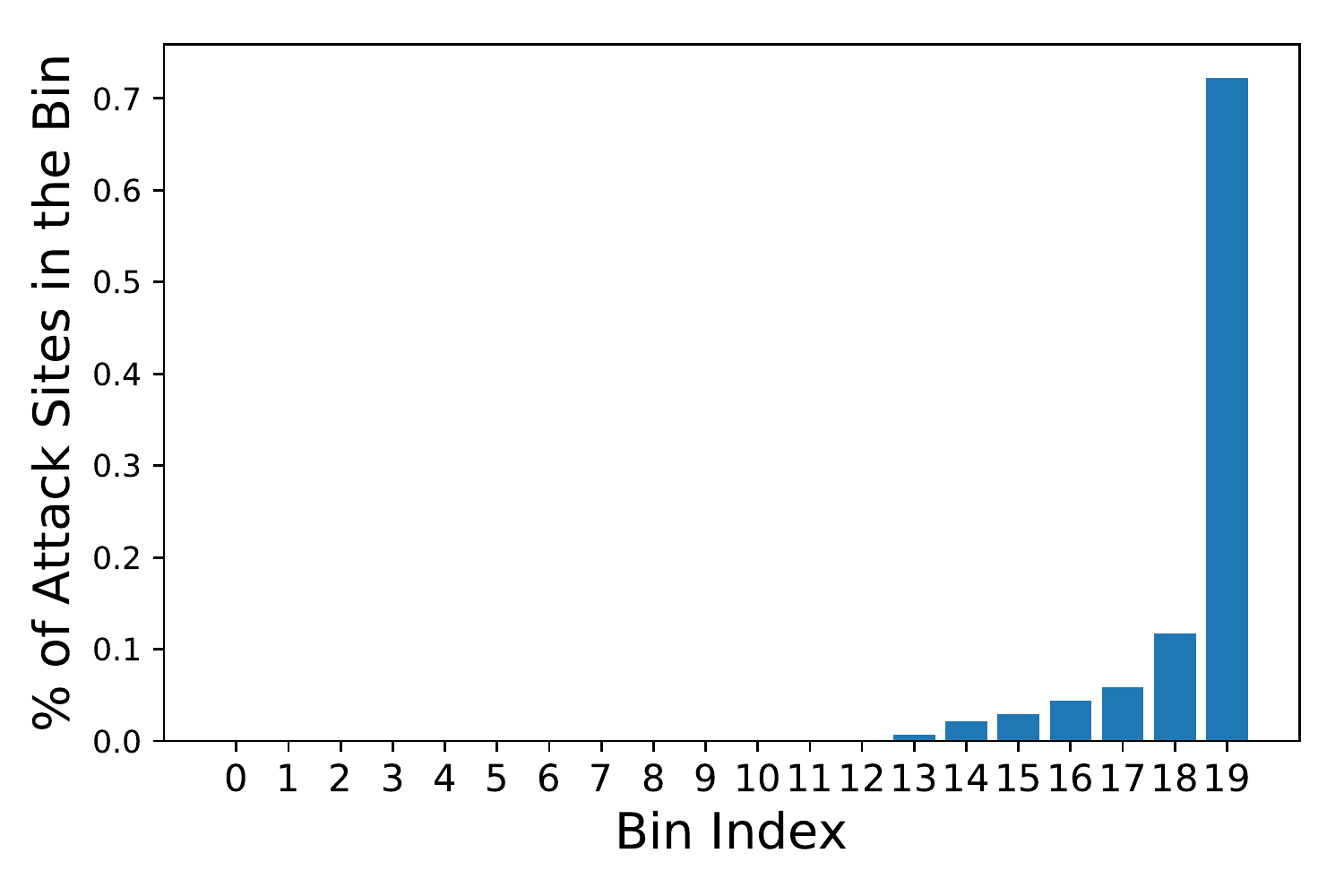}\\
    \includegraphics[scale=0.3]{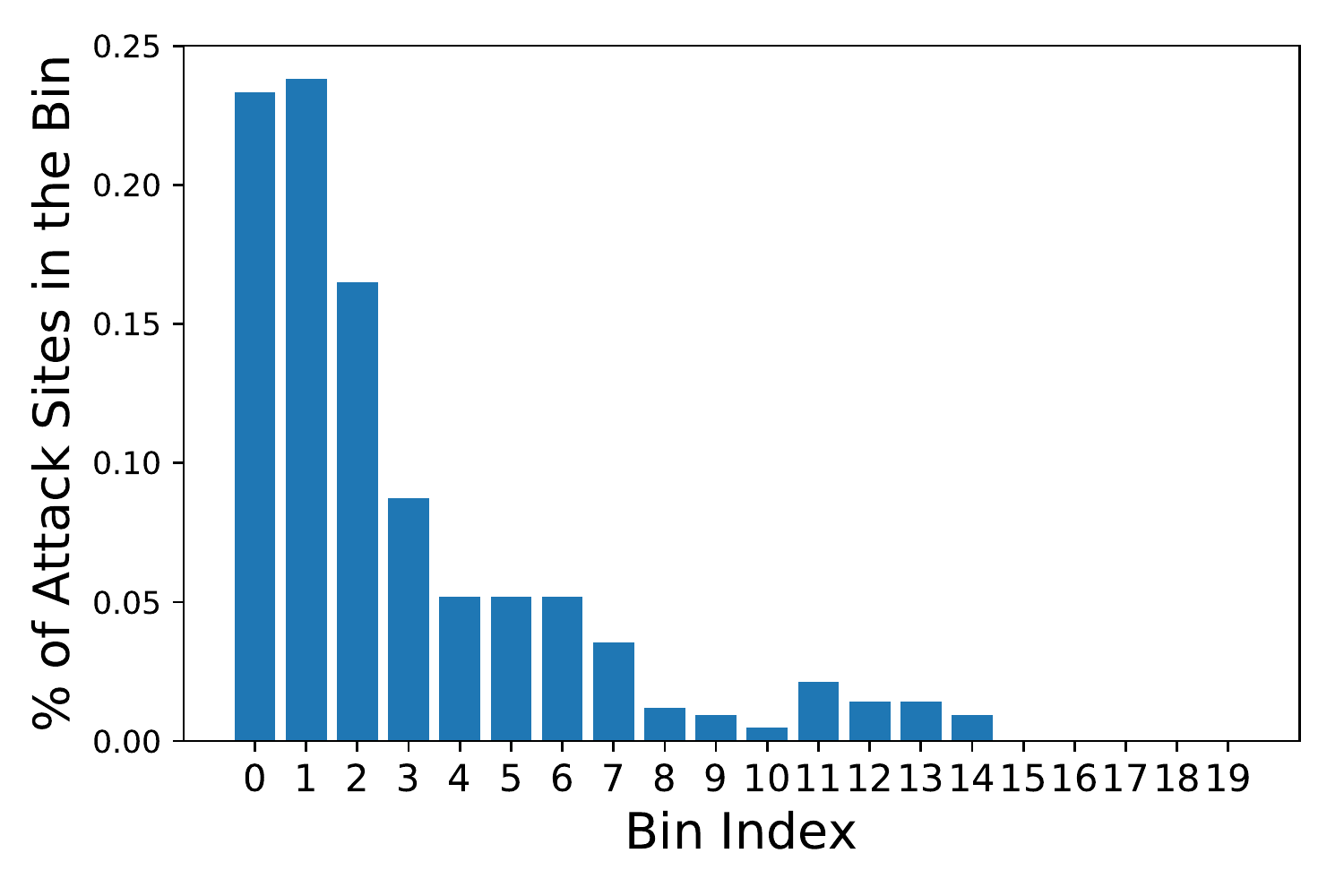}
    \includegraphics[scale=0.3]{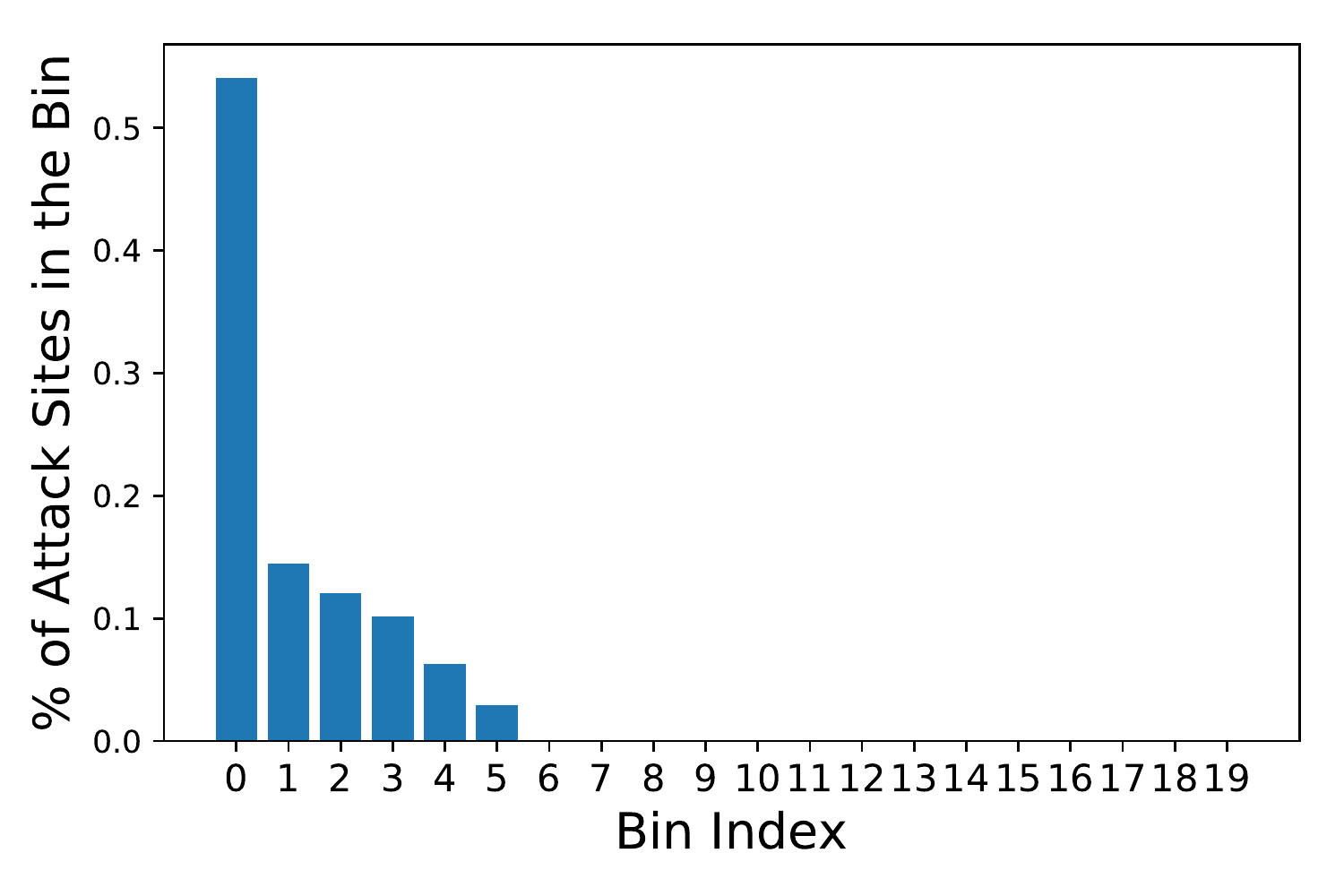} 
    \includegraphics[scale=0.3]{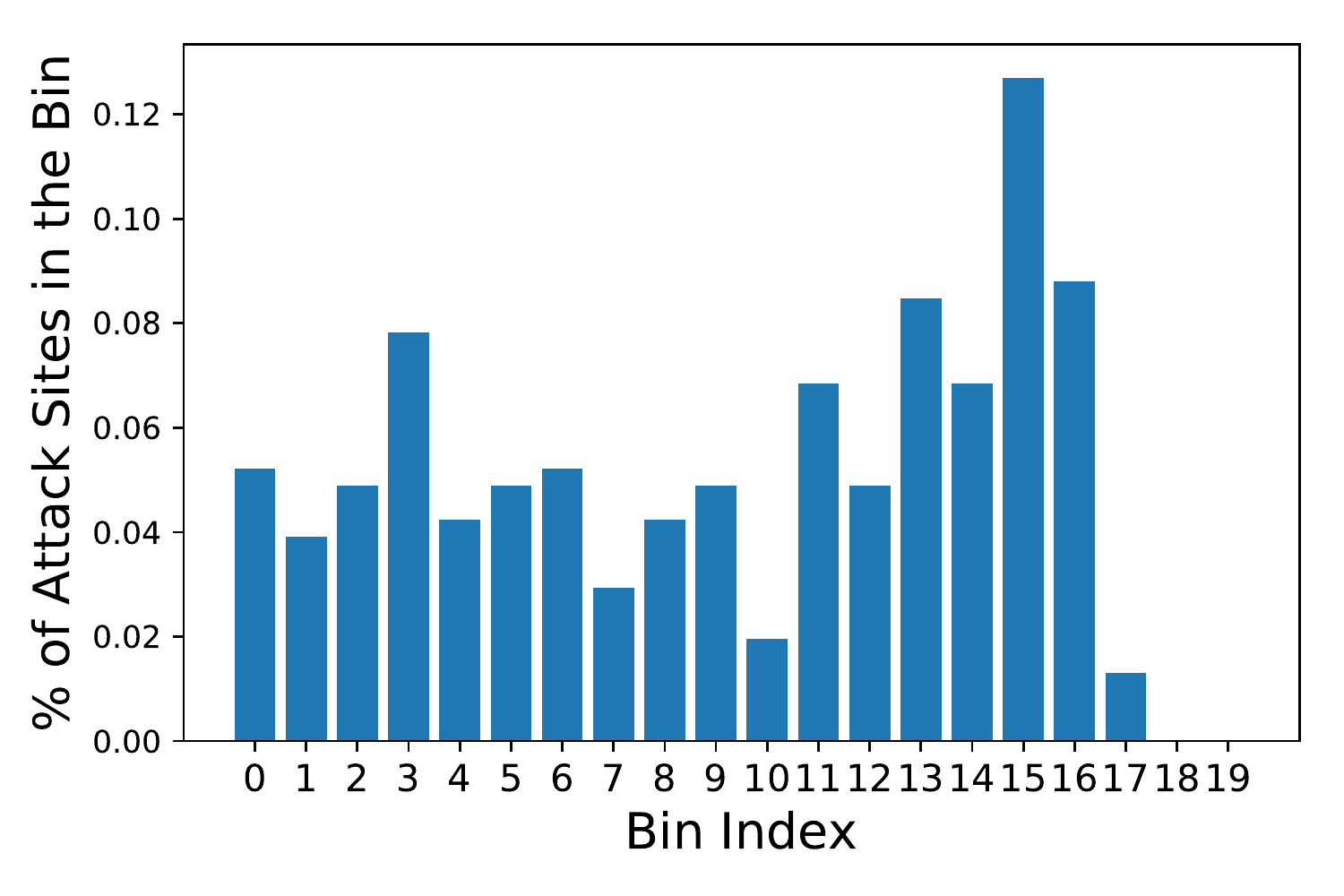}
    \captionof{figure}{Fashion MNIST Sandals v.s. Boots, Incremental \textbf{Top row:} Semi-online, \textbf{Bottom row:} Full-online. \textbf{Left to right:} Slow Decay, Fast Decay, Constant}
    \label{fig:fashion_MNISTpos}
\end{center}

\begin{center}
    \centering
    \includegraphics[scale=0.3]{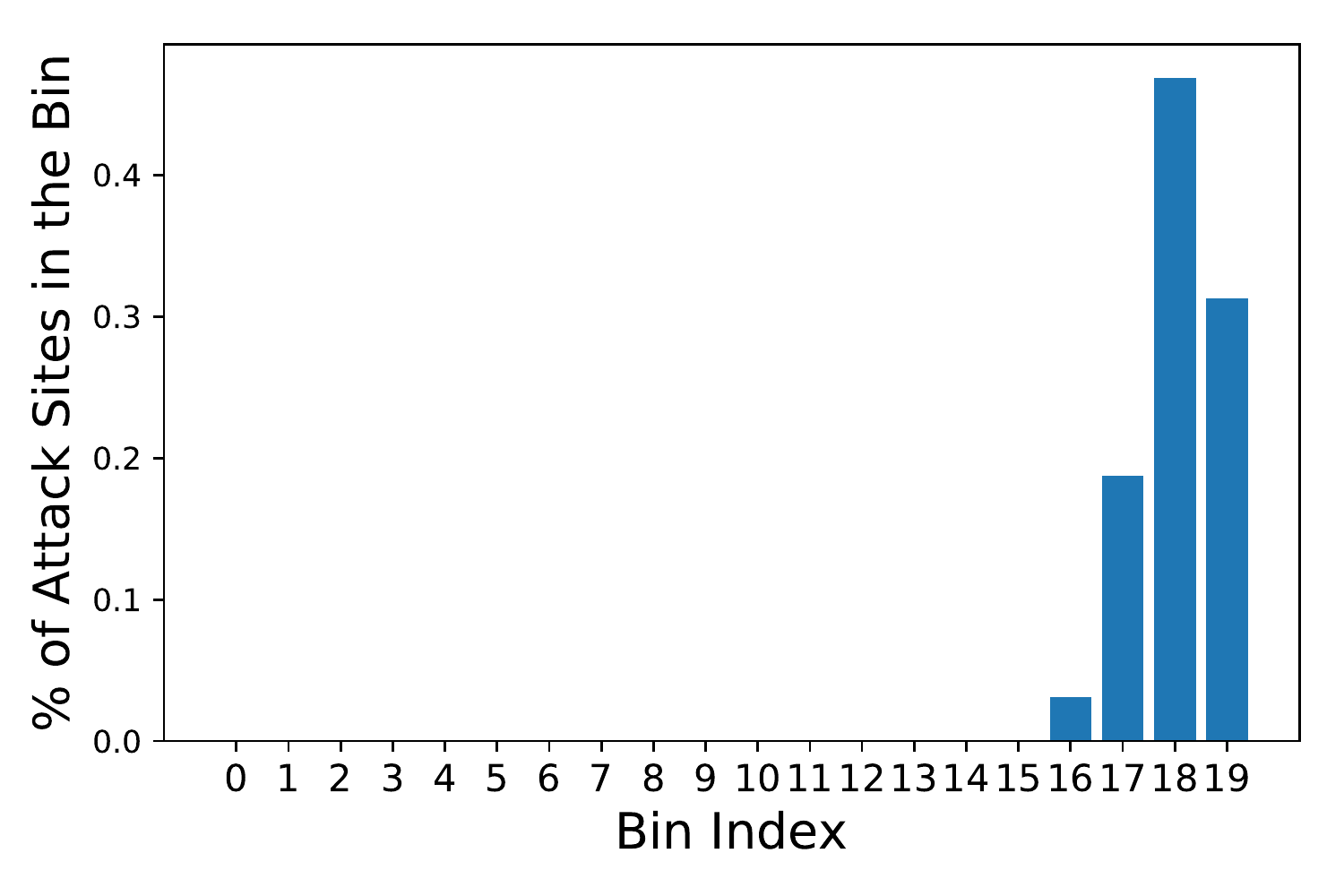}
    \includegraphics[scale=0.3]{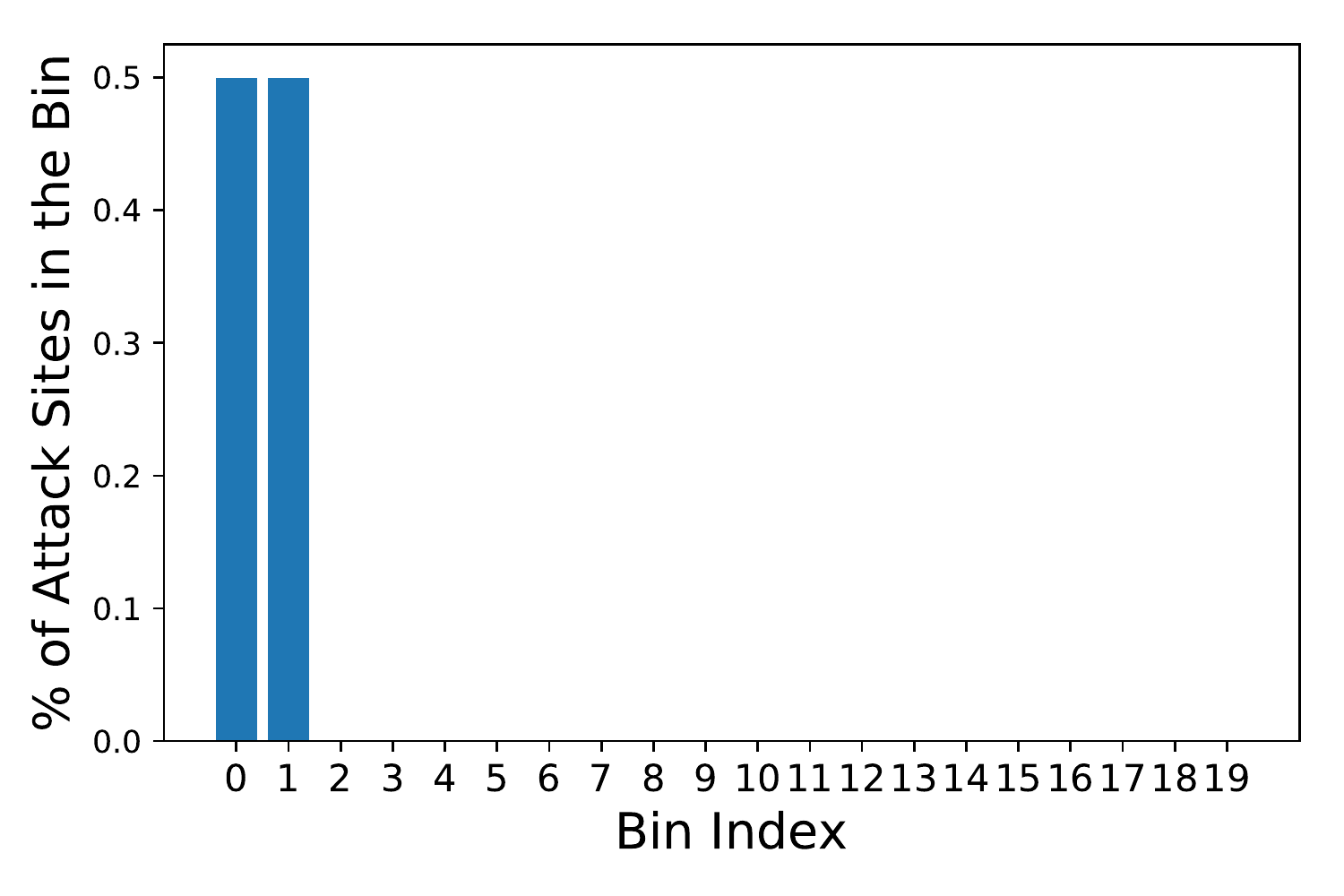} 
    \includegraphics[scale=0.3]{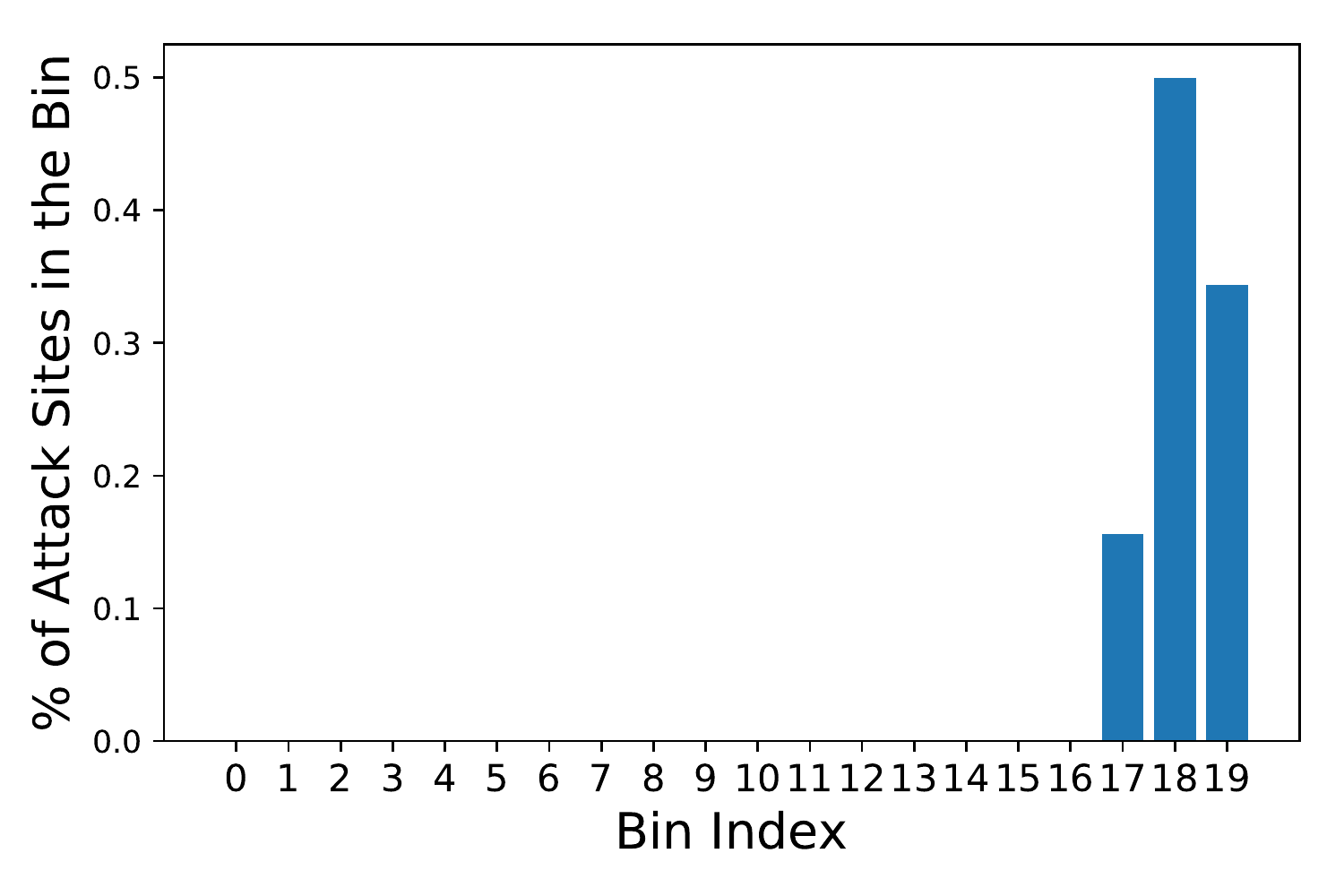}\\
    \includegraphics[scale=0.3]{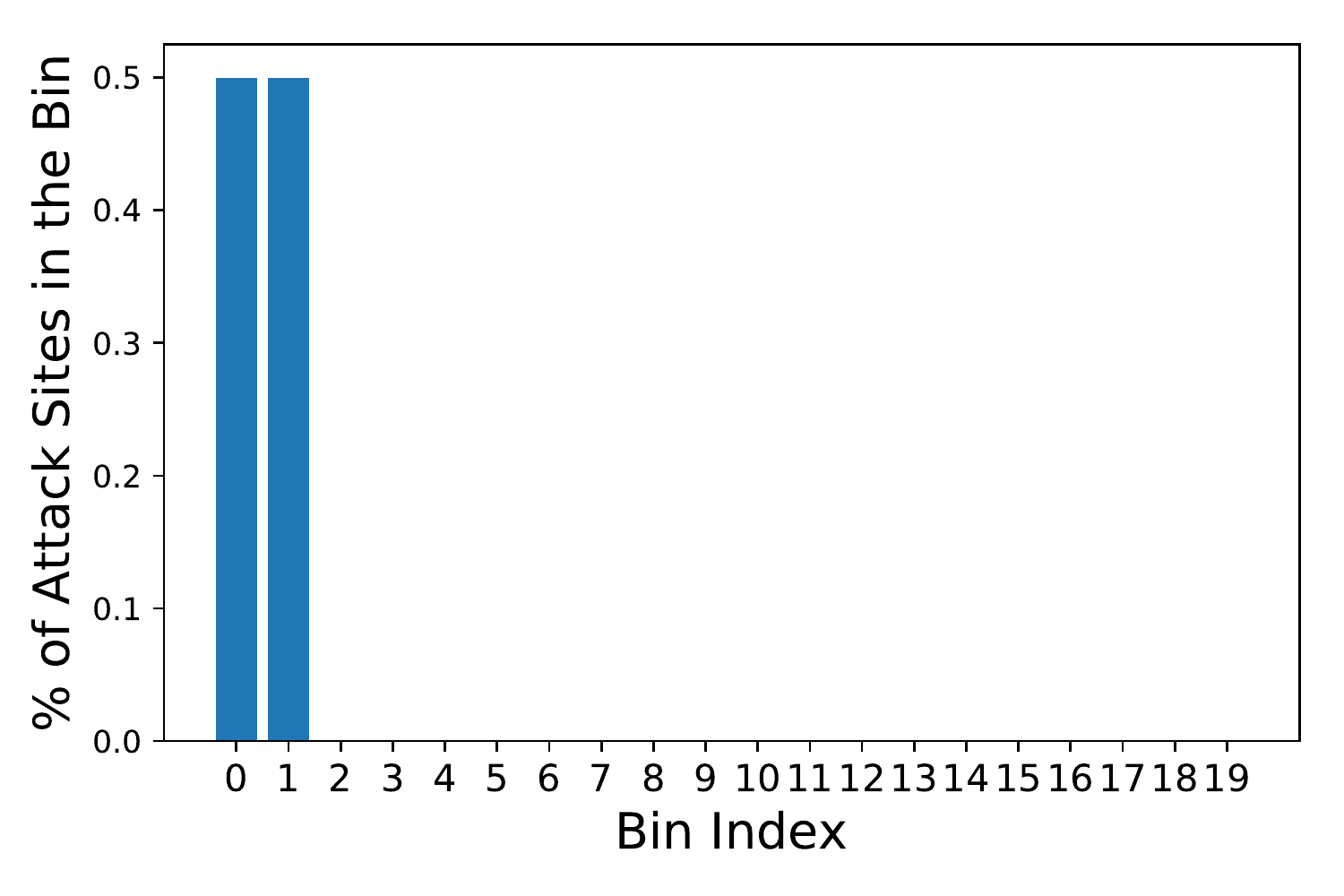}
    \includegraphics[scale=0.3]{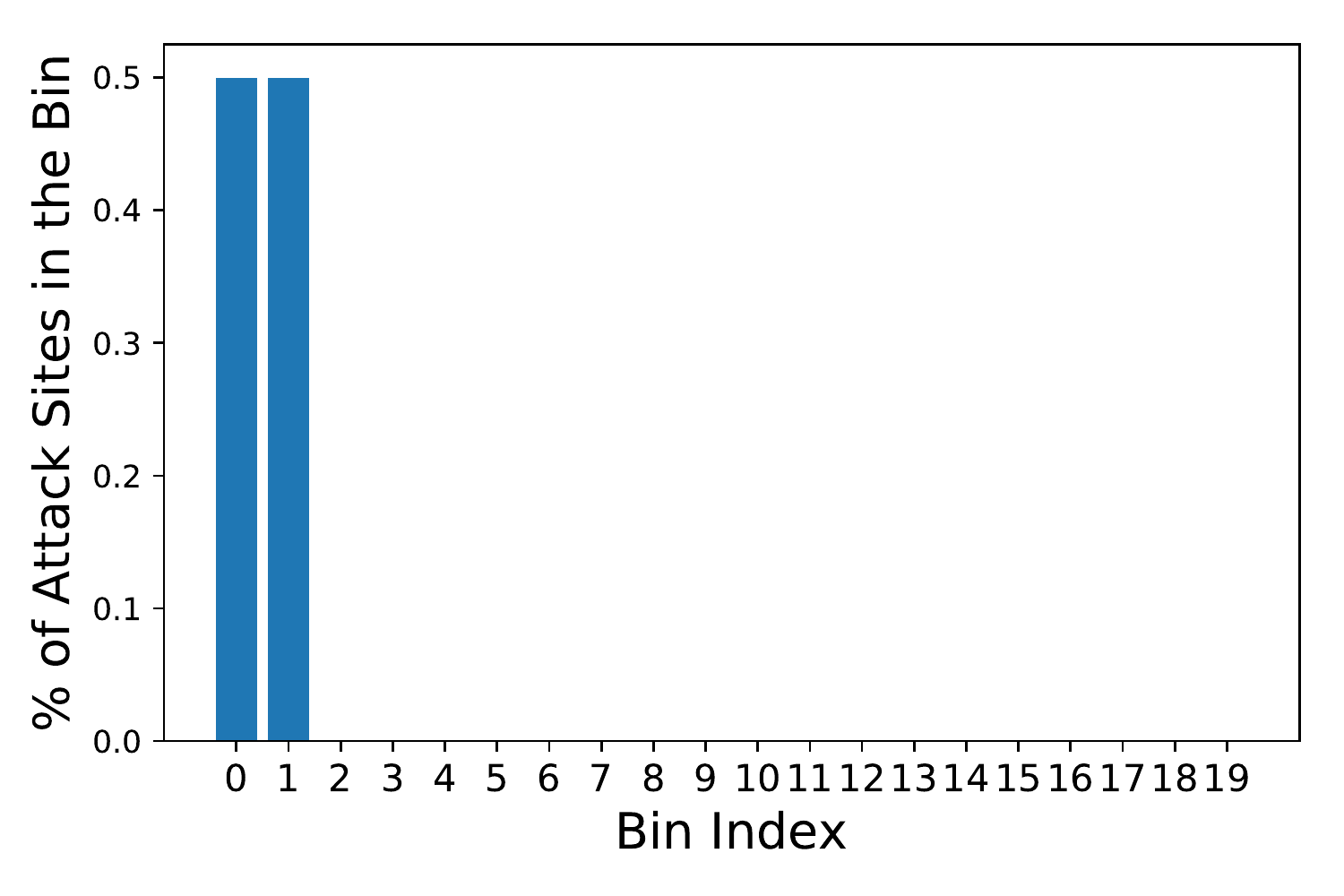} 
    \includegraphics[scale=0.3]{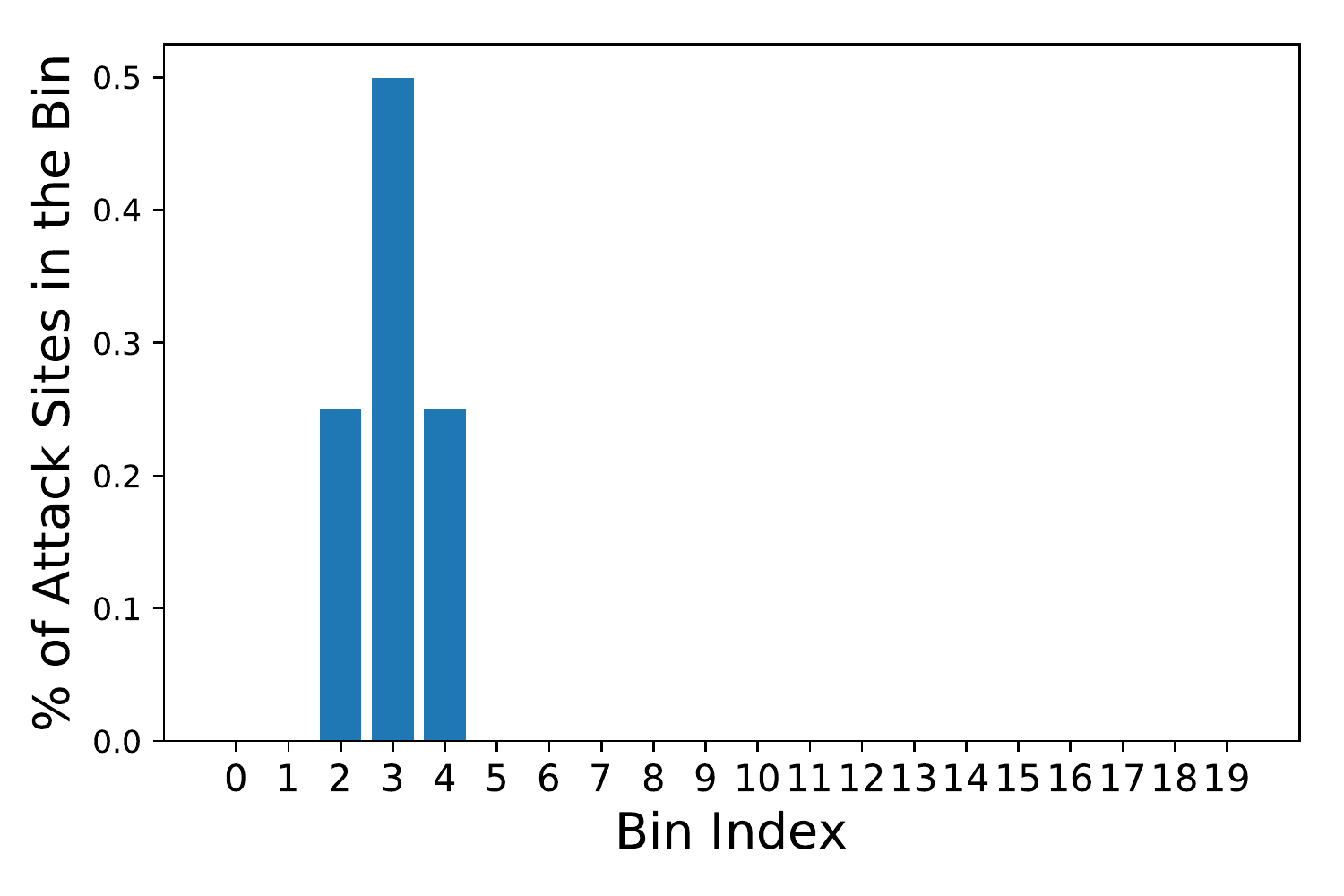}
    \captionof{figure}{Fashion MNIST Sandals v.s. Boots, Interval \textbf{Top row:} Semi-online, \textbf{Bottom row:} Full-online. \textbf{Left to right:} Slow Decay, Fast Decay, Constant}
    \label{fig:fashion_MNISTpos}
\end{center}

\subsubsection{UCI Spambase}
\begin{center}
    \centering
    \includegraphics[scale=0.3]{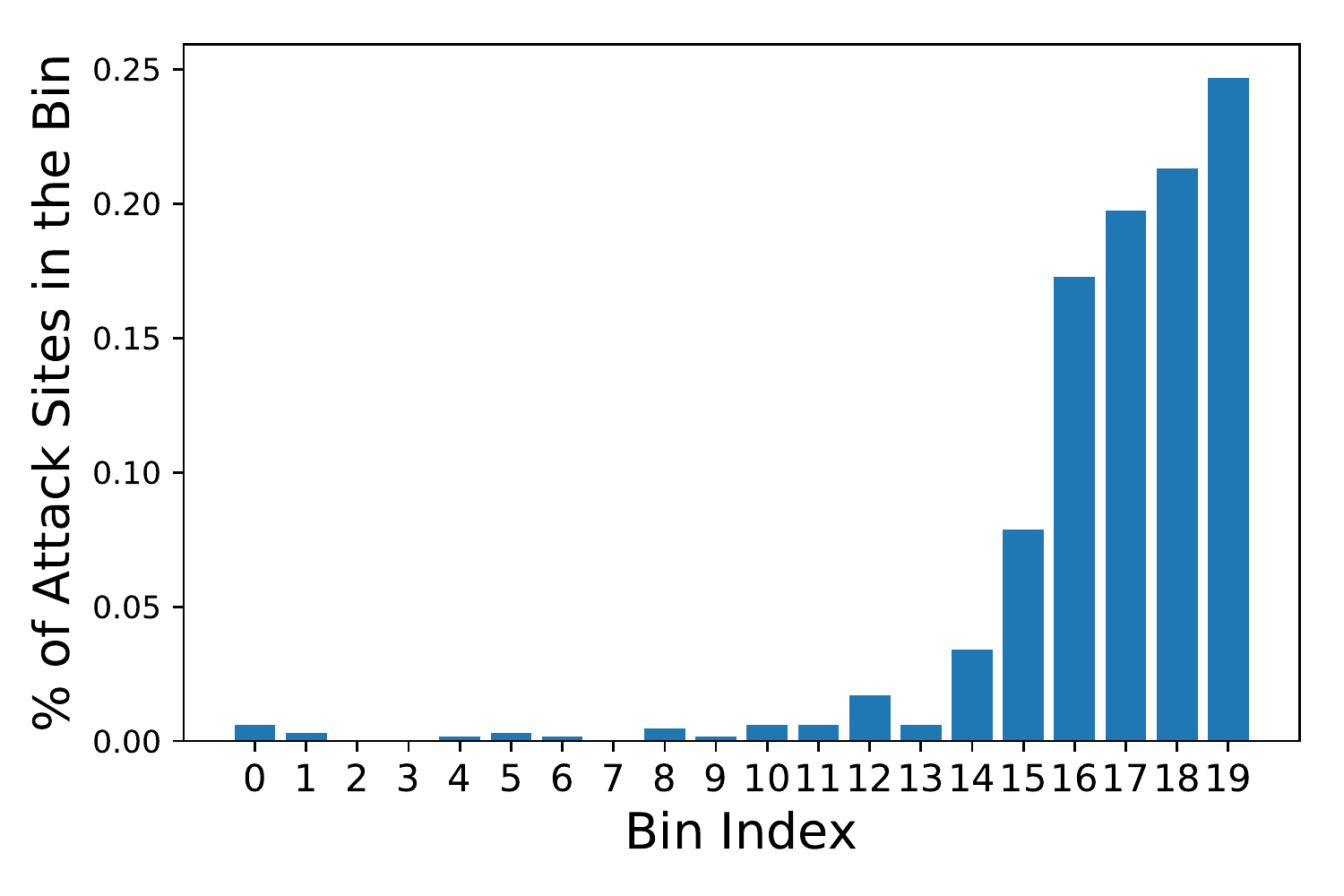}
    \includegraphics[scale=0.3]{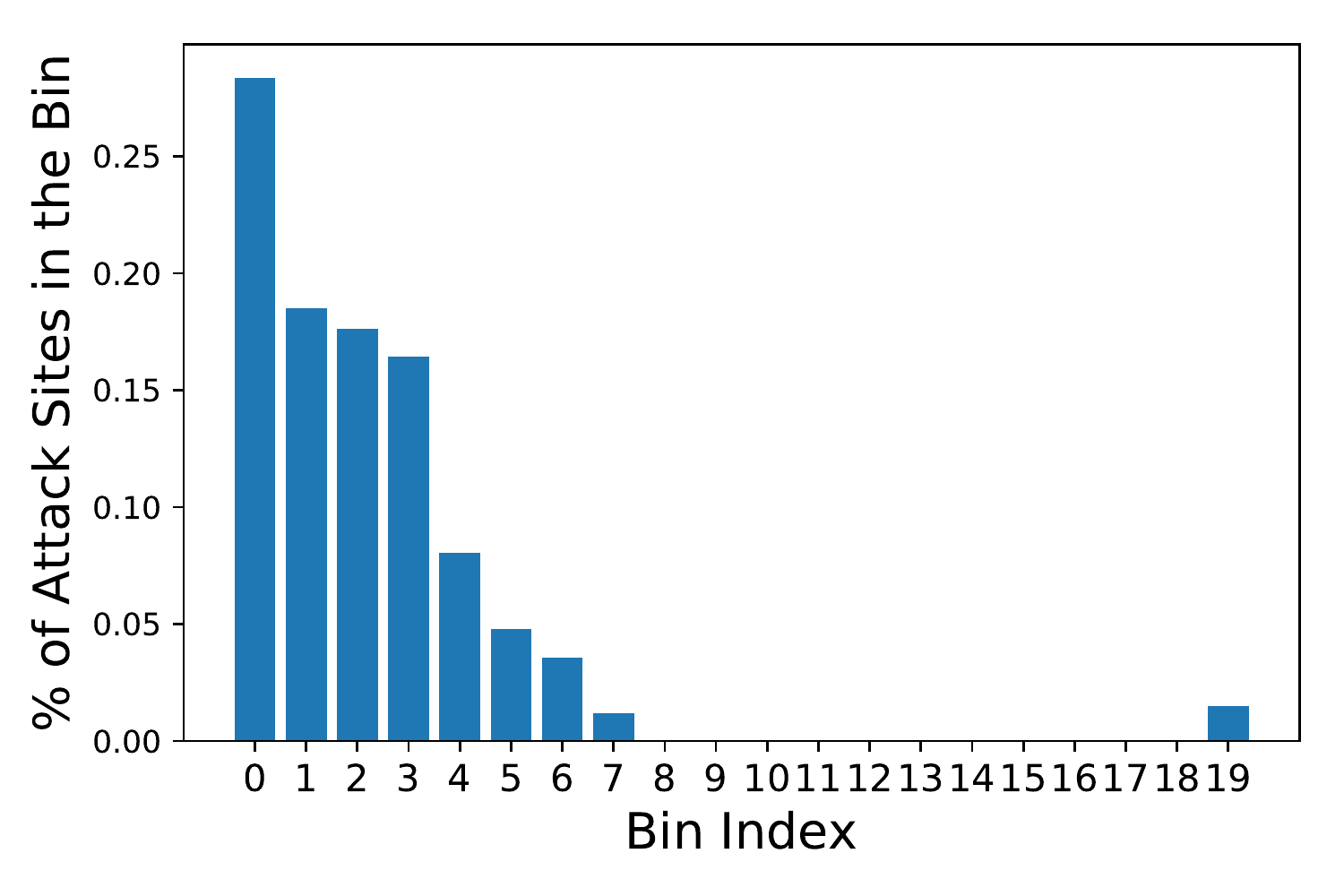} 
    \includegraphics[scale=0.3]{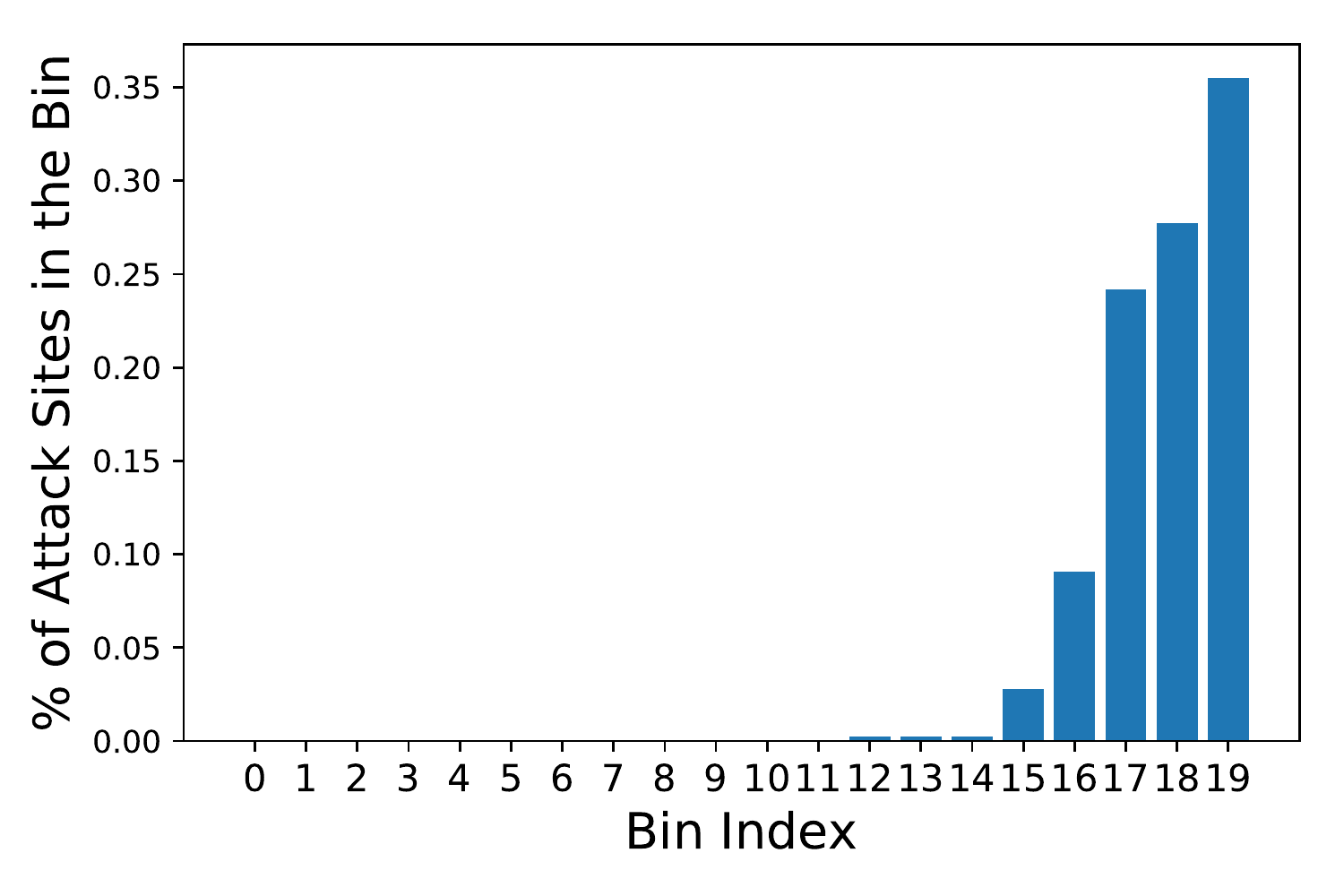}\\
    \includegraphics[scale=0.3]{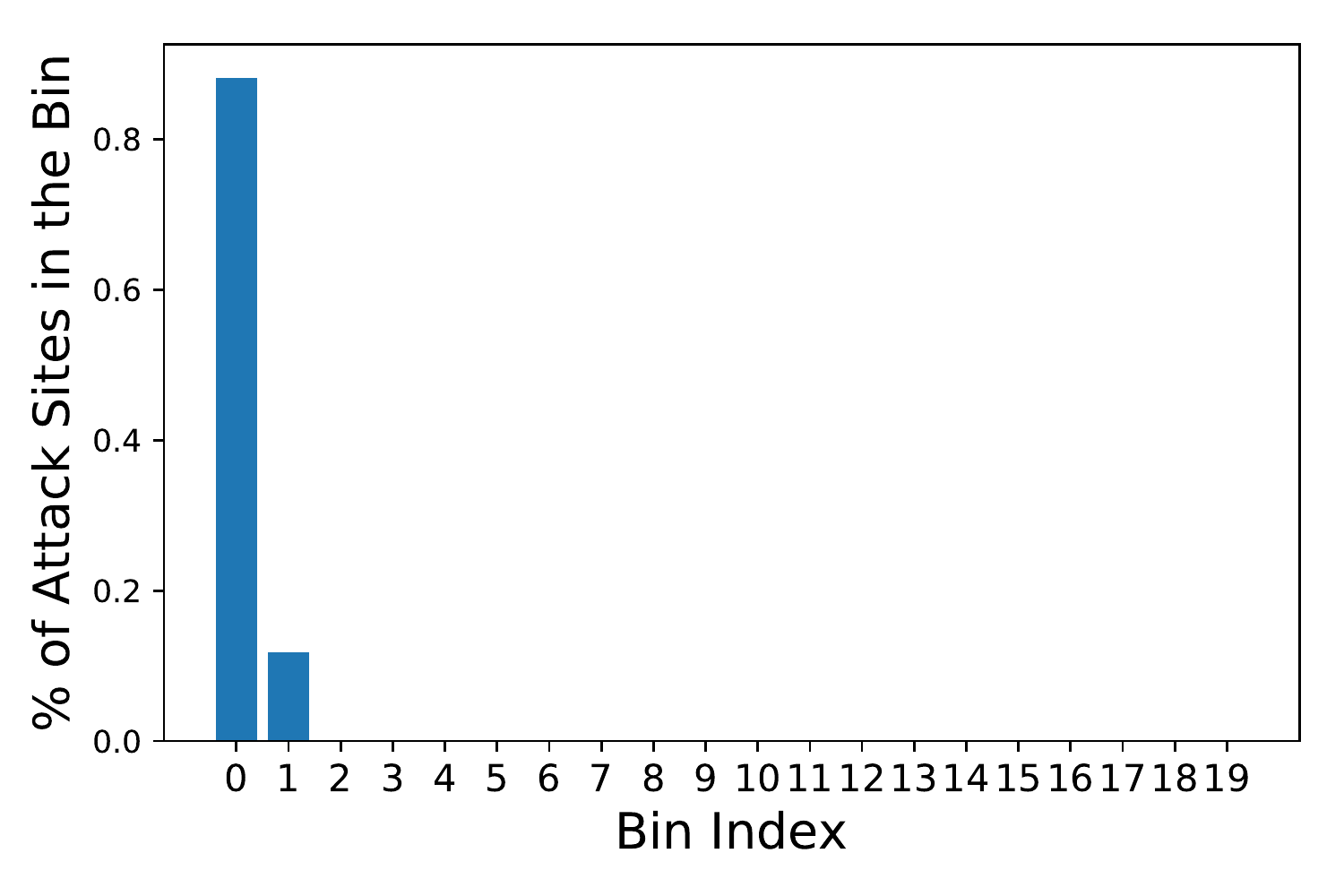}
    \includegraphics[scale=0.3]{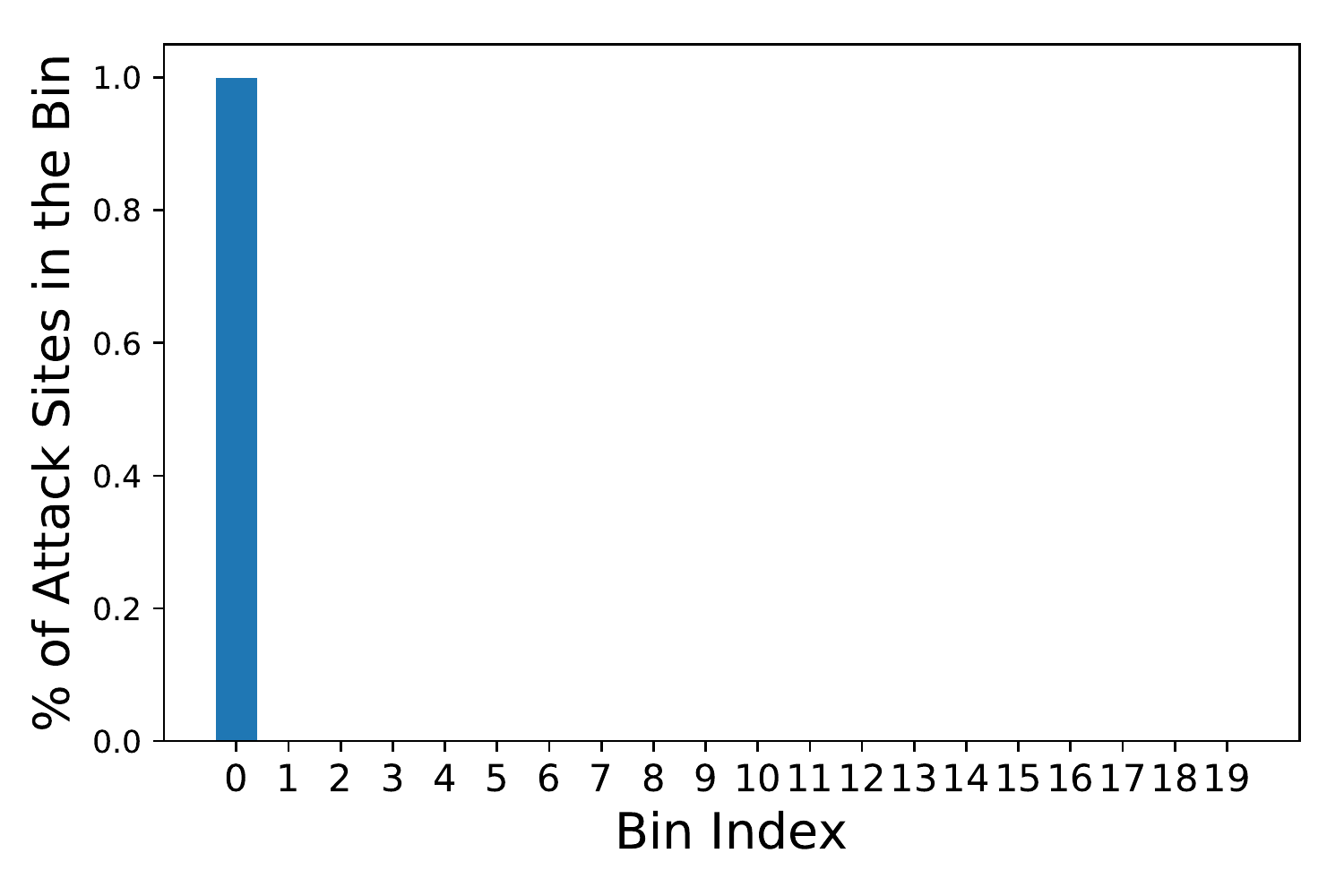} 
    \includegraphics[scale=0.3]{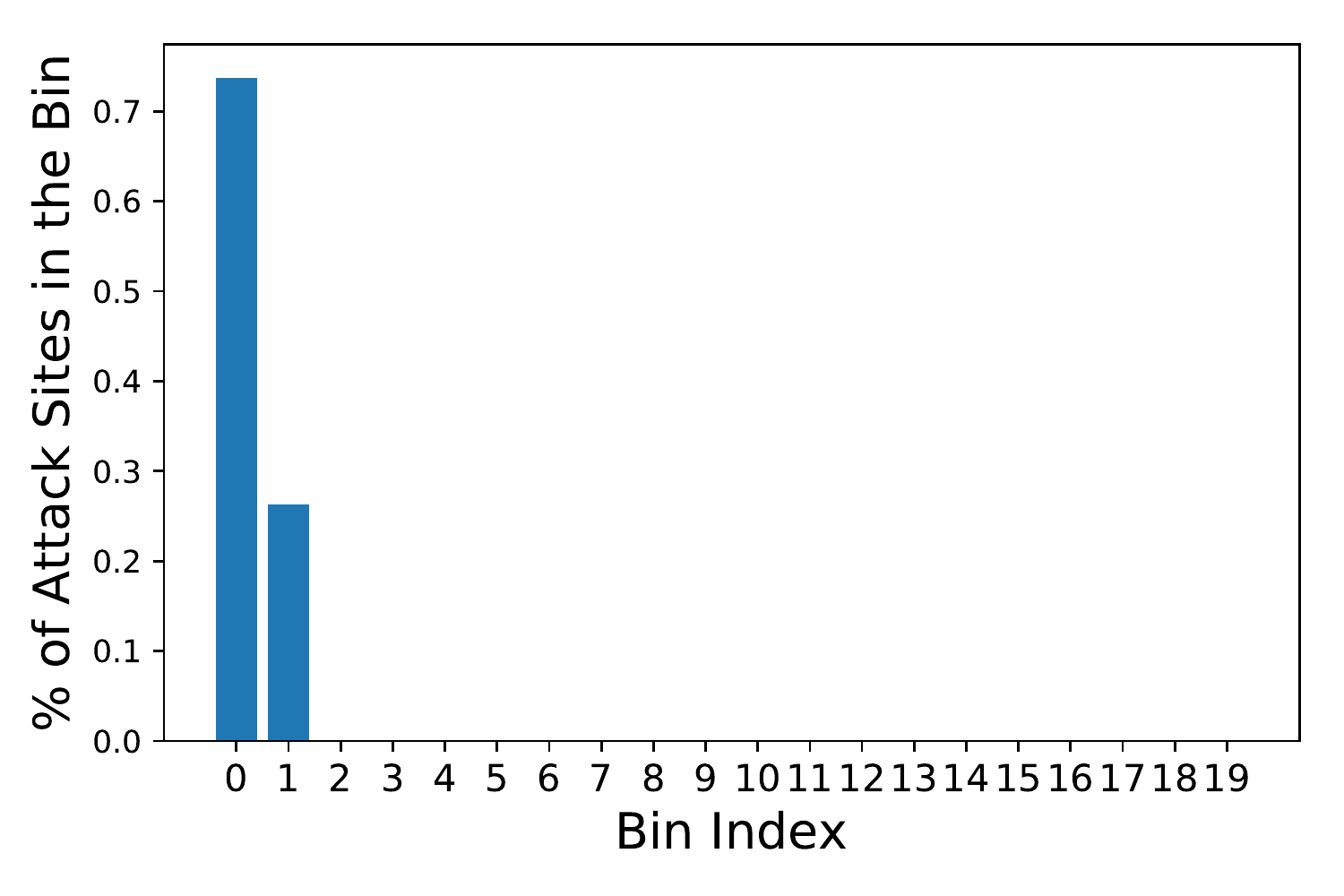}
    \captionof{figure}{UCI Spam Dataset, Incremental \textbf{Top row:} Semi-online, \textbf{Bottom row:} Full-online. \textbf{Left to right:} Slow Decay, Fast Decay, Constant}
    \label{fig:spampos}
\end{center}

\begin{center}
    \centering
    \includegraphics[scale=0.3]{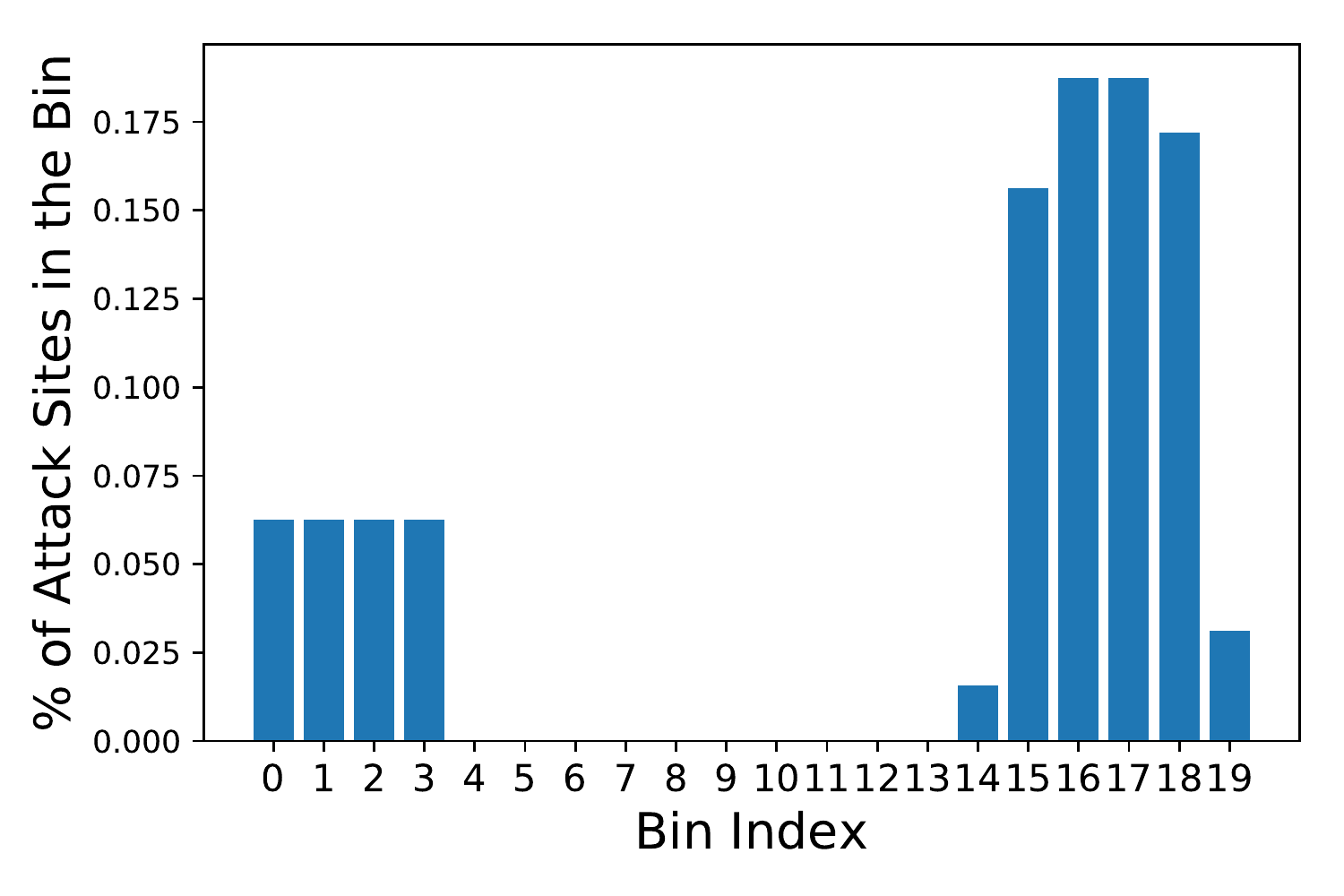}
    \includegraphics[scale=0.3]{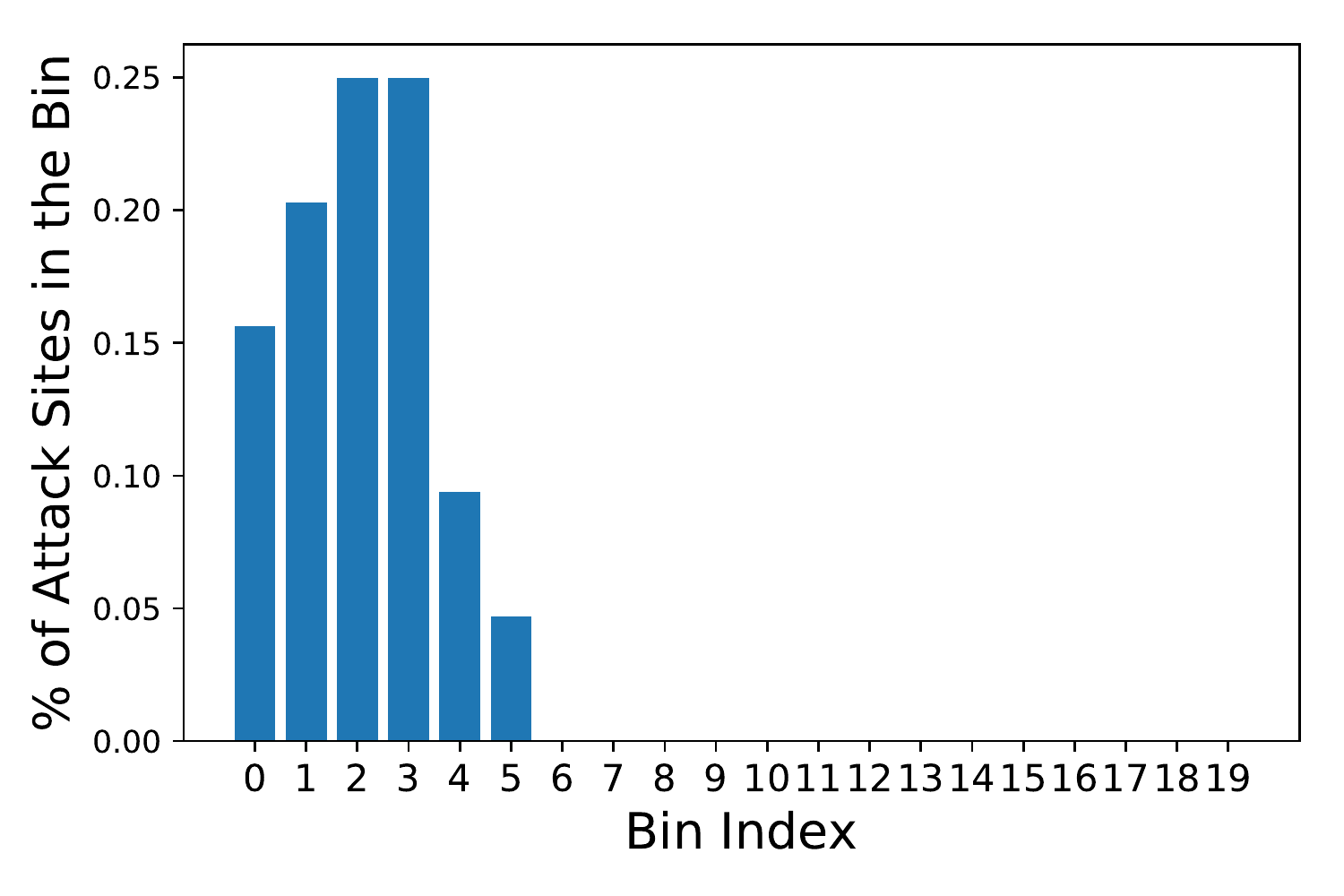} 
    \includegraphics[scale=0.3]{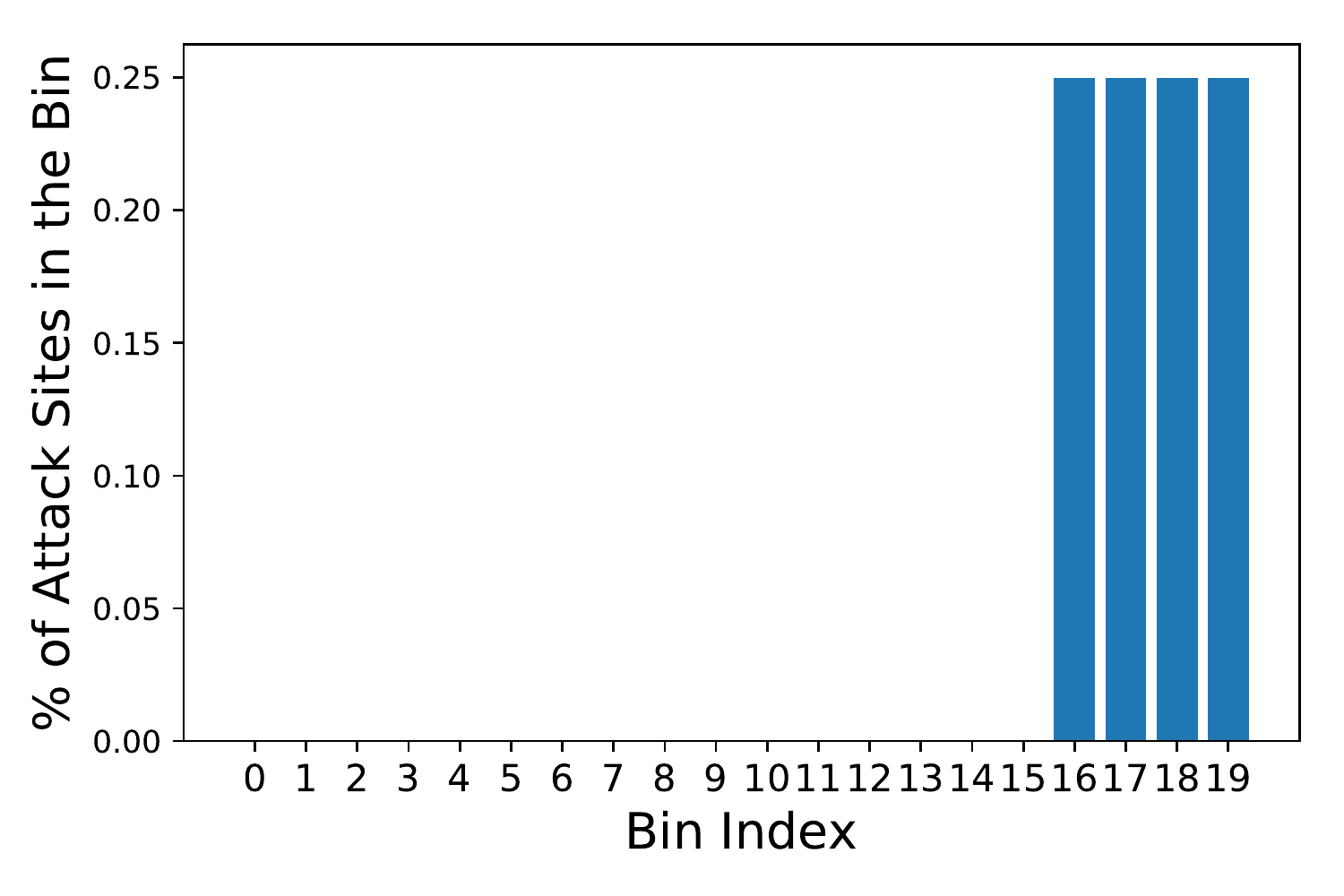}\\
    \includegraphics[scale=0.3]{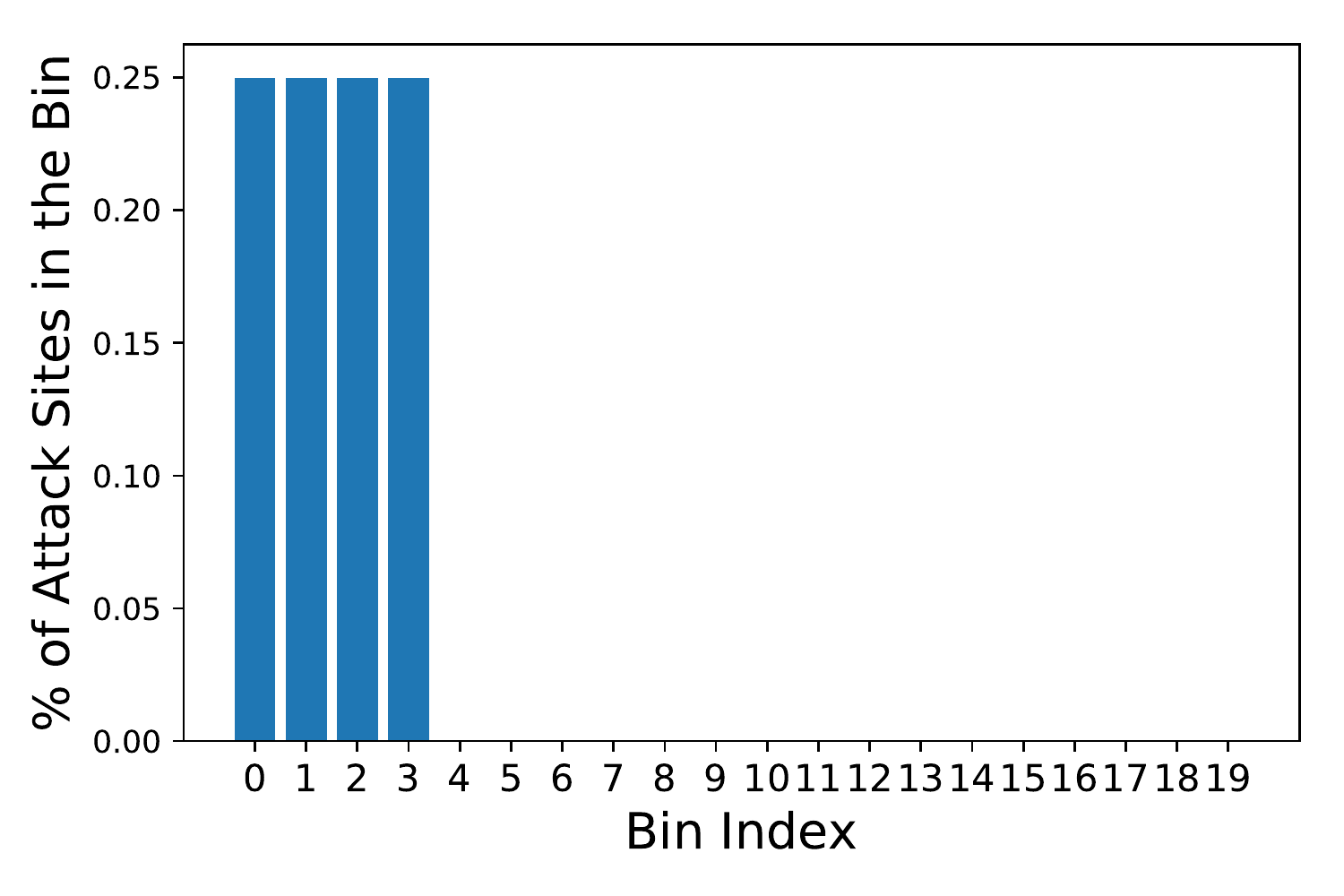}
    \includegraphics[scale=0.3]{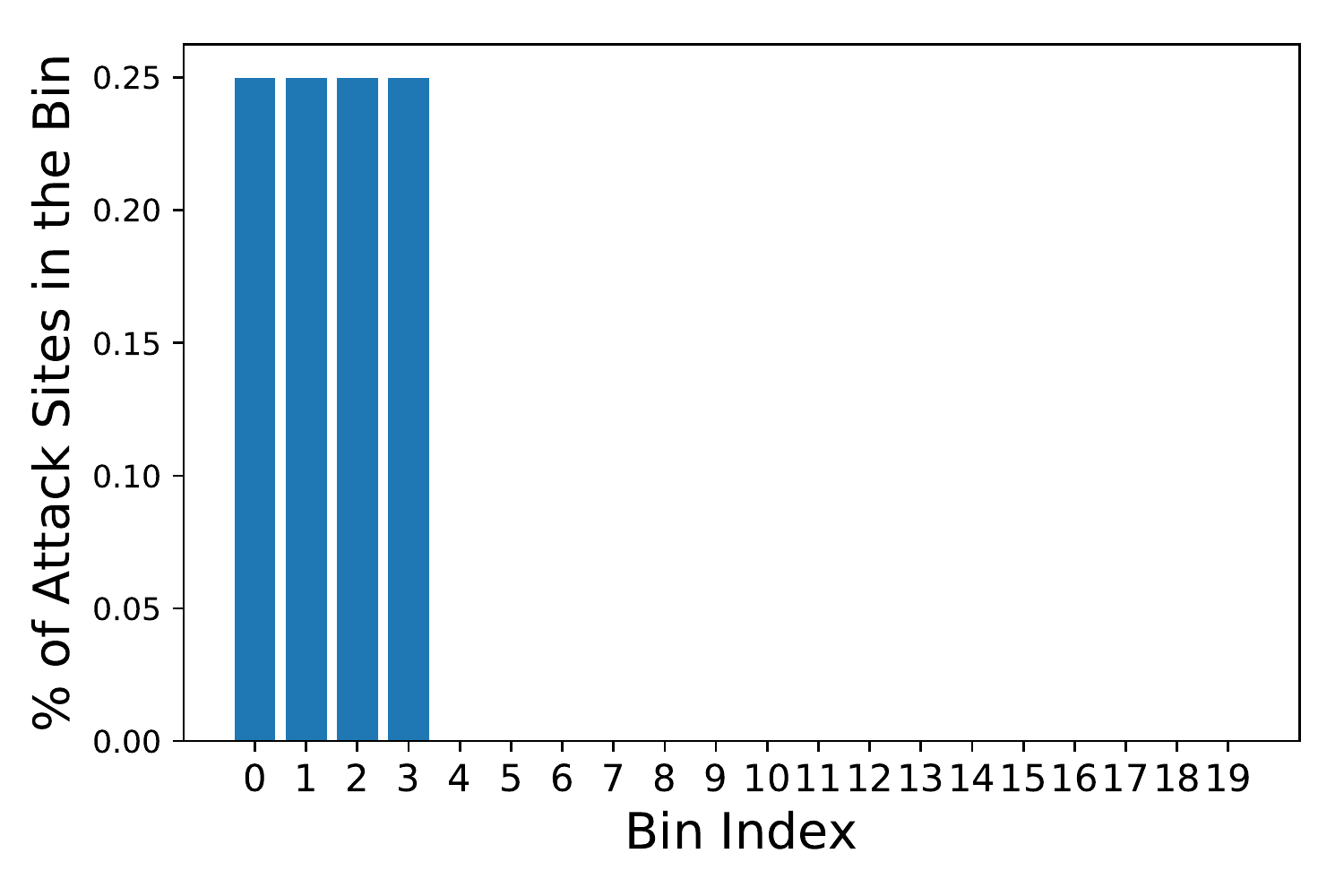} 
    \includegraphics[scale=0.3]{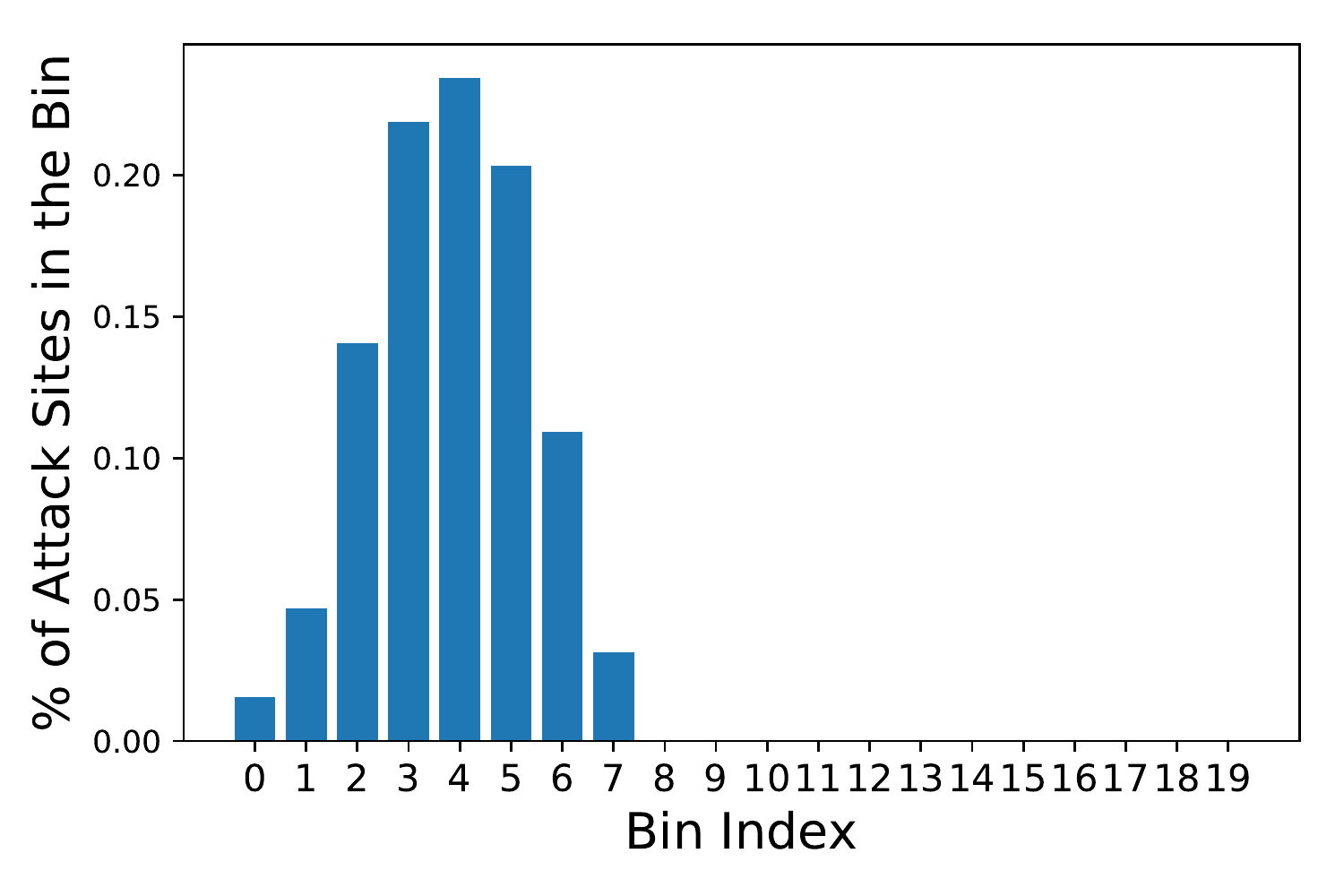}
    \captionof{figure}{UCI Spam Dataset, Interval \textbf{Top row:} Semi-online, \textbf{Bottom row:} Full-online. \textbf{Left to right:} Slow Decay, Fast Decay, Constant}
    \label{fig:spampos}
\end{center}

\end{document}